\documentclass{article}

\usepackage{amsmath,amsfonts,bm}

\def\Figref#1{Figure~\ref{#1}}

\def\Secref#1{Section~\ref{#1}}
\def\eqref#1{equation~\ref{#1}}
\def\Eqref#1{Equation~\ref{#1}}
\def\1{\bm{1}}

\DeclareMathAlphabet{\mathsfit}{\encodingdefault}{\sfdefault}{m}{sl}
\SetMathAlphabet{\mathsfit}{bold}{\encodingdefault}{\sfdefault}{bx}{n}

\DeclareMathOperator*{\argmin}{arg\,min}

\usepackage[sort,authoryear,round]{natbib}
\usepackage{fullpage}
\usepackage[utf8]{inputenc} % allow utf-8 input
\usepackage[T1]{fontenc}    % use 8-bit T1 fonts
\usepackage{xcolor}
\usepackage{hyperref}       % hyperlinks
\definecolor{myblue}{rgb}{0,0.2,0.8}
\hypersetup{ %
pdftitle={},
pdfauthor={},
pdfsubject={},
pdfkeywords={},
pdfborder=0 0 0,
pdfpagemode=UseNone,
colorlinks=true,
linkcolor=myblue,
citecolor=myblue,
filecolor=myblue,
urlcolor=myblue,
pdfview=FitH}
\usepackage{authblk}
\usepackage{url}            % simple URL typesetting
\usepackage{booktabs}       % professional-quality tables
\usepackage{amsfonts}       % blackboard math symbols
\usepackage{amsthm}
\usepackage{nicefrac}       % compact symbols for 1/2, etc.
\usepackage{microtype}      % microtypography

\usepackage{graphicx}
\usepackage{xspace}
\usepackage[varqu,varl,var0,scaled=0.97]{inconsolata}
\usepackage{caption}
\usepackage{overpic}
\usepackage{wrapfig}
\usepackage{tcolorbox}
\usepackage{enumitem}
\usepackage{threeparttable}
\usepackage{subfig}

\usepackage{listings}
\usepackage{color}
\usepackage{lipsum}

\definecolor{dkgreen}{rgb}{0,0.6,0}
\definecolor{gray}{rgb}{0.5,0.5,0.5}
\definecolor{mauve}{rgb}{0.58,0,0.82}

\newcommand{\eg}{\emph{e.g.},\xspace}
\newcommand{\ie}{\emph{i.e.},\xspace}

\definecolor{ao(english)}{rgb}{0.0, 0.5, 0.0}

\usepackage[capitalize,noabbrev]{cleveref}

\newtheorem{theorem}{Theorem}[section]

\newtheorem{corollary}[theorem]{Corollary}

\title{\bf{Subspace-Configurable Networks}}
\author[1]{Dong Wang$^*$}
\author[1,2]{Olga Saukh$^*$}
\author[3]{Xiaoxi He}
\author[4]{Lothar Thiele}
\affil[1]{Graz University of Technology, Austria}
\affil[2]{Complexity Science Hub Vienna, Austria}
\affil[3]{University of Macao, China}
\affil[4]{ETH Zurich, Switzerland}
\affil[ ]{}
\affil[ ]{\texttt{\{dong.wang, saukh\}@tugraz.at, hexiaoxi@um.edu.mo, thiele@tik.ee.ethz.ch}}

\begin{document}
\date{}
\maketitle

\def\thefootnote{*}\footnotetext{Equal contribution}

\begin{abstract}
While the deployment of deep learning models on edge devices is increasing, these models often lack robustness when faced with dynamic changes in sensed data. This can be attributed to sensor drift, or variations in the data compared to what was used during offline training due to factors such as specific sensor placement or naturally changing sensing conditions. Hence, achieving the desired robustness necessitates the utilization of either an invariant architecture or specialized training approaches, like data augmentation techniques. Alternatively, input transformations can be treated as a domain shift problem, and solved by post-deployment model adaptation. In this paper, we train a parameterized subspace of \emph{configurable networks}, where an optimal network for a particular parameter setting is part of this subspace. The obtained subspace is low-dimensional and has a surprisingly simple structure even for complex, non-invertible transformations of the input, leading to an exceptionally high efficiency of subspace-configurable networks (SCNs) when limited storage and computing resources are at stake.
Our source code is online.\footnote{\url{https://github.com/osaukh/subspace-configurable-networks}}
\end{abstract}

\section{Introduction}

In real-world applications of deep learning, it is common for systems to encounter environments that differ from those considered during model training. There are many reasons for this difference between training and post-deployment such as sensor drift, device-to-device  variation, and domain shift in the data compared to what was used during offline training due to factors like a different sensor placement or changing sensing conditions. 
To address the above challenge, there are two primary approaches: designing robust, invariant models and employing domain adaptation techniques. Both strategies aim to mitigate the performance degradation resulting from the discrepancies between the source and the target domains.

Invariant architectures focus on making the model robust, insensitive, or invariant to specific transformations of the input data. This can be achieved by various means, including training with data augmentation~\citep{Botev2022,Geiping2022}, canonicalization of the input data~\citep{Jaderberg2015,kaba2022equivariance}, adversarial training~\citep{Engstrom2017}, and designing network architectures that inherently incorporate the desired invariances~\citep{Marcus2018,kauderer2018cnninv}. 
% ~\citep{Marcus2018,kauderer2018cnninv,Blything2020,Biscione2021}. 
Domain adaptation, on the other hand, seeks to transfer the knowledge acquired from a source domain to a target domain, where the data distributions may differ. This approach leverages the learned representations or features from the source domain and fine-tunes or adapts them to better align with the target domain~\citep{RussoCTC17,ruijiaxu2018}.

Unlike traditional invariant architectures, configurable networks explicitly define invariances by parameterizing desired input data transformations. We introduce subspace-configurable networks (SCNs), which are trained with weights residing in a subspace formed by a few base models. By receiving a parameter vector for an input transformation, SCNs select appropriate high-accuracy model weights from this subspace, effectively isolating the targeted invariances.
\begin{figure*}[t]
    \centering    
    \includegraphics[height=.235\linewidth]{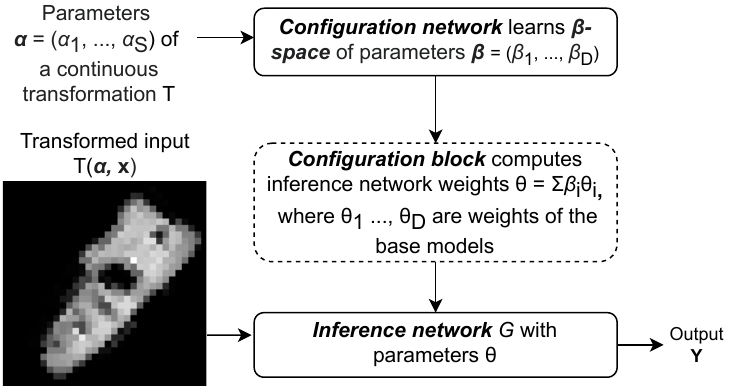}  
    \hspace{-0.2cm}
    \includegraphics[height=.244\linewidth]{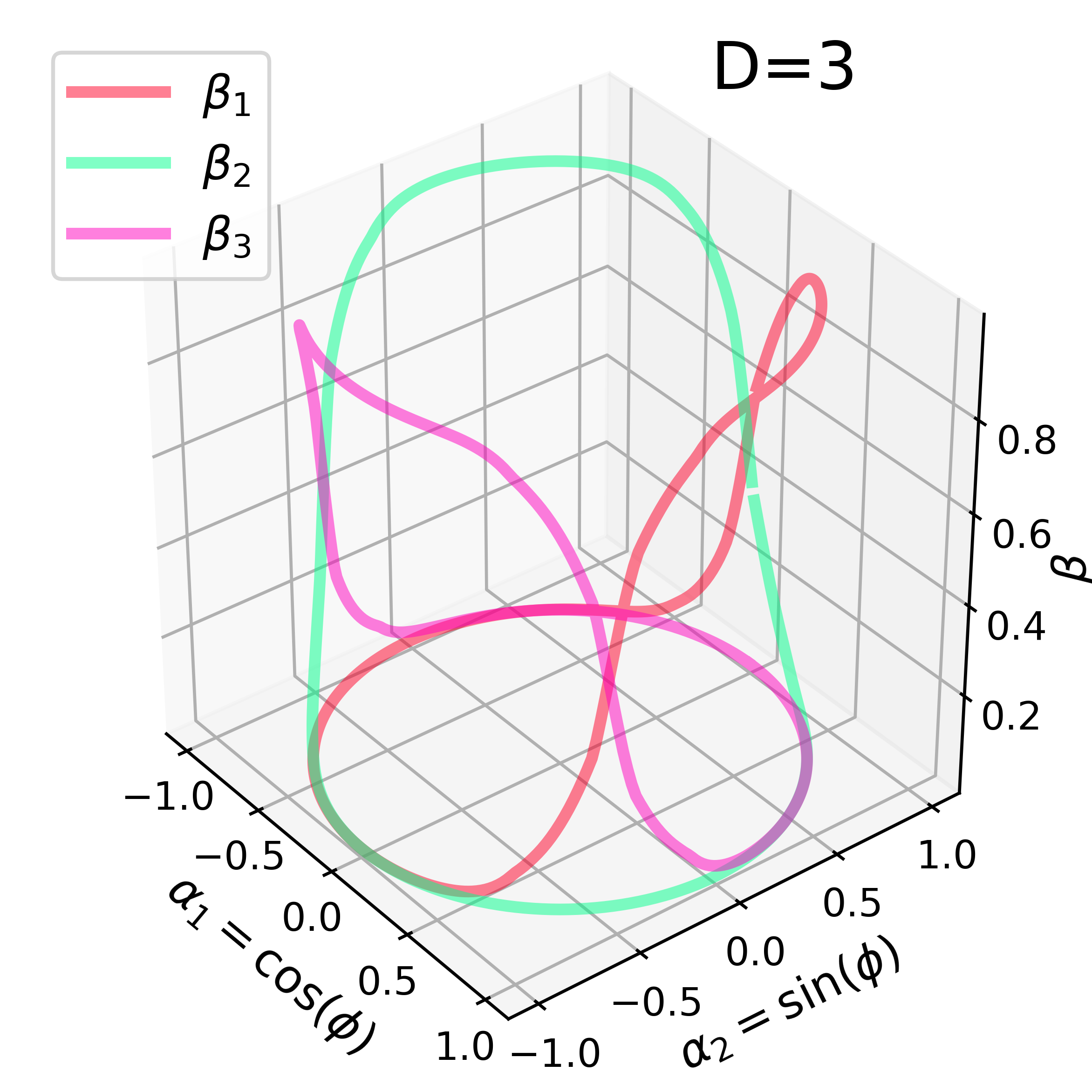}
    \includegraphics[height=.238\linewidth]{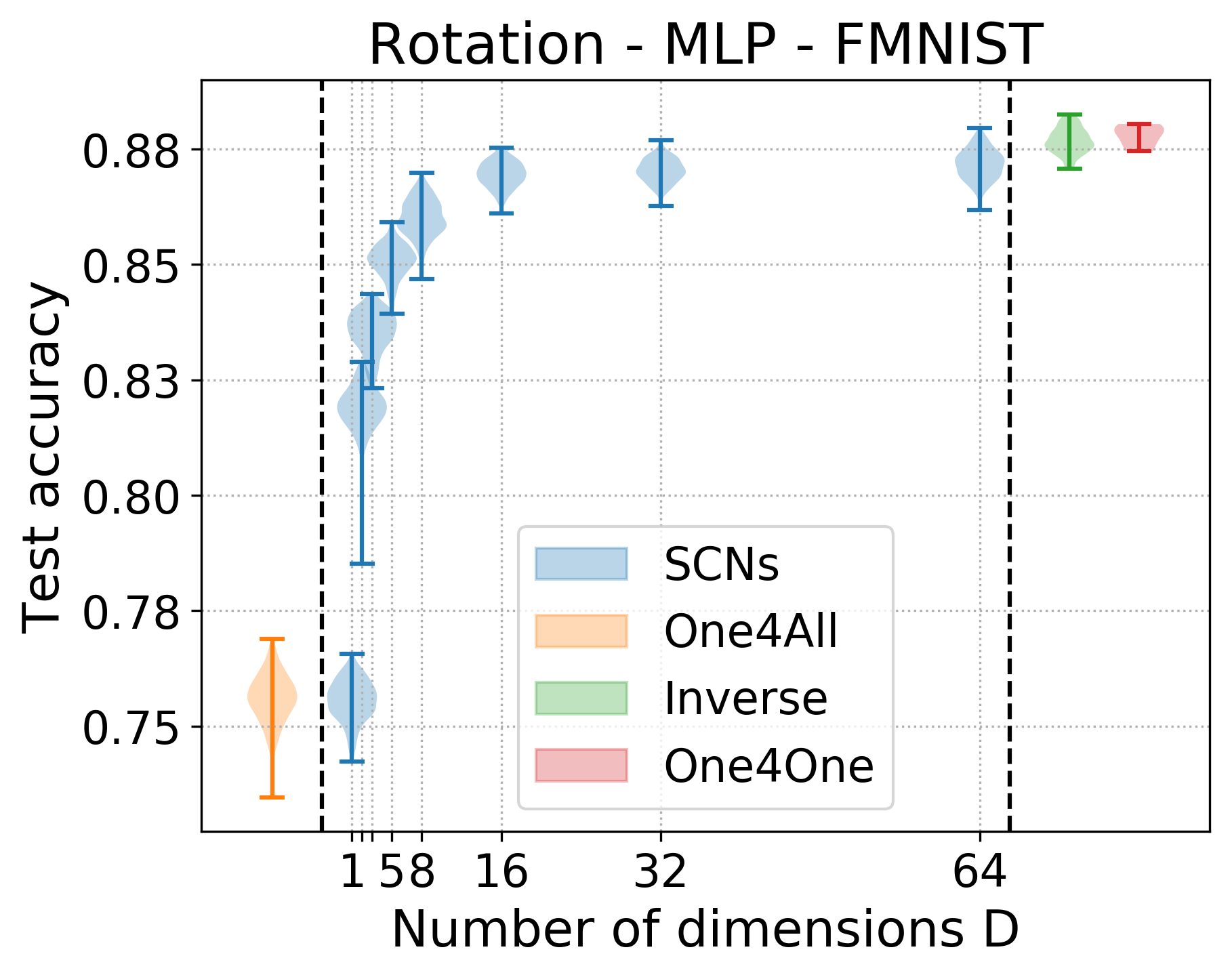}
    \caption{\textbf{Training subspace-configurable networks (SCNs)}, where an optimal network for a fixed transformation parameter vector is part of the subspace retained by few configuration parameters. 
    \textbf{Left:} Given input transformation parameters $\alpha$, \eg a rotation angle for a 2D rotation, we train a \emph{configuration network} which yields a $D$-dimensional configuration subspace (\emph{$\beta$-space}) to construct an efficient inference network with weights $\theta = \sum \beta_i \cdot \theta_i$, where $\theta_i$ are the weights of the \emph{base models}, and $\beta$ is a \emph{configuration vector}.
    \textbf{Middle:} Optimal model parameters in the configuration subspace as function of the rotation angle $\alpha$ given by $(\cos(\phi), \sin(\phi))$ for 2D rotation transformations applied to FMNIST~\citep{xiao2017fashionmnist}. Here SCN has three base models with parameters $\theta_i$ and three configuration vectors $\beta_i$ to compose the weights of the 1-layer MLP inference model. 
    \textbf{Right:} Test accuracy of SCNs with $D=1..64$ dimensions compared to a single network trained with data augmentation (One4All), classifiers trained on canonicalized data achieved by applying inverse rotation transformation with the corresponding parameters (Inverse), and networks trained and tested on datasets where all images are rotated by a fixed degree (One4One).\protect\footnotemark\xspace Each violin shows the performance of a model on all degrees with a discretization step of $1^\circ$, expect for One4One where the models are independently trained and evaluated on $0$, $\pi/6$, $\pi/4$, $\pi/3$, $\pi/2$ rotated input. 
    }
    \label{fig:firstfigure}
\end{figure*}
We evaluate our SCN models by studying 2D translation, scaling, translation, and complex irreversible transformations like 3D rotation-and-projection.
In the appendix, we also evaluate a wide range of real-world transformations covering both computer vision and audio signal processing domains, along with dedicated network architectures. To uncover configuration subspaces for a set of input transformations, SCNs leverage a hypernet-inspired architecture~\citep{ha2016hypernet} to learn optimal inference models for each specific transformation in the set (\Figref{fig:firstfigure} left).
To offer additional insights, we visualize the relation between the input transformation parameters and the configuration vector in the configuration subspace for a number of transformations (an example of the configuration subspace for 2D rotation is shown in \Figref{fig:firstfigure} middle). 
Interestingly, the configuration parameter vectors form well-structured geometric objects, highlighting the underlying structure of the optimal parameter subspaces.
If the inference network capacity is fixed, usually due to severe resource constraints of edge devices, SCNs can quickly beat training with data augmentation (One4All) and match or outperform solutions trained for input transformation parameters optimized for each input transformation separately (One4One), see \Figref{fig:firstfigure} right.
The contributions of this paper are summarized as follows:

\footnotetext{One4One = \emph{one} model for \emph{one} parameter setting, \ie a fixed rotation degree. One4All = \emph{one} model for \emph{all} parameter settings, \ie a model trained with data augmentation for the considered %continuous 
range of parameter values.}

\begin{itemize}
    \item We design \emph{subspace-configurable networks (SCNs)} to learn the configuration subspace and generate optimal networks for specific transformations. The approach presents a highly resource-efficient alternative to model adaptation through retraining and specifically targets resource-constrained devices. Our approach and theoretical insights are covered in Section~\ref{sec:scn} and Appendix~\ref{sec:theory}.
    \item SCNs are evaluated on ten common real-world transformations, using five backbones and five standard benchmark datasets from computer vision and audio signal processing domains. The results are detailed in Section~\ref{sec:eval} and Appendix~\ref{sec:cn:accuracy}-\ref{sec:3d:appendix}.
    \item SCNs take transformation parameters as input, yet these parameters can be estimated from the input data. We provide an algorithm to build a transformation-invariant model on top of SCNs in Section~\ref{sec:scn} and Appendix~\ref{sec:search:appendix}.
    \item In practical IoT scenarios the parameter supply can be replaced with a correlated sensor modality. We implemented SCNs on two resource-constrained devices and show in Section~\ref{sec:eval:iot:fruit} and Appendix~\ref{sec:eval:iot} their outstanding performance and remarkable efficiency.
\end{itemize}
\Secref{sec:conclusion} concludes this paper with a discussion of limitations and outlining future research directions. Further related work is presented in Appendix~\ref{sec:relatedwork}.

\section{Subspace-Configurable Networks}
\label{sec:scn}

\subsection{Transformations and their parameterization}

Let $\mathbb{X} \times \mathbb{Y} = \{(x, y)\}$ be a dataset comprising labelled examples $x \in \mathbb{X} \subset \mathbb{R}^N$ with class labels $y \in \mathbb{Y} \subset \mathbb{R}^M$. 
We apply a transformation $T: \mathbb{R}^S \times \mathbb{R}^N \rightarrow \mathbb{R}^N$ parameterized by the vector $\alpha = (\alpha_1, \cdots, \alpha_S) \in \mathbb{A} \subseteq \mathbb{R}^S$ to each input example $x$. 
A transformed dataset is denoted as $T(\alpha, \mathbb{X}) \times \mathbb{Y} := \{(T(\alpha, x), y)\}$. 
For instance, let $\mathbb{X}$ be a collection of $P \times P$ images, then we have $x\in \mathbb{R}^{P ^2}$ where each dimension corresponds to the pixel intensity of a single pixel.
Transformation $T(\alpha, \mathbb{X}) : \mathbb{A} \times \mathbb{R}^{P ^2} \rightarrow \mathbb{R}^{P ^2}$ is modulated by pose parameters $\alpha$, such as rotation, scaling, translation or cropping. 
We assume that data transformations $T(\alpha, \mathbb{X})$ preserve the label class of the input and represent a continuous function of $\alpha \in \mathbb{A}$, \ie for any two transformation parameters $\alpha_1$ and $\alpha_2$ there exists a continuous curve in $\mathbb{A}$ that connects two transformation parameters. Note that by changing $\alpha$ we transform all relevant data distributions the same way, \eg the data used for training and test.
The set $\{T(\alpha, x)\,|\,\alpha \in \mathbb{A}\}$ of all possible transformations of input $x$ is called an \emph{orbit} of $x$. We consider an infinite orbit defined by a continuously changing $\alpha$.

We consider an inference network to represent a function $g : \mathbb{X} \times \mathbb{R}^L \rightarrow \mathbb{Y}$ that maps data $x \in \mathbb{X}$ from the input space $\mathbb{X}$ to predictions $g(x, \theta) \in \mathbb{Y}$ in the output space $\mathbb{Y}$, where the mapping depends on the weights $\theta \in \mathbb{R}^L$ of the network. 
$E(\theta, \alpha)$ denotes the expected loss of the inference network and its function $g$ over the considered training dataset $T(\alpha, \mathbb{X})$. Since the expected loss may differ for each dataset transformation parameterized by $\alpha$, we write $E(\theta, \alpha)$ to make this dependency explicit. Optimal network parameters $\theta^*_\alpha$ are those that minimize the loss $E(\theta, \alpha)$ for a given transformation vector $\alpha$.

\subsection{Learning configurable networks}

The architecture of SCNs is sketched in \Figref{fig:firstfigure} (left). Excited by the hypernet~\citep{ha2016hypernet} design, we train a configuration network with function $h(\cdot)$ and an inference network with function $g(\cdot)$ connected by a linear transformation of network parameters $\theta = f(\beta)$ computed in the configuration block:
\begin{equation}
    \label{eq:5}
    \theta = f(\beta) = \sum_{i = 1}^{D} \beta_i \cdot \theta_i,
\end{equation}
where $\theta_i \in \mathbb{T} \subseteq \mathbb{R}^L$ for $i \in [1, D]$ denote the static weights (network parameters) of the base models that are the result of the training process. The configuration network with the function $h: \mathbb{R}^S \rightarrow \mathbb{R}^D$ yields a low-dimensional \emph{configuration space} of vectors $\beta \in \mathbb{R}^D$, given transformation parameters $\alpha \in \mathbb{A}$. Along with learning the mapping $h$, we train the $D$ \emph{base models} with weights $\theta_i \in \mathbb{R}^L$ to construct the weights of inference networks $\theta = f(\beta)$.
The SCN training process minimizes the expected loss $E(\theta,\alpha) = E(f(h(\alpha)), \alpha)$ to determine the configuration network with function $h$ and the base model parameters $\theta_i$. We use the standard categorical cross-entropy loss in all our experiments. During inference, a transformed input example $\hat{x} = T(\alpha, x)$ is classified by the inference network with weights $\theta$, $\beta = h(\alpha)$ and $y = g(\hat{x}, f(h(\alpha)))$. 

Note that degenerated solutions, where $\beta$ is constant for all $\alpha$ are part of the solution space. 
In this case, SCN ends up representing a single model for all transformation parameters $\alpha \in \mathbb{A}$, which is essentially the One4All model, \ie a model trained with data augmentation over all transformation parameters $\alpha \in \mathbb{A}$. To avoid degenerated cases, we enhance the cross-entropy loss function with a regularization term as a squared cosine similarity $\cos^2(\beta^{(1)},\beta^{(2)})$ between the configuration vector $\beta^{(1)}$ for a randomly chosen $\alpha^{(1)}$ applied to transform the current batch, and a vector $\beta^{(2)}$ obtained from the configuration network for another randomly sampled $\alpha^{(2)} \in \mathbb{A}$. The applied regularization (with a weighting factor of 1.0) improves the performance of SCNs by reinforcing them to construct unique dedicated inference networks for different transformation parameters $\alpha$.

For a continuous transformation $T(\alpha)$, in the next section and in Appendix~\ref{sec:theory} we provide theoretical results that help to understand the structure of the $\beta$-space using continuity arguments of the optimal solution space.

\subsection{Continuity of the learned subspaces}
\label{sec:continuity}

\Figref{fig:firstfigure} (middle) exemplifies a learned $\beta$-space for the 2D rotation transformation learned by SCN for $D=3$ with a MLP inference network architecture trained on FMNIST. Transformation parameters $(\alpha_1, \alpha_2) = (\cos(\phi), \sin(\phi))$ with $\phi=0..2\pi$ yield 3-dimensional $\beta$ vectors $(\beta_1, \beta_2, \beta_3)$ with each $\beta_i$ being in charge of a specific contiguous region of the $\alpha$-space. Transitions between regions are covered by models that are a linear combination of optimal solutions for other $\alpha$ values. This structural property of the $\beta$-space is independent of the dataset and architecture we used to train SCNs, as shown in the next section. Moreover, we observe a continuous change of $\beta$ as we change $\alpha$. 

Another observation we make from \Figref{fig:firstfigure} (right) is that SCNs match the high performance of the baselines already for small number of dimensions $D$ of the linear subspace. In other words, the solution subspace spanned by only $D$ base models learned by SCN pays no penalty for its very simple structure.

We provide theoretical results that help to understand the structure of the $\beta$-space using continuity arguments of the optimal solution space. Informally, the following theorem shows under certain conditions that for every continuous curve connecting two transformation parameters in $\mathbb{A}$, there exists a corresponding continuous curve in the network parameter space $\mathbb{T}$. These two curves completely map onto each other where the network parameters are optimal for the corresponding data transformations. In particular, the curve in the network parameter space $\mathbb{T}$ is continuous. 

To simplify the formulation of the theorems (see Appendix~\ref{sec:theorem:continuity} for the respective proofs), we suppose that the set of admissible parameters $\theta \in \mathbb{T}$ is a bounded subspace of $\mathbb{R}^L$ and all optimal parameter vectors (weights) $\theta^*_\alpha$ are in the interior of $\mathbb{T}$.

\begin{theorem}[\textbf{Continuity}] \label{th:1:short}
Suppose that the loss function $E(\theta, \alpha)$ satisfies the Lipschitz condition
\begin{equation}\label{eq:lipshitz}
    \begin{split}
        |E(\theta, \alpha^{(2)}) - E(\theta, \alpha^{(1)}| &\leq K_\alpha ||\alpha^{(2)} - \alpha^{(1)} ||_2
    \end{split}
\end{equation}
\noindent for $\alpha^{(1)}, \alpha^{(2)} \in \mathbb{A}$, and $E(\theta, \alpha)$ is differentiable w.r.t. to $\theta$ and $\alpha$.
Then, for any continuous curve $\alpha(s) \in \mathbb{A}$ with $0 \leq s \leq \hat{s}$ in the parameter space of data transformations there exists a corresponding curve $\theta(t) \in \mathbb{T}$ with $0 \leq t \leq \hat{t}$ in the parameter space of network weights and a relation $(s, t) \in R$ such that 
\begin{itemize}[topsep=0pt]
\itemsep0em 
\item 
    the domain and range of $R$ are the intervals $s \in [0, \hat{s}]$ and $t \in [0, \hat{t}]$, respectively, and
\item 
    the relation $R$ is monotone, \ie if $(s_1, t_1), (s_2, t_s) \in R$ then $(s_1 \geq s_2) \Rightarrow (t_1 \geq t_2)$, and 
\item
    for every $(s, t) \in R$ the network parameter vector $\theta(t)$ minimizes the loss function $E(\theta, \alpha)$ for the data transformation parameter $\alpha(s)$.
\end{itemize} 
\end{theorem}

We are also interested in the relation between $\alpha$ and corresponding optimal vectors $\beta$ that define optimal locations on the linear subspace of admissible network parameters as defined by (\ref{eq:5}). To simplify the formulation of the further theoretical result and proof, we suppose that $\beta \in \mathbb{B}$ where $\mathbb{B}$ is a bounded subspace of $\mathbb{R}^D$, and all basis vectors $\theta_j$ that define $f(\beta)$ in (\ref{eq:5}) have bounded elements. Under these assumptions, we can derive a corollary from Theorem~\ref{th:1:short}. 

\begin{corollary} \label{th:2:short}
Suppose that the loss function $E(\theta, \alpha)$ satisfies (\ref{eq:lipshitz}), and for any $\alpha \in \mathbb{A}$ all local minima of $E(f(\beta), \alpha)$ w.r.t. $\beta$ are global. Then the following holds: For any continuous curve $\alpha(s) \in \mathbb{A}$ with $0 \leq s \leq \hat{s}$ in the parameter space of data transformations there exists a corresponding curve $\beta(t) \in \mathbb{B}$ with $0 \leq t \leq \hat{t}$ on the linea r network parameter subspace according to (\ref{eq:5}) and a relation $(s, t) \in R$ such that 
\begin{itemize}[topsep=0pt]
\itemsep0em
\item 
    the domain and range of $R$ are the intervals $s \in [0, \hat{s}]$ and $t \in [0, \hat{t}]$, respectively, and
\item 
    the relation $R$ is monotone, \ie if $(s_1, t_1), (s_2, t_s) \in R$ then $(s_1 \geq s_2) \Rightarrow (t_1 \geq t_2)$, and 
\item
    for every $(s, t) \in R$ the network parameter vector $\beta(t)$ minimizes the loss function $E(f(\beta), \alpha)$ for the data transformation parameter $\alpha(s)$.
\end{itemize} 
\end{corollary}

The proof of the corollary is in Appendix~\ref{sec:theorem:continuity}. The above corollary provides a theoretical argument for the continuous curves in Figure~\ref{fig:firstfigure} (middle), \ie the curves explicitly show the relation $R$. The existence of such a relation $R$ is also apparent for all the dataset-architecture pairs used to empirically evaluate SCNs in \Secref{sec:eval}. 

Appendix~\ref{sec:theorem:smallchanges} contains a further result: Small changes to transform parameters $\alpha$ result in small changes of optimal configuration vectors $\beta^*_\alpha$ for suitable loss functions $E(f(\beta), \alpha)$. In other words, the relation $R$ can be represented as a continuous function $r$ with $t = r(s)$, \ie the parameter vector $\beta(t)$ that minimizes the loss function can be determined as a function of the data transformation $\alpha(s)$.

\subsection{Search in the $\alpha$-space and practical value of SCNs}
\label{sec:searchalpha}
SCNs take transformation parameter $\alpha$ as input. However, one can also use a search algorithm to estimate $\alpha$ from the input data, aiming for low-entropy, confident classification results. This approach, inspired by previous techniques \citep{Wortsman2020supsup,Hendrycks2016}, utilizes the basin hopping method~\cite{Iwamatsu_2004} targeting hard nonlinear optimization problems with a mix of global and local search phases. To enhance performance, SCN training includes a regularizer to optimize output entropy based on the correct $\alpha$. Despite the computational demands of the $\alpha$-search algorithm, it achieves high accuracy, allowing SCNs to serve as transformation-invariant networks. In practice, however, search in the $\alpha$-space can often be avoided. 
If the transformation of interest is discrete and limited to a few cases, $\alpha$-search can be reduced to running inference over a few candidate models. Most importantly, however, the parameter $\alpha$ can be inferred from a correlated sensor modality. We demonstrate this in two edge applications discussed in Section~\ref{sec:eval:iot:fruit} and Appendix~\ref{sec:eval:iot}.

Several peculiarities of SCNs' design make them well-suited for resource-constrained devices. 
(1) SCNs take advantage of memory hierarchies. Fast memory, such as SRAM, offers access times in the range of nanoseconds, but it is usually available in smaller capacities. In contrast, slow memory like Flash or EEPROM, provides larger storage capacities, but with slower access times. Slow memory is ideal for storing less time-sensitive SCN reconfiguration data, such as the $D$ base models holding $\theta_i$ and the parameters of the configuration network. At the same time, the inference network weights $\theta$ should better completely fit into RAM to support fast inference. 
(2) SCNs yield the most benefit if memory and thus the network capacity are limited. Given unlimited resources, One4All can match the performance of SCNs. Exceptions include corner cases where the transformation parameter $\alpha$ is used to break symmetries.
(3) SCNs draw inspiration from the recent linear mode connectivity literature~\citep{entezari2021role, ainsworth2022git}, and empirically show that a linear reconfiguration function $f(\beta)$ yields great performance of models for different transformation parameters. This decision allows implementing efficient memory access strategies for SCN reconfiguration, such as reuse of memory pages. Also computation of \eqref{eq:5} can be efficiently realized with hardware-friendly linear vector operations, \ie using MLA and FMA instructions, vectorization, and pipelining.

\section{Experimental Results}
\label{sec:eval}

\begin{figure*}[t]
    \centering
    \begin{minipage}[c]{.62\linewidth}
    \includegraphics[width=.49\linewidth]{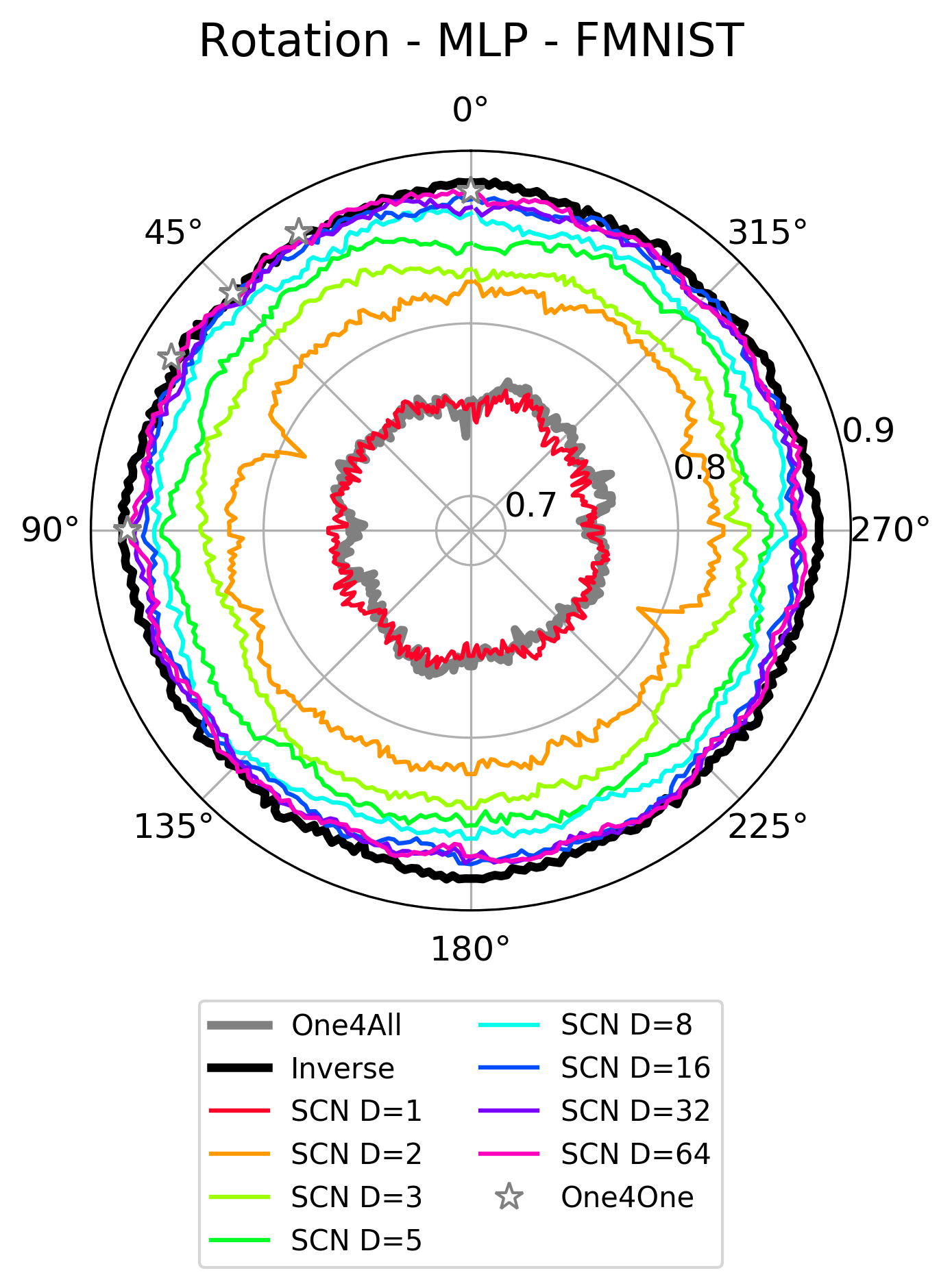}
    \includegraphics[width=.49\linewidth]{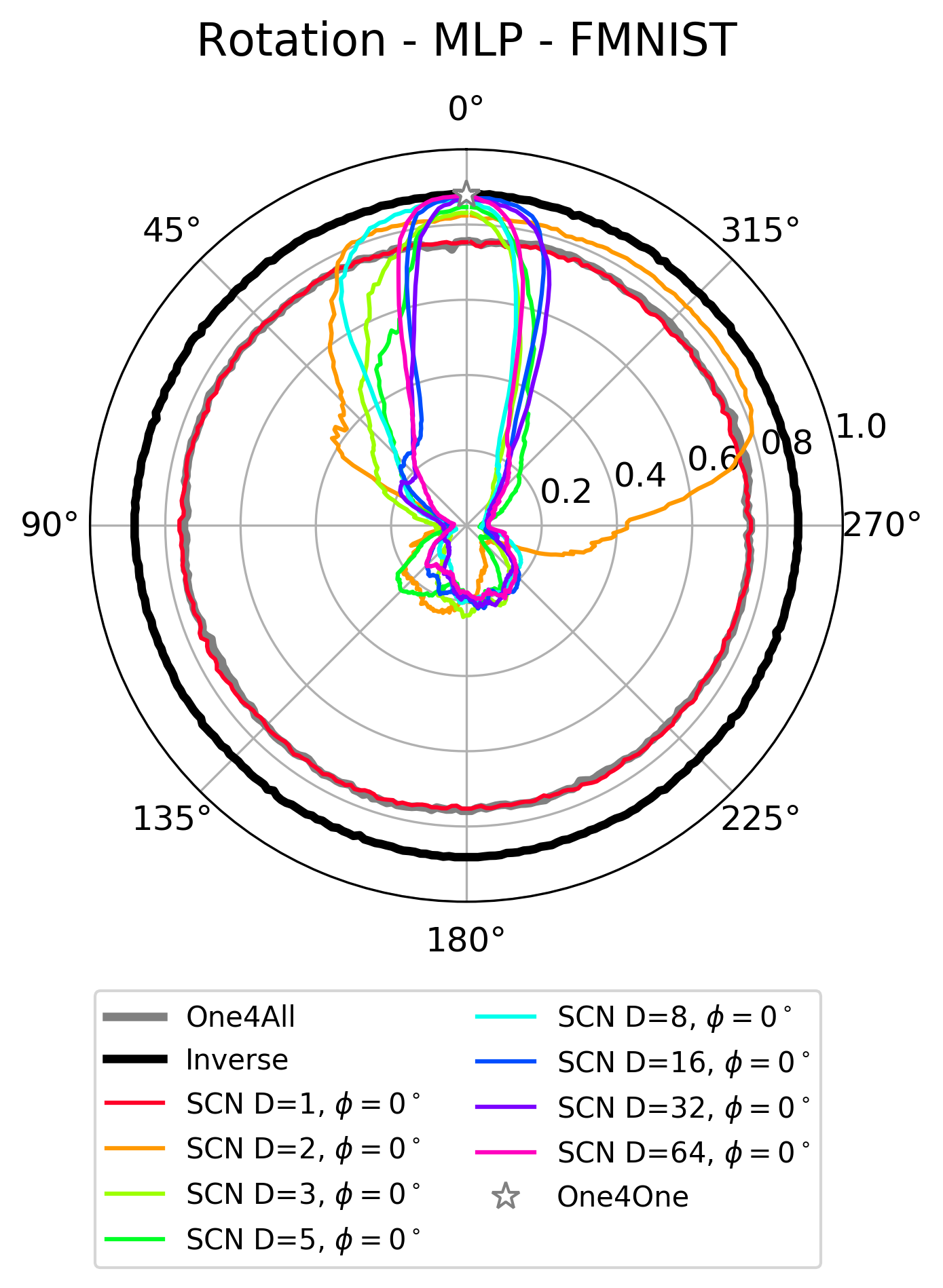}
    \end{minipage}
    \hspace{0.7cm}
    \begin{minipage}[c]{.27\linewidth}
        \centering
        \includegraphics[width=\linewidth]
        {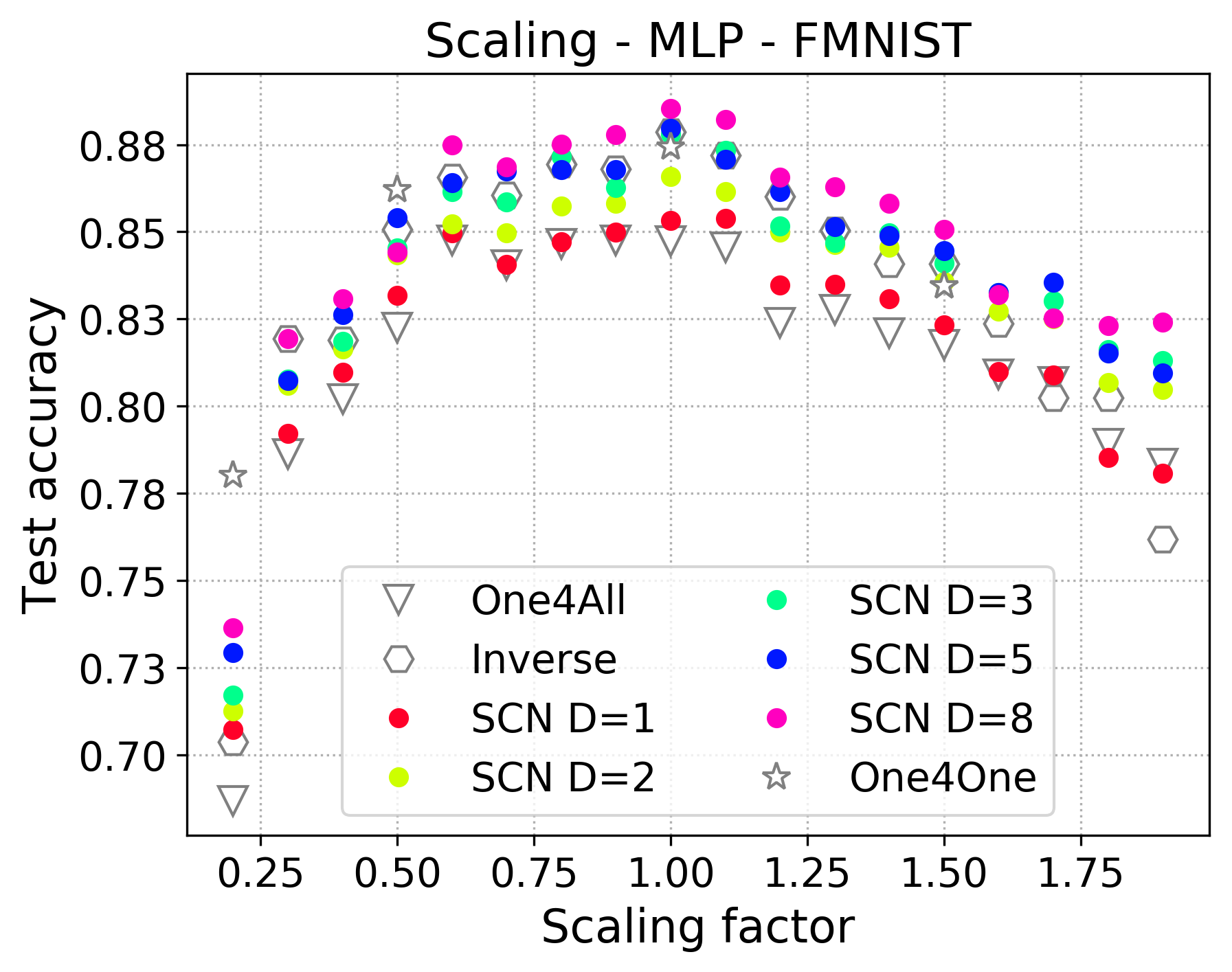} 
        
        \includegraphics[width=\linewidth]
        {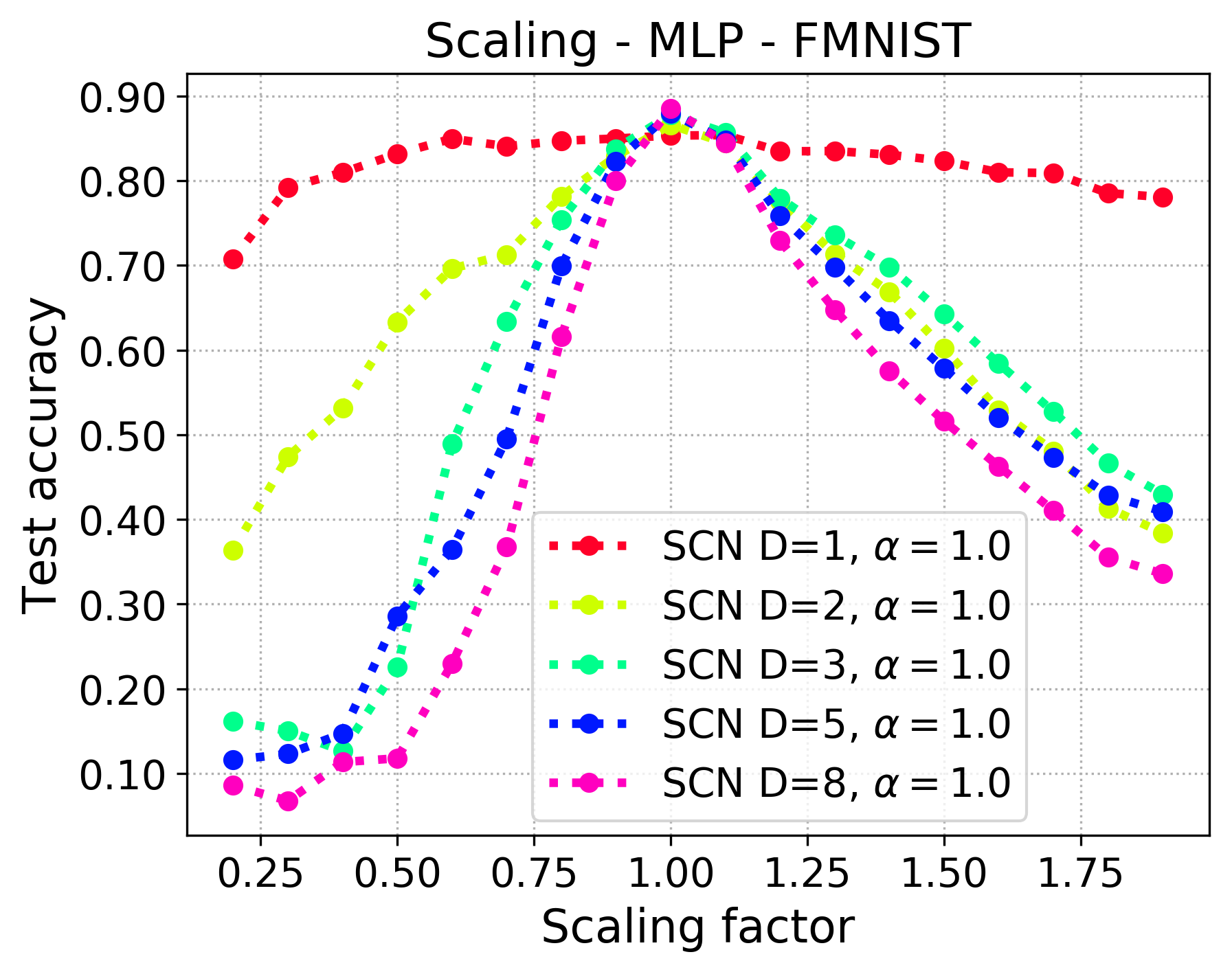}
    \end{minipage}

    \caption{\textbf{SCN test accuracy for 2D rotation and scaling transformations.} 
    \textbf{Left and middle:} 2D rotation parameterized by a rotation degree $\phi=0..2\pi$ input to the configuration network as $\alpha=(\cos(\phi), \sin(\phi))$. For each $\alpha$, SCN determines a configuration vector $\beta$ used to build a dedicated model for every angle shown on the right. The middle polar plot shows the performance of a single model ($\phi=0^\circ$) on all angles. The model works best for the input transformed with $T(\phi=0^\circ)$. Inference network architecture is a 1-layer MLP with 32 hidden units trained on FMNIST. The models constructed by SCN outperform One4All approaching Inverse and One4One accuracy already for small $D$.
    \textbf{Right top:} Scaling transformation parameterized by the scaling factor $\alpha=0.2..2.0$. \textbf{Right bottom:} SCN performance of a single model ($\alpha=1.0$) on all inputs. The dedicated model gets increasingly specialized for the target input parameters with higher $D$. Inference network is a 5-layer MLP with 32 hidden units in each layer trained on FMNIST. Also see Appendix~\ref{sec:cn:accuracy} and videos showing SCN inference models for each parameter setting.\protect\footnotemark
    }
    \label{fig:mlpb:accuracy}
\end{figure*}

\footnotetext{\url{https://tinyurl.com/2nb8k644}}

We evaluate the performance of SCNs on 10 popular transformations (2D rotation, scaling, translation, brightness, contrast, saturation, sharpness, 3D rotation-and-projection, pitch shift and audio speed change) and five dataset-architecture pairs from computer vision and audio signal processing domains (MLPs on FMNIST~\citep{xiao2017fashionmnist}, ShallowCNNs~\citep{neyshabur2020learning} on SVHN~\citep{SVHN}, LeNet-5~\citep{lecun1998} on ModelNet10~\citep{3dShapeNets}, ResNet18~\citep{he2015resnet18} on CIFAR10~\citep{cifar100}, and M5~\citep{DaiDQLD16} on Google Speech Commands Dataset~\citep{warden2019SC}). All considered transformations are continuous, and their parameterization is straightforward. For example, a rotation angle for a 2D rotation.
The main paper exemplifies the obtained results to highlight SCN performance, while additional plots can be found in Appendix~\ref{sec:cn:accuracy}-\ref{sec:3d:appendix}. Training hyper-parameters, architectural choices, dataset description and samples of the transformed images are presented in Appendix~\ref{sec:implementationdetails}.

We compare SCNs to the following baselines. \emph{One4All} represents a network trained with data augmentation obtained by transforming the input by randomly chosen parameters $\alpha \in \mathbb{A}$. \emph{Inverse} classifier is trained on a canonicalized data representation achieved by first applying the inverse transformation to the transformed input. Note that 2D rotation is a fully invertible transformation in theory, yet introduces small distortions in practice due to rounding effects. Translation is fully invertible if the relevant part of the input stays within the input boundaries. Scaling brings significant distortion to the input, and inversion leads to a considerable loss of input quality. Finally, \emph{One4One} represents a set of networks, each trained and tested on the dataset transformed with a fixed parameter vector $\alpha$. Note that for a fixed architecture, dataset, and loss function, a well-trained One4One baseline achieves the best in-distribution generalization. In this sense, it upper bounds the performance which can be achieved by any domain adaptation method using the same data. When comparing model performance throughout this work, all baselines feature \emph{the same} model architecture and have \emph{the same} capacity as the SCN inference network. 
We use a 1-layer MLP with 64 hidden units as the configuration network architecture to learn the configuration subspace $\beta = h(\alpha)$. Our main evaluation metric is the test accuracy, but we also analyze the impact of SCN dimensionality $D$ on its performance, and the structure of the $\beta$-space.

\begin{figure*}[t]
    \centering
     \includegraphics[width=.24\linewidth]{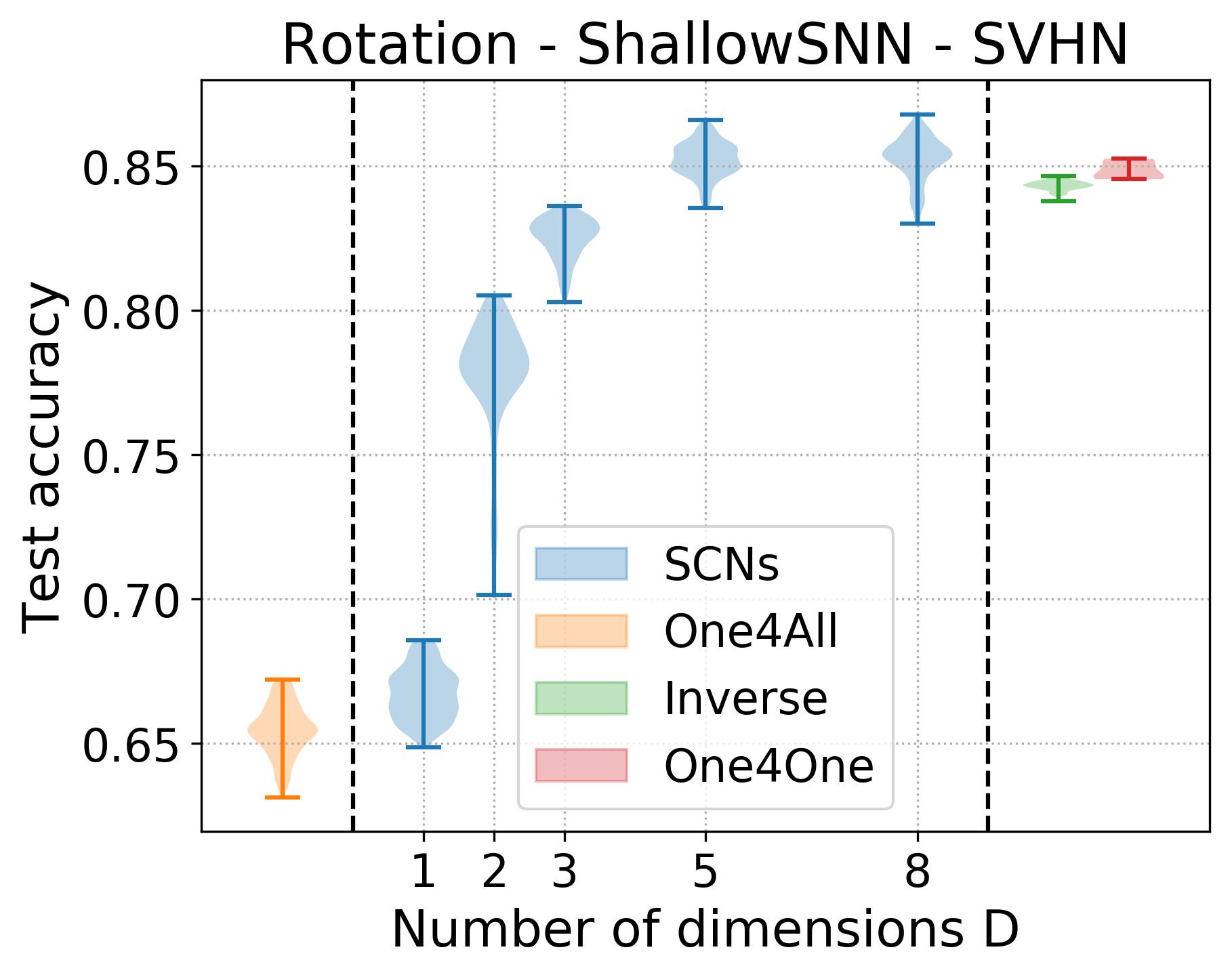}
     \includegraphics[width=.24\linewidth]{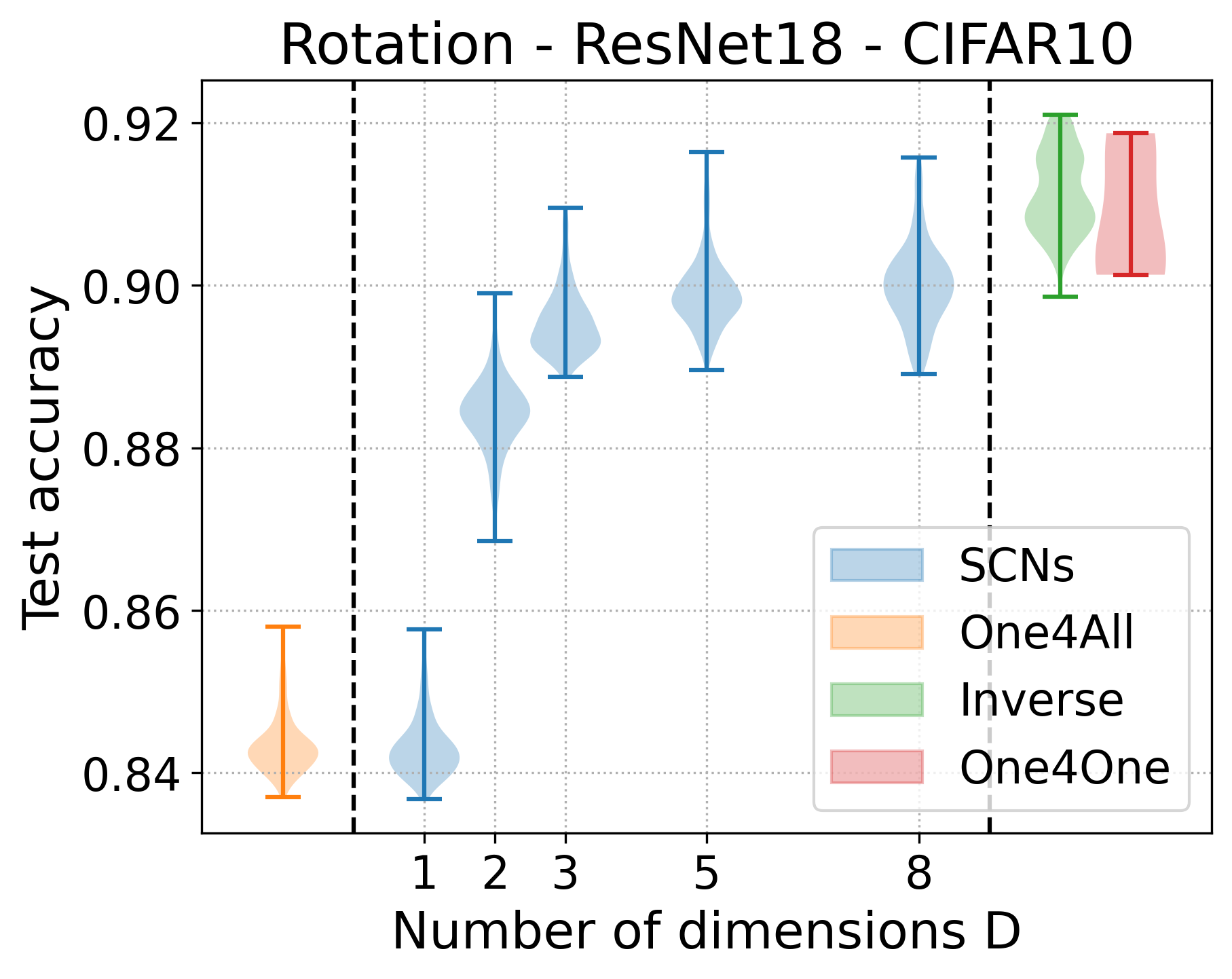}
     \includegraphics[width=.24\linewidth] {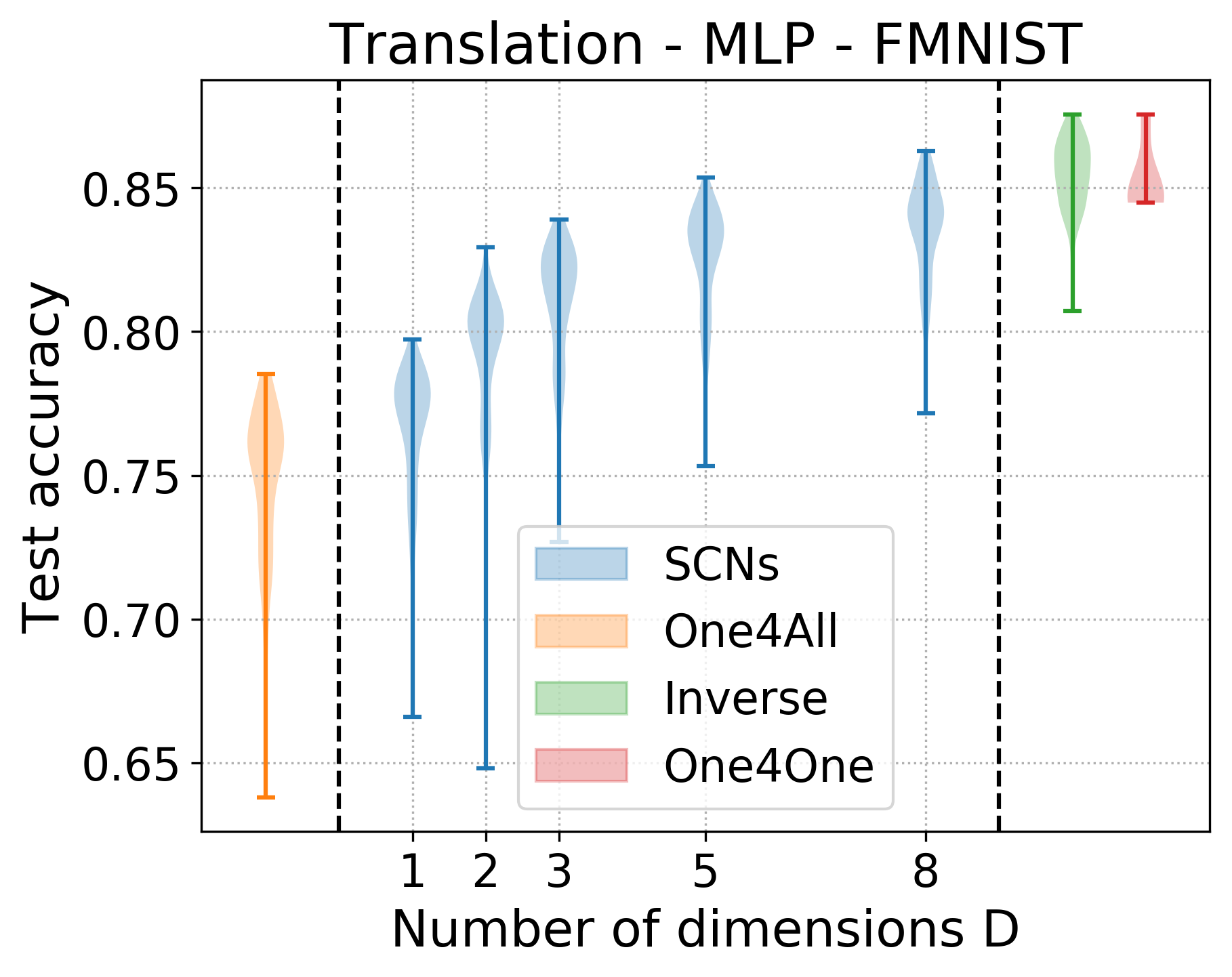}
     \includegraphics[width=.24\linewidth] {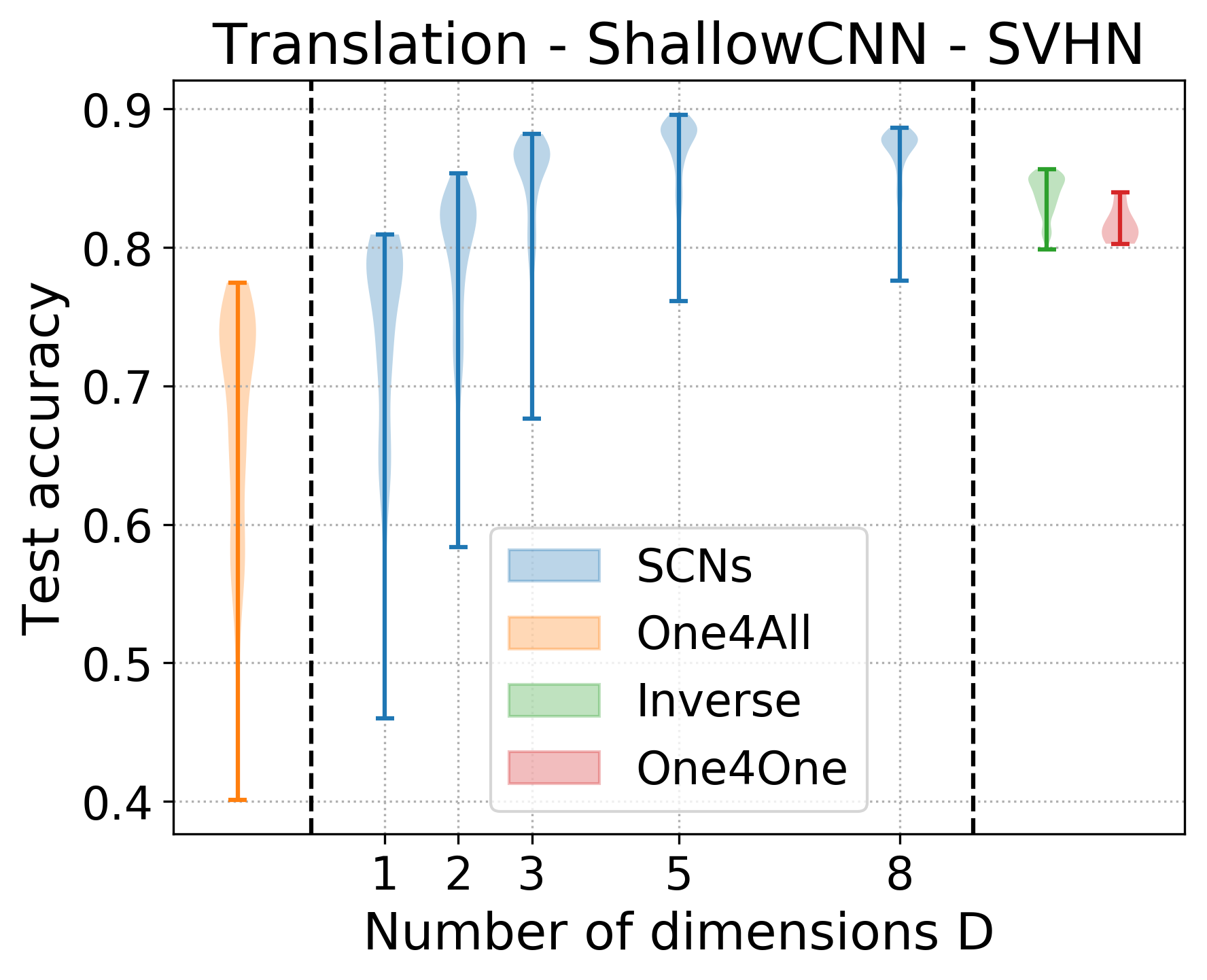}
    \caption{\textbf{SCNs achieve high test accuracy already for low $D$}, outperforming One4All and approaching (and in some cases outperforming) both Inverse and One4One baselines. 
    \textbf{2 plots on the left:} 2D rotation on ShallowCNN--SVHN and ResNet18--CIFAR10. 
    \textbf{2 plots on the right:} Scaling on MLP--FMNIST and ShallowCNN--SVHN. The plots are complementary to \Figref{fig:mlpb:accuracy} evaluating the performance of SCN on different transformations and dataset-architecture pairs. For translation, the violin for One4One comprises prediction accuracy of independently trained models for (0,0) and ($\pm$8,$\pm$8) shift parameters. A detailed evaluation of SCNs for translation is detailed in 
    Appendix~\ref{sec:translation}.}
    \label{fig:effect:d}
\end{figure*}

\subsection{SCN test set accuracy}

\Figref{fig:mlpb:accuracy} and \Figref{fig:effect:d} present different views on the SCN test accuracy as a function of the number of dimensions $D$ when the concept is applied to different transformations, datasets, and architectures. \Figref{fig:mlpb:accuracy} left shows the performance of SCNs on $0-2\pi$ rotation angles. The test accuracy for $D=1$ matches One4All but quickly approaches Inverse and One4One baselines for higher $D$. For the scaling transformation shown in \Figref{fig:mlpb:accuracy} top right, SCNs for $D>1$ easily outperform One4All. They also outperform Inverse for $\alpha < 0.3$ and $\alpha > 1.2$ already for small $D$. Non-invertible transformations introduce significant distortion to the input data complicating feature re-use across inputs for different $\alpha$. A large performance gap between One4One and Inverse for $\alpha=0.2$ suggests that at small scales different features in the input become more relevant than in the original dataset. In some cases, SCNs achieve higher accuracy than One4One networks trained and tested only on the transformed data for some fixed value of $\alpha$, since One4One does not make use of data augmentation but SCN implicitly does due to its structure given in \Eqref{eq:5}.

\Figref{fig:effect:d} presents an aggregated view on the SCN test accuracy for 2D rotations on ShallowCNN--SVHN and ResNet18--CIFAR10, and also for translation on MLP--FMNIST and ShallowCNN--SVHN. Each violin comprises accuracies achieved by models tested on all parameter settings traversed with a very small discretization step (with a granularity of $1^\circ$ and 1 pixel for 2D rotation and translation respectively). The only exception here is the One4One baseline, where a violin comprises the performance of five models independently trained and tested on the transformed inputs for a fixed parameter setting. 
The fixed parameters are chosen to cover $\mathbb{A}$ from the most beneficial (\eg $\alpha=(0,0)$ for translation) to the most suboptimal ($\alpha=(\pm8,\pm8)$ for translation) setting. This is why the violins for the translation transformation have a long tail of low accuracies. The performance of SCNs is consistent across dataset-architecture pairs, matching the best performing baselines already for a small number of dimensions $D$ (also see Appendix~\ref{sec:cn:accuracy}).

\begin{figure*}[t]
    \centering
    \includegraphics[width=\linewidth]{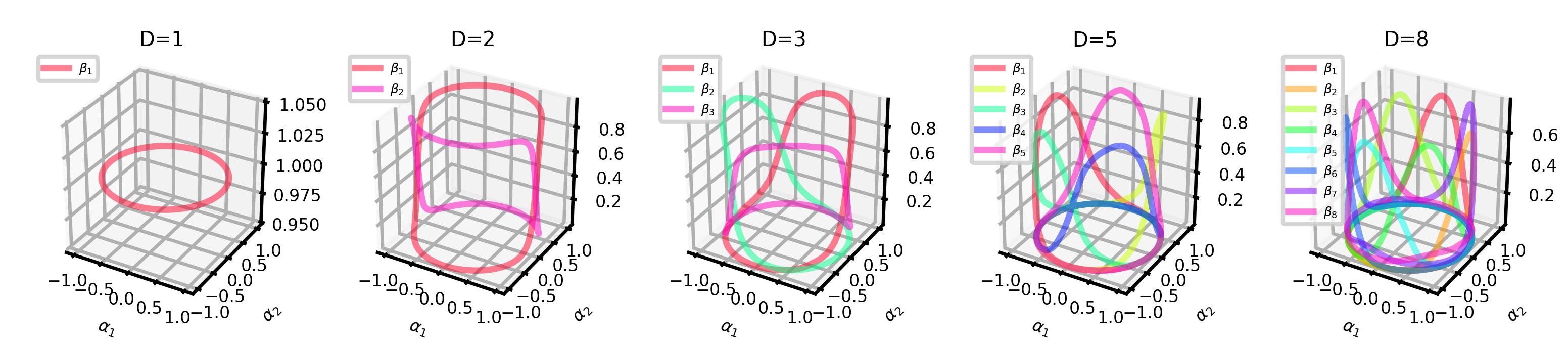}
    \vskip -0.1cm
    \includegraphics[width=\linewidth]{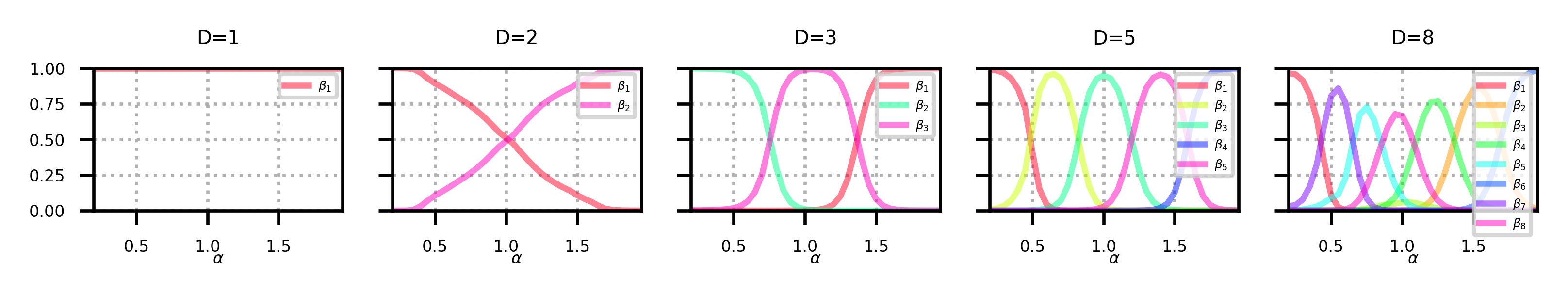}
    \vskip -0.5cm
    \includegraphics[width=\linewidth]{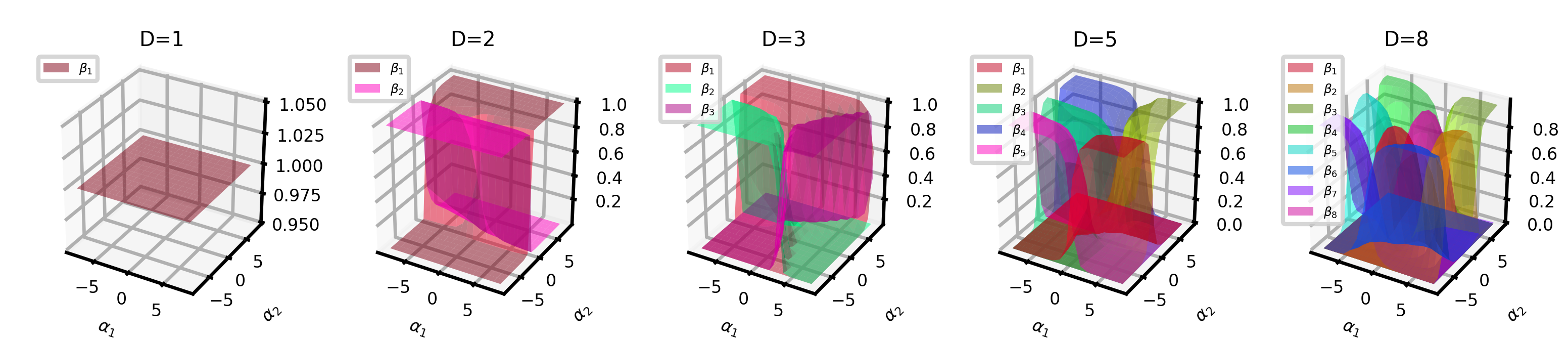}
    \caption{\textbf{A typical view of the $\beta$-space for 2D rotation, scaling and translation, $D=1..8$}. The $\beta$-space is nicely shaped, with each $\beta$ being responsible for a specific range of inputs with smooth transitions.
    \textbf{Top:} SCNs for 2D rotation on ResNet18--CIFAR10. Transformation parameters are a vector $\alpha = (\alpha_1, \alpha_2) = (\cos(\phi), \sin(\phi))$, with $\phi$ being a rotation angle. 
    \textbf{Middle:} SCNs for scaling on ShallowCNN--SVHN, with a scaling factor $\alpha$ between 0.2 and 2.0. 
    \textbf{Bottom:} SCNs for translation on MLP--FMNIST.  A shift is specified by two parameters $(\alpha_x, \alpha_y)$ varying in the range (-8,8) along $x$ and $y$ axes. A visualization for other dataset-architecture pairs is presented in Appendix~\ref{sec:betaspace}. 
    }
    \label{fig:beta}
\end{figure*}

\subsection{Structure of the configuration subspace}

The $\beta$-space learned by the configuration network $h$ for different transformations, datasets, and inference network architectures is shown in \Figref{fig:beta}. For 2D rotation, the transformation parameters $\alpha = (\cos(\phi), \sin(\phi))$ are drawn from a circle and result in all $\beta_i$ being continuous curves arranged around the cylinder in our $\alpha$-$\beta$ visualization. For all transformations, if $D=1$, the SCN training yields $\beta_1=1$ due to the use of softmax in the last layer of the configuration network and a single base model. For $D \neq 1$, each $\beta_i$ is high for a certain contiguous range of $\alpha$s and low outside of this range. For small $D$, the regions of high $\beta$s are largely disjoint, yet overlap as $D$ is scaled up. Interestingly, the shape of the learned transformation is preserved across datasets and inference network architectures, although minor differences do exist, see Appendix~\ref{sec:betaspace}. 

We claim that the subspace of optimized configurations for data transformations parameterized by $\alpha$ is \emph{nicely structured}: (i) We achieve good accuracies even for a linear subspace of low degrees of freedom $D$. (ii) We observe a nice structure of optimal solutions in the space, as represented by the function $\beta = h(\alpha)$ and supported by our theoretical results. 
This finding is related to the recent literature on linear mode connectivity of independently trained solutions~\citep{entezari2021role}, solutions that share a part of the training trajectory~\citep{frankle2020linear}, and those trained on data splits~\citep{ainsworth2022git, keller2022repair}. SCNs establish linear mode connectivity between models trained for different transformation parameters, enhancing the existing literature. 

Although the inference network architecture seems to have little impact on the shape of the learned $\beta$-space, there are interesting exceptions. How does configuration subspace look like if the inference network architecture is invariant to the applied transformation? We trained a SCN for translation with a translation-invariant CNN as inference network architecture. The learned configuration space, in this case, appears degenerated with only one translation-invariant model being learned, regardless of $D$, \ie all but one $\beta_i$ are zero for all transformation parameters $\alpha$ (see Appendix \ref{sec:translation}).

\begin{figure*}[t]
\centering
     \includegraphics[width=.24\linewidth]
     {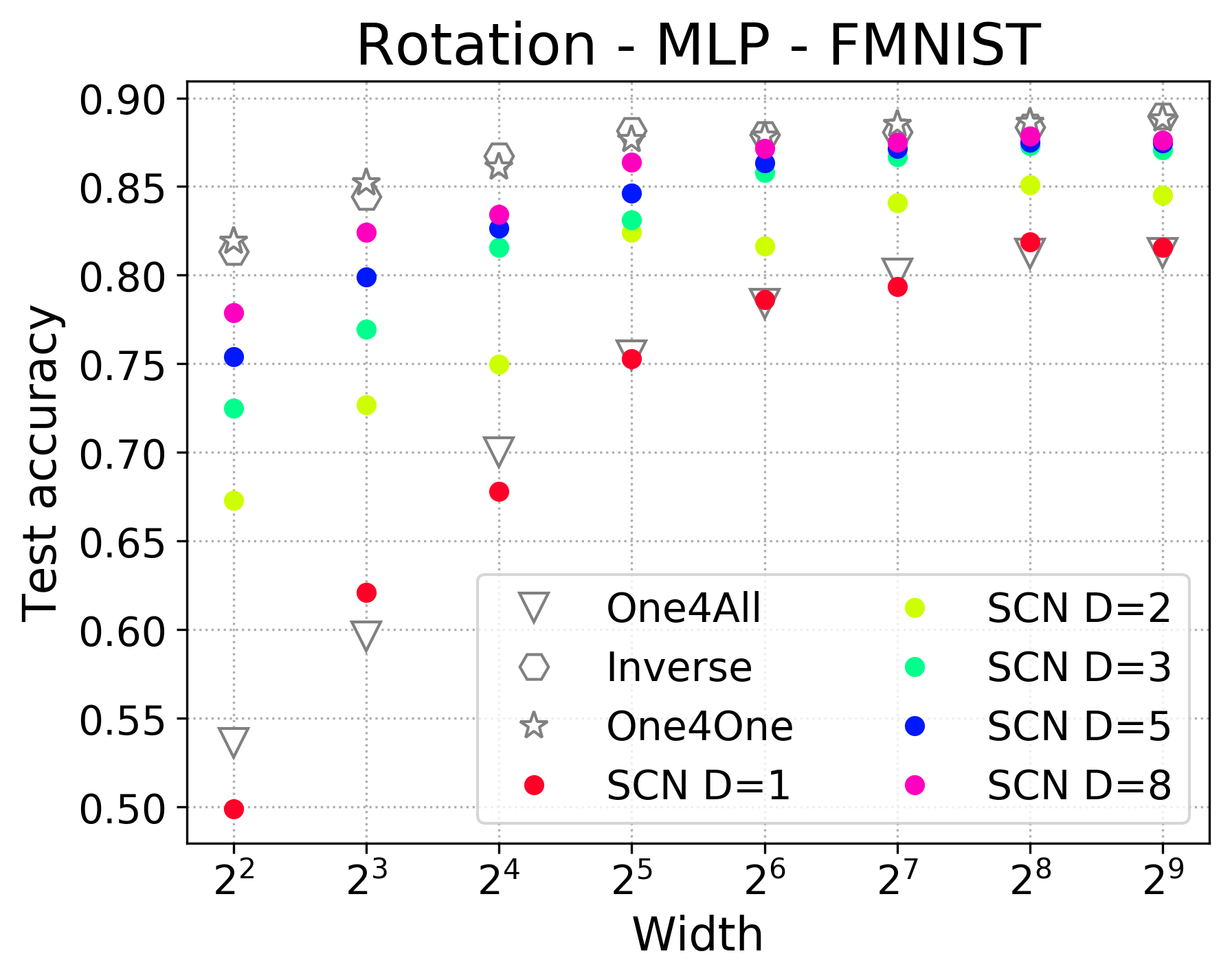}
     \includegraphics[width=.24\linewidth]
     {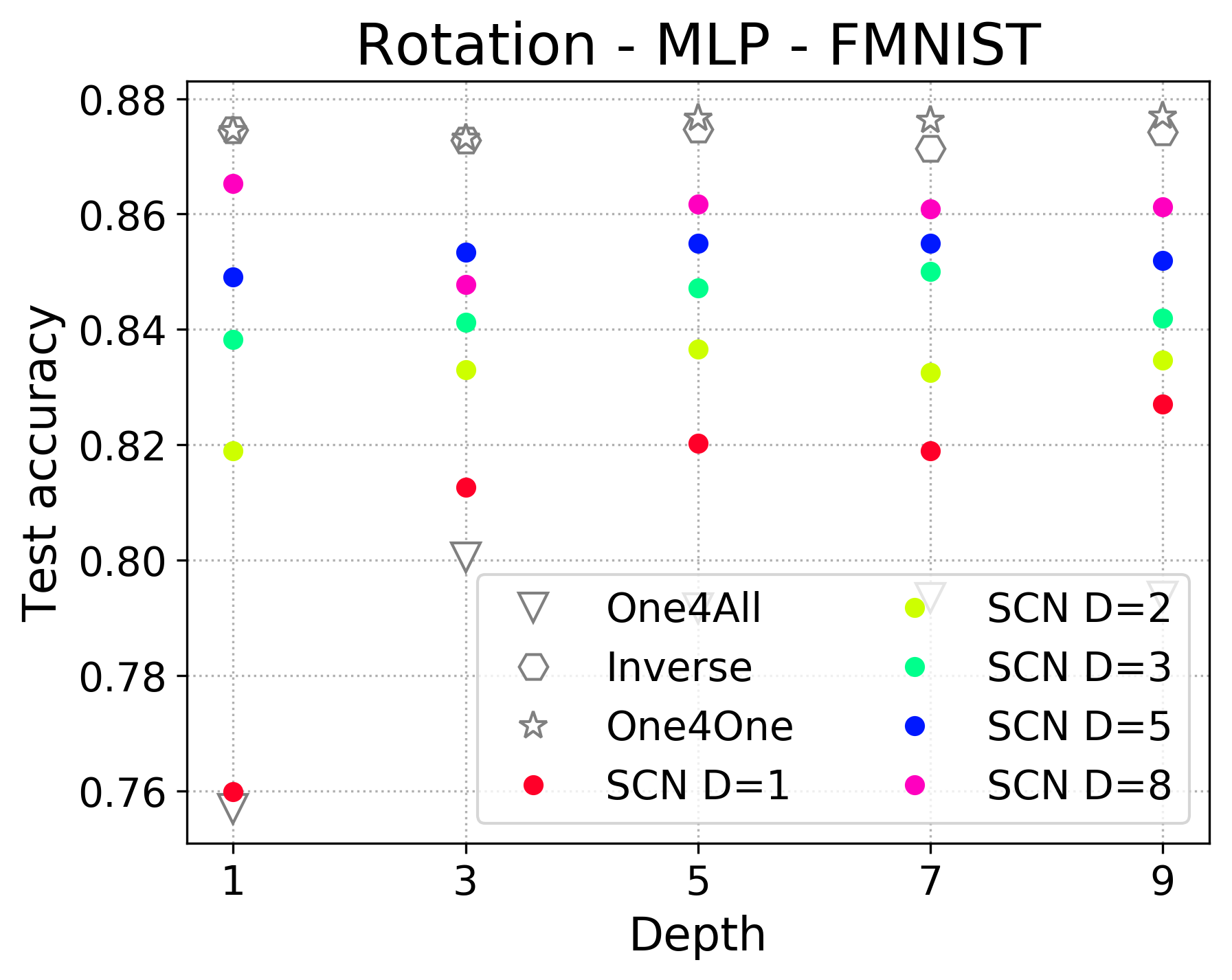}
     \includegraphics[width=.24\linewidth]
     {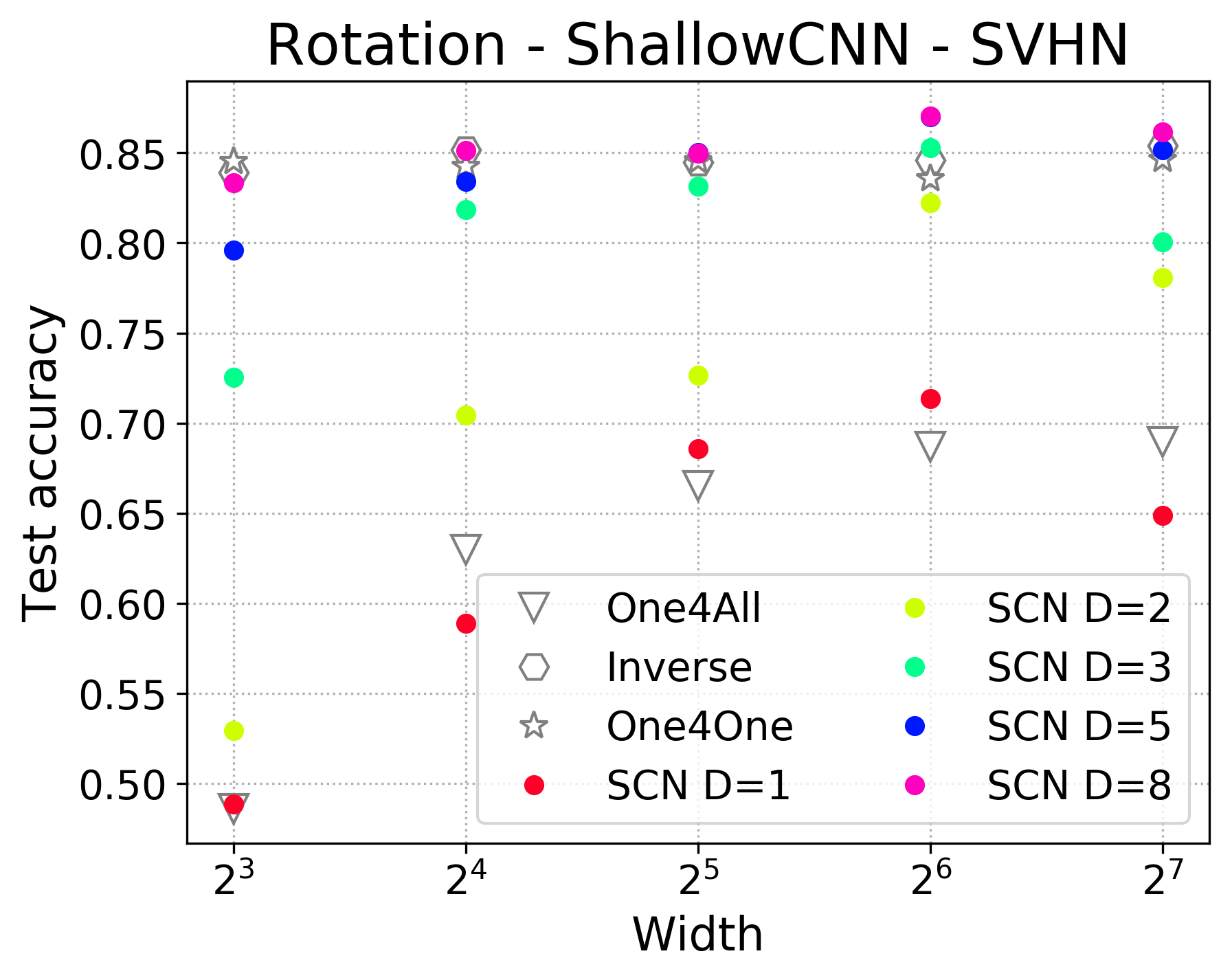}
     \includegraphics[width=.24\linewidth]
     {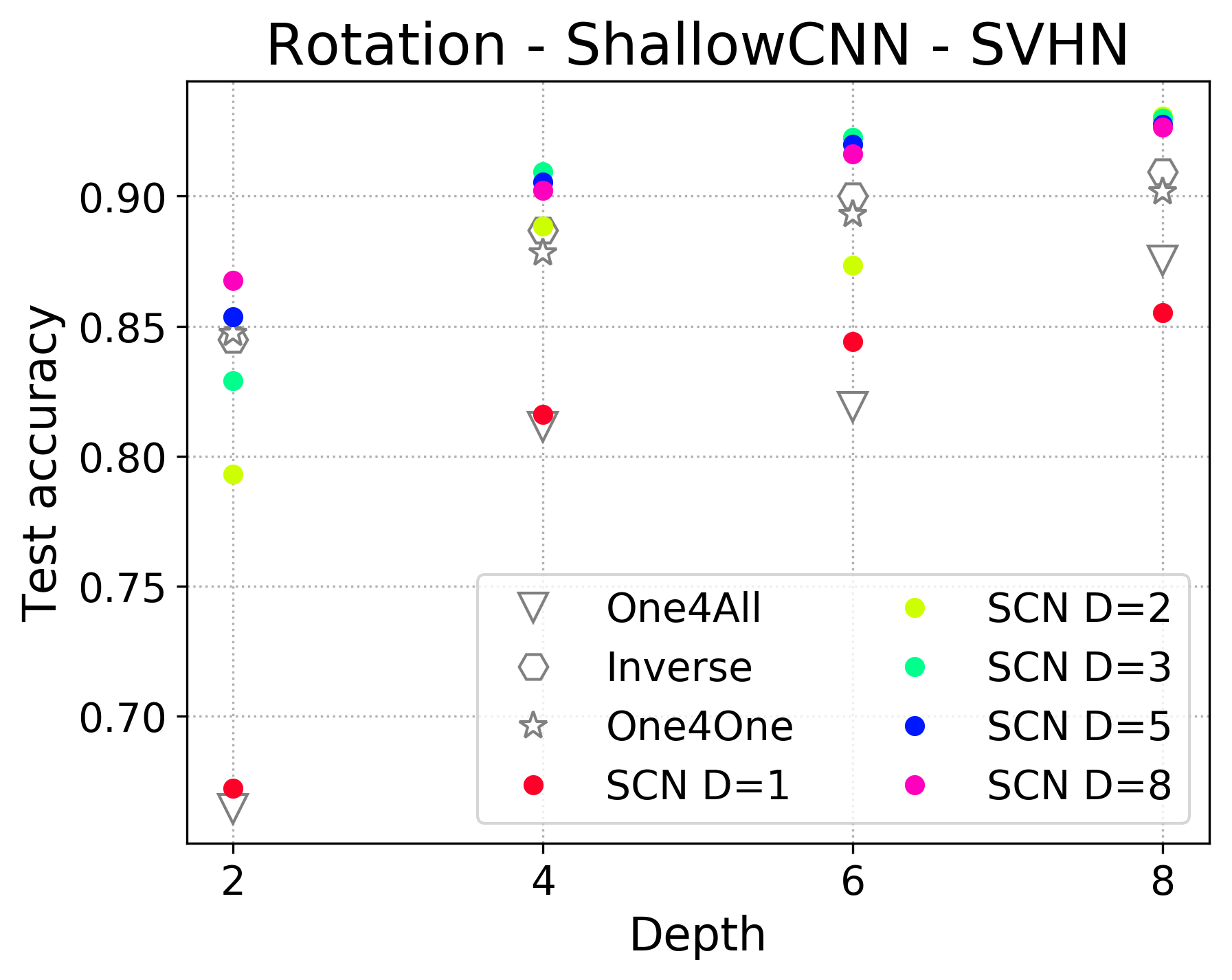}
    \caption{\textbf{Effect of network capacity on SCN test accuracy for 2D rotation}. We vary inference network width and depth to obtain the models of different capacity. \textbf{2 plots on the left:} Effect of width and depth for MLPs on FMNIST. In the width experiments, all networks are 1-layer MLPs. In the depth experiments, network width is fixed to 32 hidden units. \textbf{2 plots on the right:} Effect of width and depth for ShallowCNNs on SVHN. In the width experiments, the depth is fixed to two layers scaled together. In the depth experiments, the width of the hidden layers is fixed to 32 channels.}
    \label{fig:effect:capacity}
\end{figure*}

\subsection{SCN dimensionality and capacity constraints}
SCNs yield high performance already for low $D$, and we observe diminishing returns from adding dimensions for all tested transformations and dataset-architecture pairs, see \Figref{fig:effect:d}. SCN dimensionality $D$ impacts the overhead of training SCN, including the weights of the configuration network to compute $\beta$s and the weights $\theta_i$ of the base models. It also affects the overhead of reconfiguring an inference model if the transformation parameters $\alpha$ change, \eg to adapt an object detection model to a new camera position. Our empirical evaluation suggests that small $D$ is sufficient to achieve high SCN performance. 

The optimal $D$ depends on the inference network architecture and capacity. These trade-offs are explored when scaling inference network architectures along the width and depth dimensions in \Figref{fig:effect:capacity}. We present the capacity scaling results for the 2D rotation transformation for MLPs on FMNIST, and for ShallowCNNs on the SVHN dataset. Both architectures incorporate BatchNorm~\citep{ioffe2015batch} layers in their deeper versions to facilitate training. Although increasing width proves to be more effective for all baselines for MLPs on FMNIST, higher depth leads to better test accuracy for ShallowCNNs on SVHN. Even for small values of $D$, SCNs consistently deliver performance improvements that are on par with the high accuracy achieved by models trained with specific parameter settings. We note that as capacity constraints get increasingly relaxed, One4All models can quickly improve performance approaching One4One and Inverse.

\subsection{3D rotation-and-projection transformation}
\label{sec:3d}
\begin{figure*}[t]
    \centering
    \includegraphics[width=\linewidth]{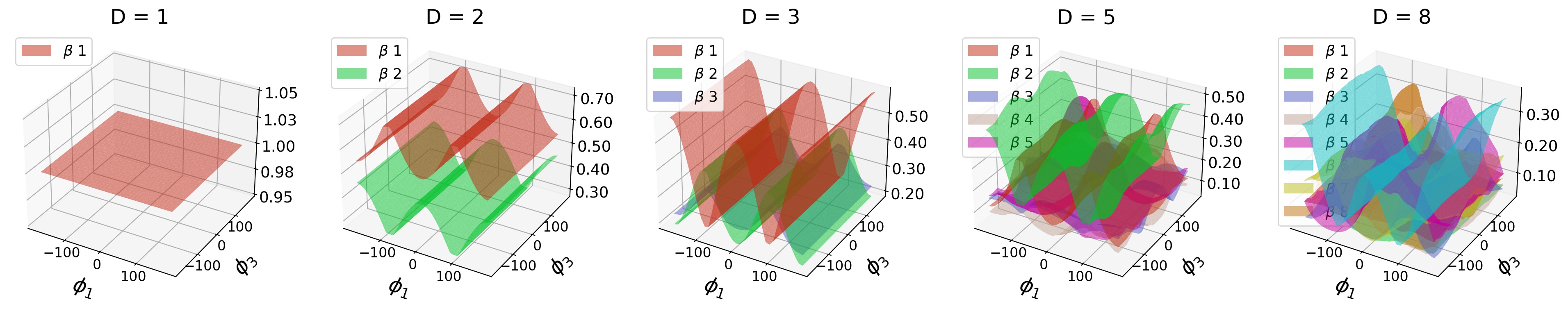}
    \caption{\textbf{A typical view of the SCN $\beta$-space for 3D rotation on LeNet5--ModelNet10}. Transformation parameters are a vector of ordered Euler angles $(\phi_1, \phi_2, \phi_3)$, each taking values from $(-\pi, \pi)$. We show the learned $\beta$-space for $\phi_{2} = -\pi$ with $D=1..8$. Further views can be found in Appendix~\ref{sec:3d:appendix}.
    An interactive visualiation is avaliable\protect\footnotemark. 
    The structure follows typical sine and cosine curves along multiple dimensions.}
    \label{fig:3d:beta}
\end{figure*}

\footnotetext{\label{3dviz}\url{https://subspace-configurable-networks.pages.dev/}}

\begin{figure*}[t]
    \centering
    \includegraphics[width=0.49\linewidth]{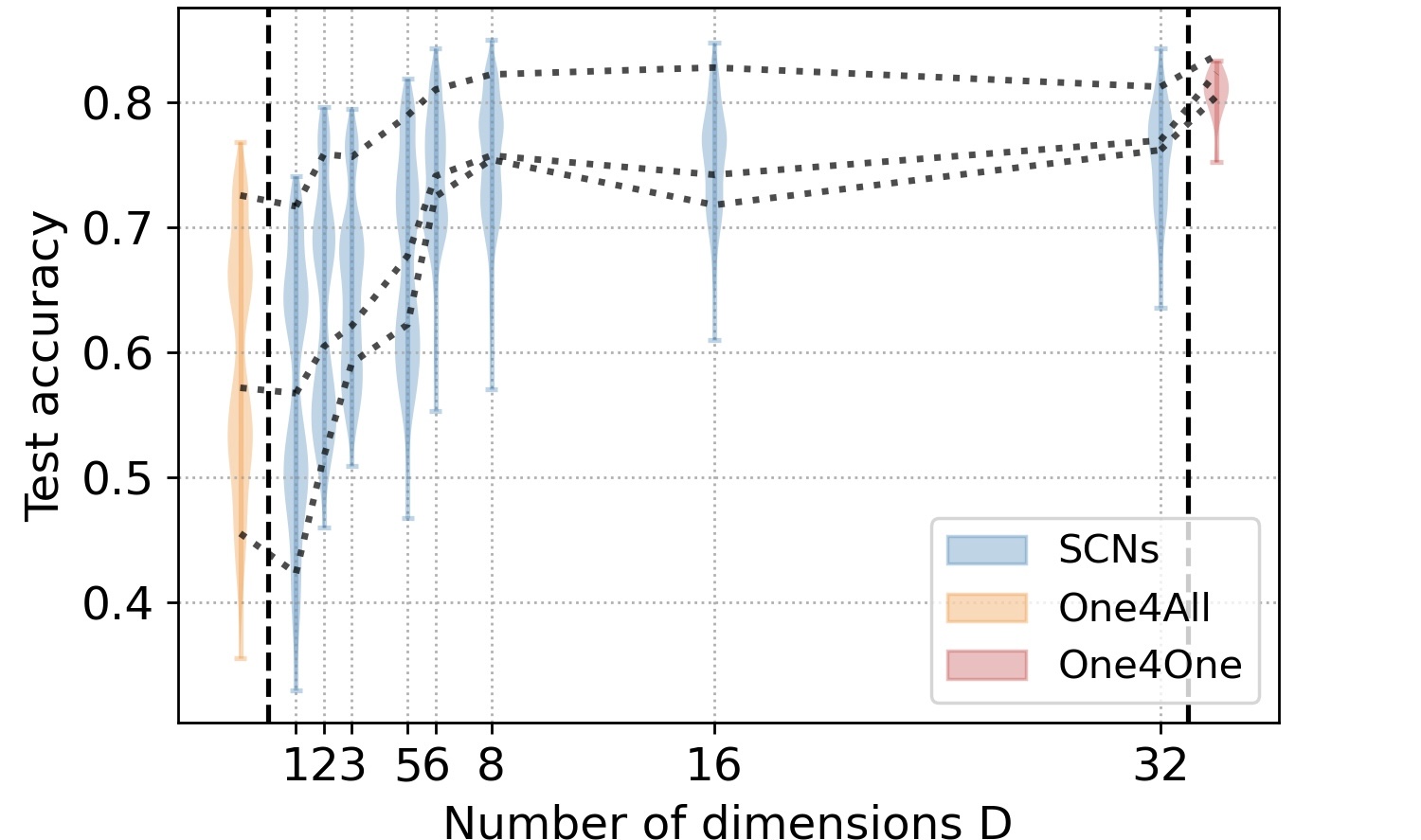}
    \caption{\textbf{Impact of $D$ on the SCN's test accuracy for 3D rotation}. 
    Each line is associated with a specific test angle and connects accuracies tested on the same rotation. %$(\phi_1, \phi_2, \phi_3)$. 
    Some rotations of a 3D object lead to a suboptimal view of the object and may significantly hurt classification accuracy.
    With increasing $D$, SCNs outperform One4All and approach the One4One baseline.}
    \label{fig:3d:accuracy:summary}
\end{figure*}

We evaluate SCNs on 3D rotations that present complex transformations with multiple suboptimal views of the object that hurt classification accuracy. We sample a point cloud from a 3D object representation part of the MobileNet10 dataset, rotate it using a vector of Euler angles $(\phi_1, \phi_2, \phi_3)$, and then project the point cloud to a 2D plane. The projection is then used as input to a classifier. We use LeNet5 as a backbone for the inference network to train SCNs. \Figref{fig:3d:beta} presents the view of the $\beta$-space as a function of two rotation angles $\phi_1$ and $\phi_3$, while $\phi_2$ is fixed at $-\pi$. The configuration space nicely reflects the characteristics of the input $\alpha$, provided as $\sin(\cdot)$ and $\cos(\cdot)$ of the input angles. 

Learned $\beta$s are insensitive to changes of $\phi_3$. Here $\phi_3$ corresponds to object rotations in the plane that does not change the object's visibility and thus leads to stable classification predictions, similarly to the 2D rotation transformation of a flat image.
The effect is best visible for low $D$ and can be verified using the interactive visualization we provide and by
inspecting further graphics in Appendix~\ref{sec:3d:appendix}.
% Note that no $\beta_i$ degenerates into a horizontal plane for $D>1$.

\Figref{fig:3d:accuracy:summary} compares SCN to One4All and One4One baselines. The Inverse is not feasible due to a projection of the point cloud on the 2D plane. Each violin comprises the model test accuracy evaluated on 30 randomly chosen angles. By comparing the accuracy for the same rotation angle (dotted lines in the plot), we observe a positive correlation between $D$ and SCN test accuracy. The result is similar to the SCN performance on 2D transformations.

\subsection{Search in the $\alpha$-space and I-SCNs}
\label{sec:iscn:search}

\begin{figure*}[t]
    \centering
    \includegraphics[width=.32\linewidth]{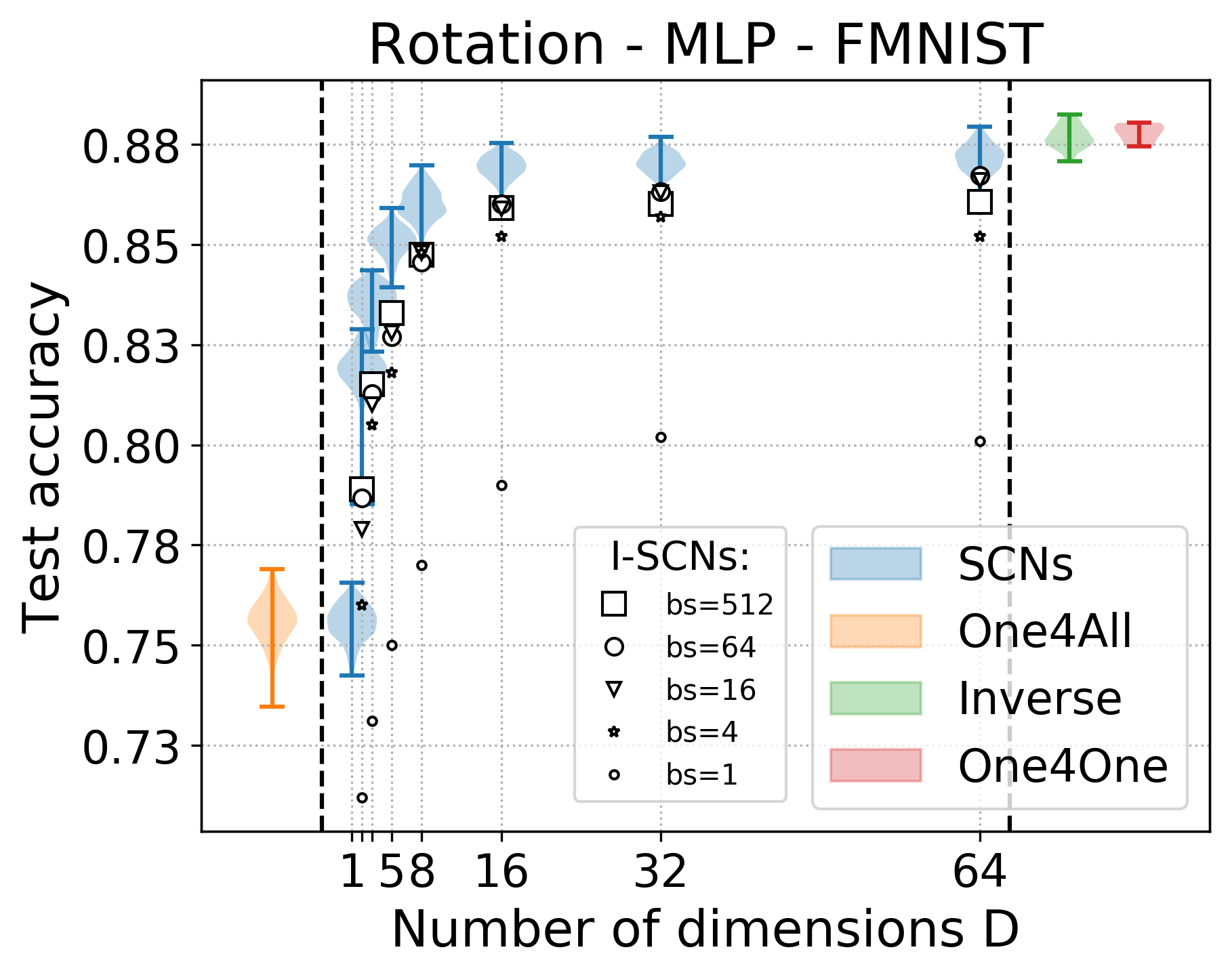}
    \includegraphics[width=.32\linewidth]{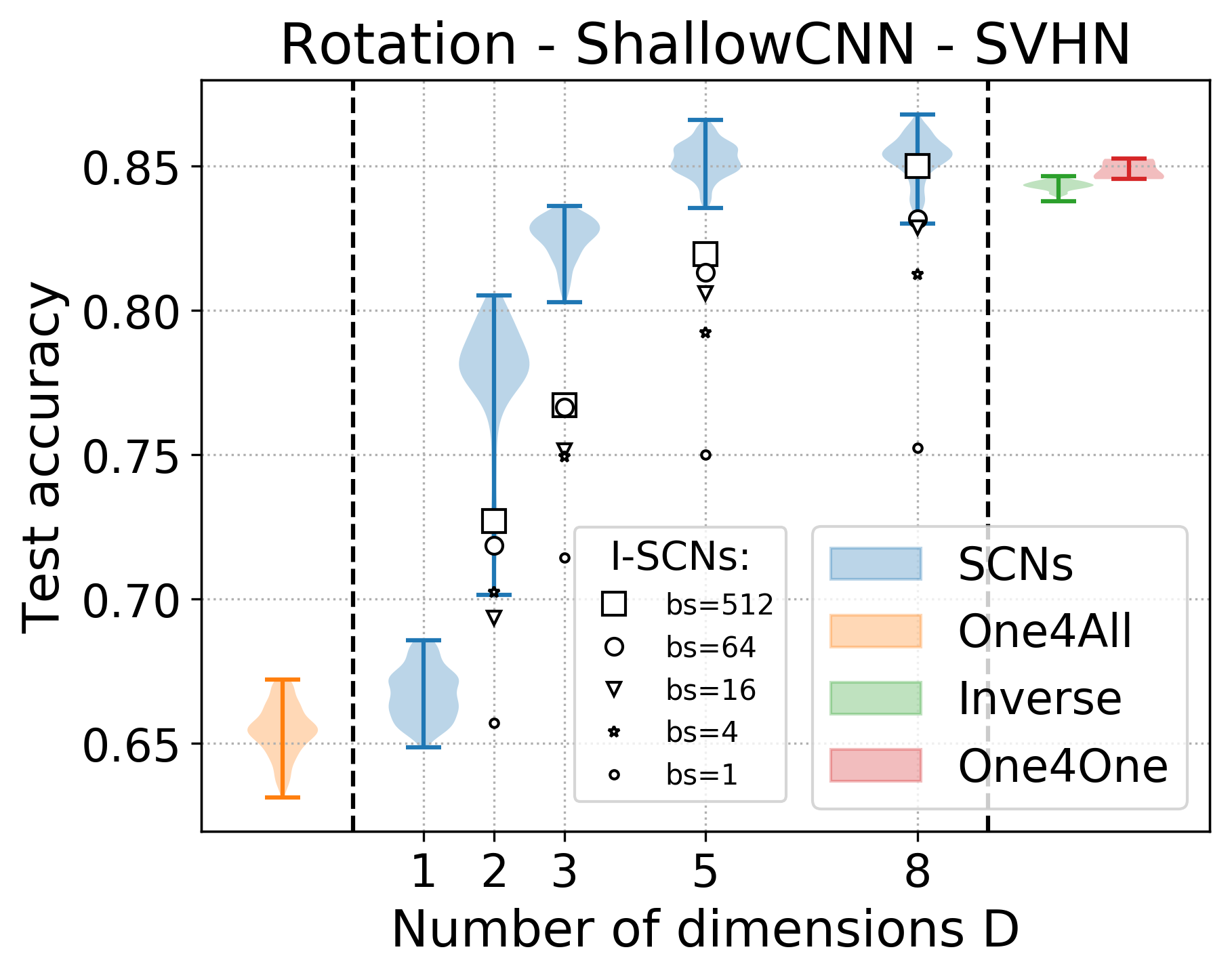}
    \includegraphics[width=.32\linewidth]{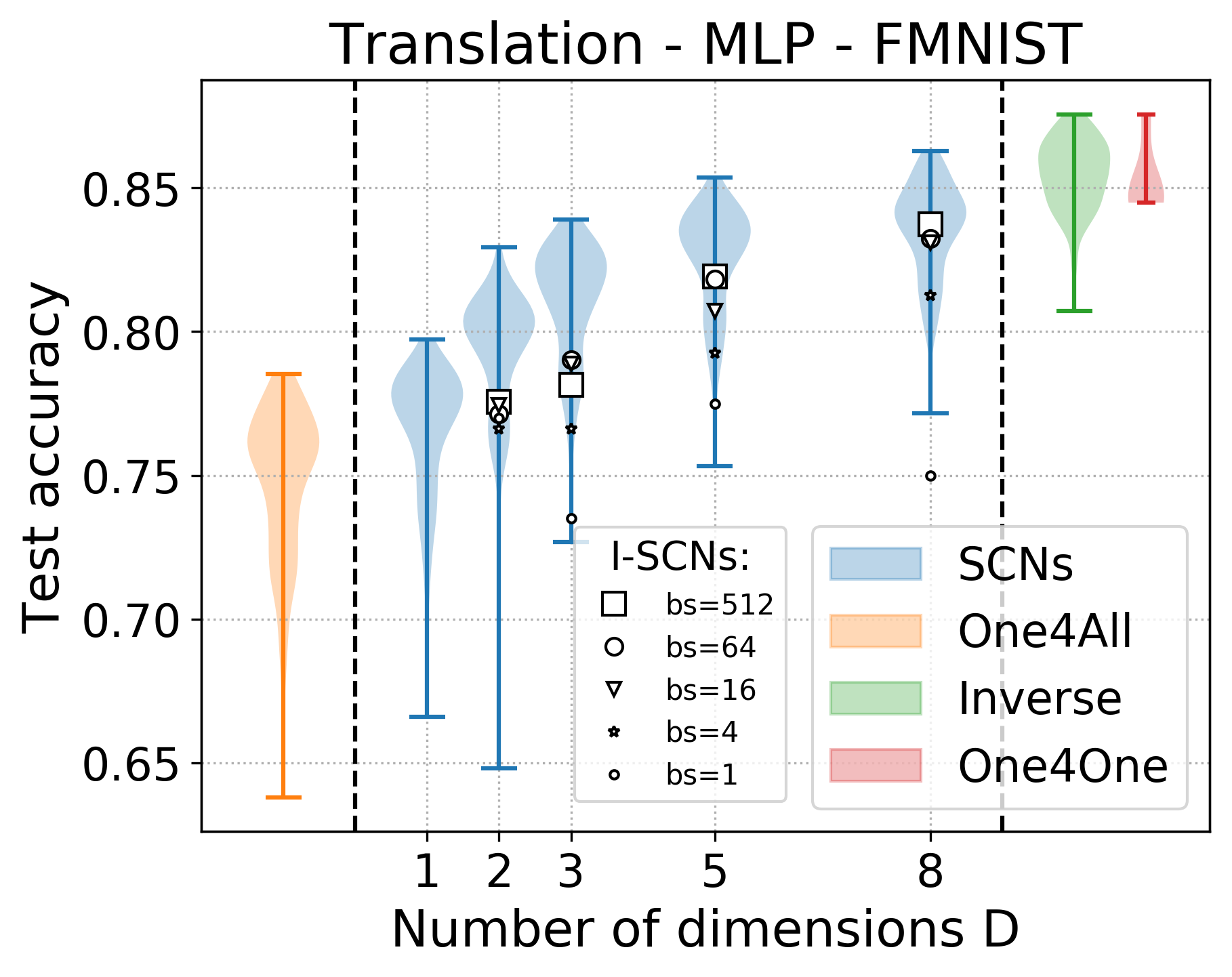}
    \caption{    
    \textbf{Performance of the search algorithm in the $\alpha$-space.} We enhance SCN evaluation plots in \Figref{fig:firstfigure} right and \Figref{fig:effect:d} left and middle right with the performance of the presented search algorithm in the $\alpha$-space. For higher batch sizes ($\geq$4) the search algorithm performs close to the respective SCNs with known $\alpha$.}
    \label{fig:iscn:search}
\end{figure*}

In \Figref{fig:iscn:search}, we enhance three plots from \Figref{fig:firstfigure} and \Figref{fig:effect:d} to show the performance of the search in the $\alpha$-space. Note that the proposed input-based search algorithm allows constructing invariant SCNs, which we refer to as I-SCNs. We compare to the test accuracy achieved by the respective SCNs with known and correct input $\alpha$ to I-SCNs. The search algorithm operates on batches (\texttt{bs} = batch size). Batch size $\geq$4 allows for an accurate estimation of $\alpha$ from the input data and yields high I-SCN performance.

\begin{table}[t]
    \caption{\textbf{Comparison of SCN to rotation-invariant TI-Pooling network.} With $D$=16 SCN is more parameter-efficient and yields higher accuracy than the baseline.}
    \centering
    \begin{tabular}{l|c|r}
    \toprule
         \textbf{Model} & \textbf{Test accuracy [\%]} & \textbf{\#parameters} \\
    \midrule
         TI-Pooling                  & 88.03 & 13'308'170 \\
         SCN(D=4) [\textbf{ours}]   & 87.37 & 374'582 \\
         SCN(D=8) [\textbf{ours}]   & 88.04 & 674'146 \\
         SCN(D=16) [\textbf{ours}]  & 88.42 & 1'273'274 \\
    \bottomrule
    \end{tabular}    
    \label{tab:compare:invariant}
\end{table}

Network architectures can be designed to be invariant to transformations. For example, to achieve rotation invariance in 2D and 3D, an element-wise maxpooling layer can be utilized~\citep{LaptevSBP16, su15mvcnn, savva2016}. TI-Pooling (called Transformation-Invariant Pooling) model~\citep{LaptevSBP16} employs parallel Siamese network layers with shared weights. We compare SCN and TI-Pooling models trained on 2D rotations with $\phi$ in the range $(0,\pi)$ on the FMNIST dataset. For SCNs, the inference network architecture is a 3-layer MLP with 64 hidden units in each layer. Tab.~\ref{tab:compare:invariant} shows the average classification accuracy and the number of parameters. SCN with $D$=16 dimensions demonstrates greater parameter efficiency compared to TI-Pooling, while also achieving higher accuracy than the baseline model.

\subsection{SCNs on Low-resource Devices}
\label{sec:eval:iot:fruit}

\begin{figure*}[!t]
    \centering
        \includegraphics[width=0.49\linewidth]{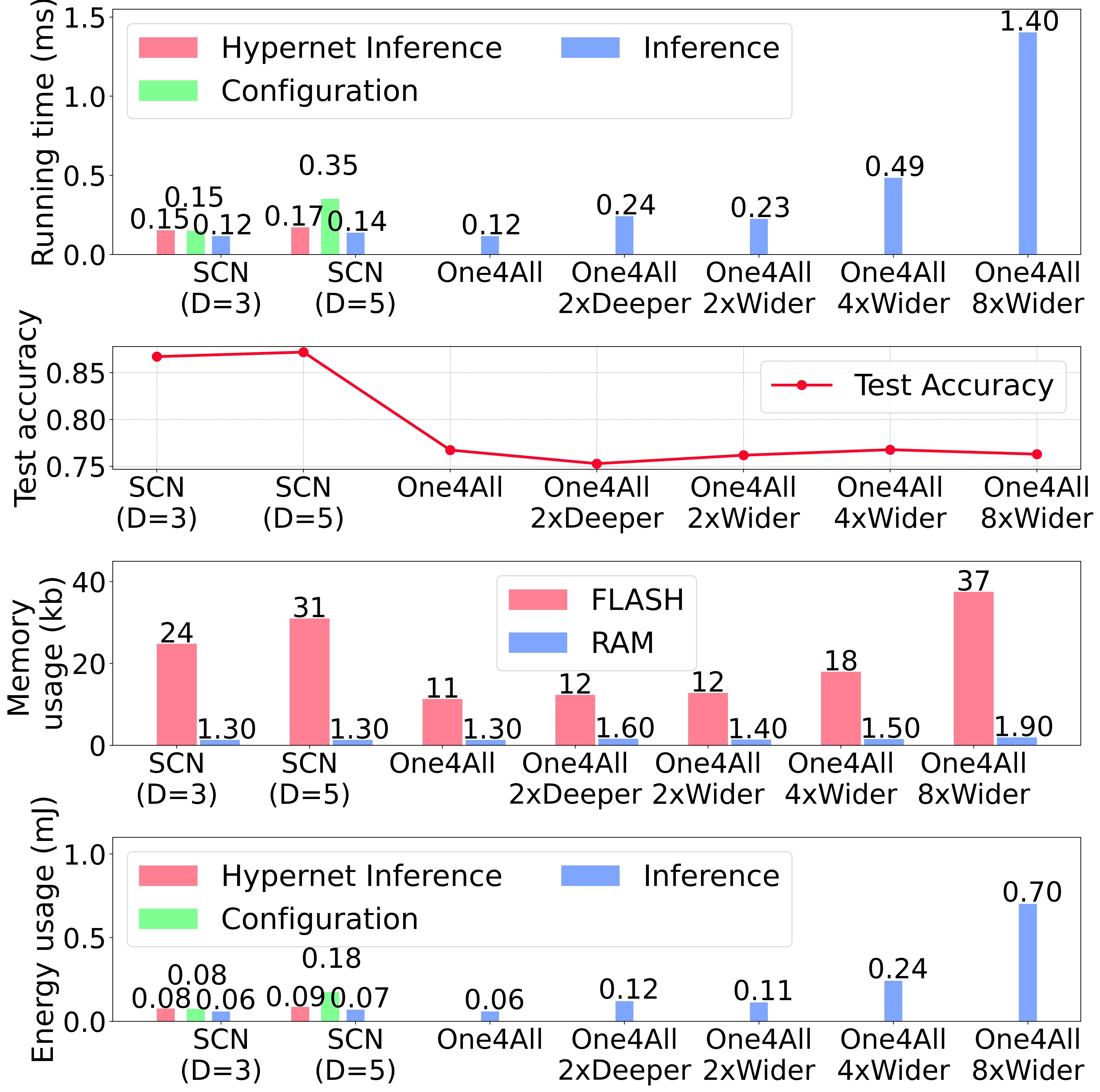}
    \caption{\textbf{SCN performance on the fruit classification task on Arduino Nano 33 BLE Sense}. From top to bottom the plots show: (1) Inference time in milliseconds, showcasing the efficiency of SCN, where the deeper and wider One4All variants lead to increased inference times. For SCNs, we also measure the execution latencies of the configuration network used to obtain vector $\beta$ ("Hypernet Inference"), and the computation time for generating $\theta$ from base models $\theta_i$ ("Configuration"). These latencies are only incurred when the deployment environment changes. (2) Test accuracy across various architectures, highlighting SCN's highly competitive performance. (3) RAM and flash memory usage in kB, indicating the increased resource consumption as the One4All model expands. (4) Energy 
    consumption in mJ.}
    \label{fig:iot:fruit:performance}
\end{figure*}

Embedded devices are frequently used in IoT applications on the edge. These devices feature severe resource constraints while running machine learning models, which should be robust to input transformations. We evaluate SCNs on two IoT applications: fruit classification using RGB sensor shown below, and traffic sign classification from camera images in Appendix~\ref{sec:eval:iot}.
We use Arduino Nano 33 BLE Sense~\citep{nano33blesense} to evaluate SCN performance on the fruit classification task. On-device APDS-9960 sensor is used to gather RGB and light intensity data, targeting bananas, apples, oranges, lemons, and kiwis under fluctuating natural light conditions.
\Figref{fig:iot:fruit:performance} shows the performance of SCN ($D=3$) and SCN ($D=5$) on classifying fruits using RGB data parameterized by the light intensity transformation. The One4All model used in this experiment has the same architecture as the SCN base model and features a LeNet5-like layout.
SCN reconfiguration overhead includes computation of parameters ("Hypernet Inference") and re-computation of inference model parameters ("Configuration"). The base model weights are stored in the on-board flash. Notice, that the wider and deeper One4All variants lead to a higher resource consumption than SCNs, yet they perform considerably worse on the fruit classification task due to color ambiguity under varying light conditions.

\section{Conclusion, Limitations, and Future Work}
\label{sec:conclusion}
This paper addresses the problem of model reconfiguration and robustness to input transformations under severe resource constraints. We design subspace-configurable networks (SCNs) that learn low-dimensional configuration subspaces and draw optimal model weights from this structure. We achieve surprisingly high accuracies even for a low number of configuration dimensions and observe a simple and intuitive structure of the subspace of optimal solutions for all investigated input transformations. 
Our findings open up practical applications of SCNs summarized below.

\paragraph{Post-deployment adaptation.}
SCNs can be used for the post-deployment model adaptation on resource-constrained devices as an alternative to costly backpropagation. SCN-configured inference networks are compact and can easily be deployed on devices with limited memory and processing power, \eg in robotics applications, edge computing, or classical embedded systems. In Section~\ref{sec:eval:iot:fruit} and Appendix~\ref{sec:eval:iot} we evaluate SCNs on two off-the-shelf MCUs (Tensilica Xtensa 32-bit LX7 dual-core and nRF52840) within the context of two example IoT applications: fruit classification using RGB sensor, and traffic sign classification from camera images. SCNs achieve a remarkable $\times$2.4 RAM savings and $\times$7.6 faster inference time compared to the One4All baseline reporting the same or higher test set accuracy.

\paragraph{SCNs as invariant architectures.}
SCNs can be used to build invariant network architectures on top of them by replacing the supply of $\alpha$ parameter vector with a search algorithm operating in the $\alpha$-space. We leverage the fact that the correct input parameters $\alpha$ should produce a confident low-entropy classification result~\citep{Wortsman2020supsup,Hendrycks2016}. Note that SCNs without the $\alpha$-search algorithm are of interest in their own right, since various sensor modalities can serve as input parameter $\alpha$, \eg using IMU (Inertial Measurement Unit) sensor measurements to determine the current 2D rotation parameter $\alpha$.

\paragraph{Limitations and future work.}
One of the major difficulties of this work is to effectively train SCNs for high $D$. The effects of the learning rate and training schedule are significant. 
With carefully chosen hyperparameters, we were able to train SCNs with $D=32$ for 3D rotation using LeNet5 without any degenerated dimension. Although the current work mainly explores continuous transformations, the extension of SCNs to discrete transformations is imaginable, yet requires rethinking of the theoretical arguments to provide a better understanding of the SCN design choices, obtained performance and limitations. Also, the generalization ability of SCN on out-of-distribution data presents the direction of our future research. Furthermore, SCNs can achieve enhanced optimization when combined with parameter-efficient model update methods, which we plan to explore in our future research. Finally, SCNs require the knowledge of a correct $\alpha$ and may present an additional vector of input manipulation and adversarial attacks. This direction requires future research and should be carefully considered before SCNs can be safely deployed.

\section*{Acknowledgements}
We thank Mitchell Wortsman and Rahim Entezari for their insightful comments on the early draft of the manuscript. This work was partly funded by the Austrian Research Promotion Agency (FFG) and Pro2Future (STRATP II 4.1.4 E‐MINDS strategic project). The results presented in this paper were computed using computational resources of ETH Zurich and the HLR resources of the Zentralen Informatikdienstes of Graz University of Technology.

\clearpage
\bibliography{arxiv.bib}

\newpage
\appendix
\section*{Appendix}
\section{Theoretical Results}
\label{sec:theory}

For a continuous transformation $T(\alpha)$, we provide theoretical results that help to understand the structure of the $\beta$-space using continuity arguments of the optimal solution space.

\subsection{Continuity of $\alpha$-to-$\beta$ subspace mapping}
\label{sec:theorem:continuity}

To simplify the formulation of the theorems and proofs, we suppose that the set of admissible parameters $\theta \in \mathbb{T}$ is a bounded subspace of $\mathbb{R}^L$ and all optimal parameter vectors $\theta^*_\alpha$ are in the interior of $\mathbb{T}$. To support theoretical analysis of the continuity of configuration space $\beta$, given continuous transformation parameters $\alpha$, we first introduce the necessary definitions and assumptions.

\paragraph{Parameterized curves.}
We define parameterized curves $c : I \rightarrow \mathbb{R}^n$ in $n$-dimensional Euclidean space that are at least one time continuously differentiable, where $I$ is a non-empty interval of real numbers. $t \in I$ is parameterizing the curve where each $c(t)$ is a point on $c$. In the following, we suppose that $c(t)$ is a natural parameterization of curve $c$, \ie the arc length on the curve between $c(a)$ and $c(b)$ is $b-a$:
\[
    b-a = \int_a^b ||c'(x)||_2 \; dx.
\]
As the shortest curve between two points is a line, we also find that $b-a \geq ||c(b) - c(a)||_2$. For example, let us define a curve $\alpha(s) = (1 - \frac{s}{d}) \alpha^0 + \frac{s}{d} \alpha^1$ where $d = ||\alpha^1 - \alpha^0 ||_2$. Then we have
\[
    \int_a^b ||c'(s)||_2 \; ds = \int_a^b \frac{1}{d} \; ||\alpha^1 - \alpha^0||_2 \; ds = b-a.
\]

\paragraph{Assumptions.}
At first, we are interested in the relation between $\alpha$ and corresponding optimal parameter vectors $\theta^*_\alpha$, \ie parameter vectors that minimize $E(\theta, \alpha)$. In order to simplify the forthcoming discussions, we suppose that the set of admissible parameters $\theta \in \mathbb{T}$ is a bounded subspace of $\mathbb{R}^L$ and all optimal parameter vectors $\theta^*_\alpha$ are in the interior of $\mathbb{T}$. 
We assume that the loss function $E(\theta, \alpha)$ is differentiable w.r.t. to $\theta$ and $\alpha$, and that it satisfies the Lipschitz condition
\begin{equation}\label{eq:1}
    \begin{split}
        |E(\theta, \alpha_2) - E(\theta, \alpha_1)| &\leq K_\alpha ||\alpha_2 - \alpha_1 ||_2 %\\
        %|E(\theta_2, \alpha) - E(\theta_1, \alpha)| &\leq K_\theta ||\theta_2 - \theta_1 ||_2
    \end{split}
\end{equation}
for some finite constant $K_\alpha$, and for all $\alpha_1, \alpha_2 \in \mathbb{A}$ and $\theta \in \mathbb{T}$. 
%for some finite constant $K_\alpha$ and $K_\theta$, and for all $\alpha, \alpha_1, \alpha_2 \in \mathbb{A}$ and $\theta, \theta_1, \theta_2 \in \mathbb{T}$.  

We are given data transformation parameters $\alpha \in \mathbb{A}$ and a network parameter vector $\theta^*_\alpha$. A point $\theta(0) = \theta^*_\alpha$ is a local minimum of the loss function if there is no curve segment $\theta(t)$ with $t \in [0, \hat{t}]$ for some $\hat{t}>0$ where $E(\theta(t), \alpha) \leq E(\theta(0), \alpha)$ and $E(\theta(\hat{t}), \alpha) < E(\theta(0), \alpha)$. All curves with $t \in [0, \hat{t}]$ for some $\hat{t}$ where $E(\theta(t), \alpha) = E(\theta(0), \alpha)$ define a maximal connected subset of locally optimal parameter values. In principle, the loss landscape for a given $\alpha$ may contain many disconnected subsets with local minima, \ie there is no path with a constant loss function between the locally minimal subsets. 

The analysis of the loss-landscape of (over-parameterized networks) is still an active area of research, see for example \citep{he2020recent, kawaguchi2019effect,liu2022spurious}. It turns out that in the case of over-parameterized networks, typical optimization methods like SGD do not get stuck in local minima when they exist, see for example \citep{allen2019convergence,kawaguchi2021recipe}. Therefore, it is reasonable to assume that all local minima found by the optimizer are also global, \ie for any given $\alpha \in \mathbb{A}$ the values of the loss functions $E(\theta^*_\alpha, \alpha)$ for all local minima $\theta^*_\alpha$ are equal.

\paragraph{Relation between data and network transformation.}
Loosely speaking, the following theorem shows that for any continuous curve that connects two transformation parameters in $\mathbb{A}$ there exists a corresponding continuous curve in the network parameter space $\mathbb{T}$. These two curves completely map onto each other where the network parameters are optimal for the corresponding data transformations. In particular, the curve in the network parameter space $\mathbb{T}$ has no jumps as is continuous. 

\begin{theorem} \label{th:1}
Suppose that the loss function $E(\theta, \alpha)$ satisfies (\ref{eq:1}), then the following holds: For any continuous curve $\alpha(s) \in \mathbb{A}$ with $0 \leq s \leq \hat{s}$ in the parameter space of data transformations there exists a corresponding curve $\theta(t) \in \mathbb{T}$ with $0 \leq t \leq \hat{t}$ in the parameter space of network parameters and a relation $(s, t) \in R$ such that 
\begin{itemize}
\item 
    the domain and range of $R$ are the intervals $s \in [0, \hat{s}]$ and $t \in [0, \hat{t}]$, respectively, and
\item 
    the relation $R$ is monotone, \ie if $(s_1, t_1), (s_2, t_s) \in R$ then $(s_1 \geq s_2) \Rightarrow (t_1 \geq t_2)$, and 
\item
    for every $(s, t) \in R$ the network parameter vector $\theta(t)$ minimizes the loss function $E(\theta, \alpha)$ for the data transformation parameter $\alpha(s)$.
\end{itemize} 
\end{theorem}

\begin{proof}
At first let us define a connected region $\mathbb{B}(\theta^*_\alpha, \delta, \alpha)$ of $\delta$-minimal loss functions values for given transformation parameters $\alpha$ and corresponding locally optimal network parameters $\theta^*_\alpha$, where $\theta^*_\alpha \in \mathbb{B}(\theta^*_\alpha, \delta, \alpha)$ and 
\begin{equation}\label{eq:2}
    \mathbb{B}(\theta^*_\alpha, \delta, \alpha) = \left\{ \theta \; | \; E(\theta, \alpha) - E(\theta^*_\alpha, \alpha) \leq \delta \right\}.
\end{equation}
In other words, within the connected region $\mathbb{B}(\theta^*_\alpha, \delta, \alpha)$ that contains $\theta^*_\alpha$, the loss function is at most $\delta$ larger than the optimal loss $E(\theta^*_\alpha, \alpha)$. Note that $\left\{ \theta \; | \; E(\theta, \alpha) - E(\theta^*_\alpha, \alpha) \leq \delta \right\}$ may be the union of many connected regions, but $\mathbb{B}(\theta^*_\alpha, \delta, \alpha)$ is the unique connected region that contains $\theta^*_\alpha$.

In order to prove the theorem, we show first that a small change in the data transformation from $\alpha(s)$ to $\alpha(s+\epsilon)$ leads to a new optimal network parameter vector $\theta^*_{\alpha(s + \epsilon)}$ that is within $\mathbb{B}(\theta^*_\alpha, \delta, \alpha)$, and $\delta$ decreases with the amount of change in $\alpha$. More precisely, we show the following statement: Given transformation parameters $\alpha(s)$, corresponding optimal network parameters $\theta^*_{\alpha(s)}$, and neighboring transformation parameters $\alpha(s+\epsilon)$ with a distance $\epsilon$ on the curve. Then the new optimal network parameter vector $\theta^*_{\alpha(s + \epsilon)}$ corresponding to $\alpha(s+\epsilon)$ is within the $\delta$-minimal region of $\theta^*_{\alpha(s)}$, namely 
\begin{equation}\label{eq:4}
    \theta^*_{\alpha(s + \epsilon)} \in \mathbb{B}(\theta^*_{\alpha(s)}, \delta, \alpha(s))
\end{equation}
if $\delta > 2 \epsilon K_\alpha$. Therefore, for an infinitesimally distance $\epsilon$ on the $\alpha$-curve, the new optimal network parameter vector $\theta^*_{\alpha(s + \epsilon)}$ is within the $\delta$-minimal region around $\theta^*_{\alpha(s)}$. Furthermore, there exists a curve segment between $\theta^*_{\alpha(s)}$ and $\theta^*_{\alpha(s + \epsilon)}$ where every point $\theta$ on this curve satisfies $E(\theta, \alpha(s)) - E(\theta^*_{\alpha(s)}, \alpha(s)) \leq \delta$ according to (\ref{eq:2}), \ie its loss for $\alpha(s)$ is at most $\delta$ higher than the loss at the beginning of the curve segment. Such a curve segment always exists as $\mathbb{B}$ is a connected region and the curve segment can completely be within $\mathbb{B}$. The change in loss $\delta$ for $\alpha(s)$ on the curve segment decreases with $\epsilon$ and is infinitesimally small.

We now prove the above statement. From (\ref{eq:1}) we find 
\begin{equation}\label{eq:3}
    |E(\theta, \alpha(s + \epsilon)) - E(\theta, \alpha(s))| \leq K_\alpha ||\alpha(s + \epsilon) - \alpha(s) ||_2 \leq \epsilon K_\alpha.
\end{equation}
First, we show that the minimum for $\alpha(s + \epsilon)$ is within the region $\mathbb{B}(\theta^*_{\alpha(s)}, \delta, \alpha(s))$. At the border of the region we find $E(\theta, \alpha(s + \epsilon)) \geq E(\theta^*_{\alpha(s)}, \alpha(s)) + \delta$ due to (\ref{eq:2}). Combining this with (\ref{eq:3}) yields $E(\theta, \alpha(s + \epsilon)) \geq E(\theta^*_{\alpha(s)}, \alpha(s)) + \delta - \epsilon K_\alpha$. In the interior of the region we find as the best bound $E(\theta, \alpha(s + \epsilon)) \leq E(\theta^*_{\alpha(s)}, \alpha(s)) + \epsilon K_\alpha$ using (\ref{eq:3}). If the loss for $\alpha(s + \epsilon)$ is larger at the border of the region than in its interior, we know that a locally minimal loss is within the region, \ie (\ref{eq:4}) holds. Therefore, we require $E(\theta^*_{\alpha(s)}, \alpha(s)) + \epsilon K_\alpha < E(\theta^*_{\alpha(s)}, \alpha(s)) + \delta - \epsilon K_\alpha$ and therefore, $\delta > 2 \epsilon K_\alpha$. 

Using the above statement, see (\ref{eq:4}), we start from some curve $\alpha(s)$, $0 \leq s \leq \hat{s}$ and construct a corresponding optimal curve in the network parameter space $\theta(t)$ for $0 \leq t \leq \hat{t}$. We begin with some $\alpha(s)$ and an optimal network $\theta^*_{\alpha(s)} \in \argmin E(\theta, \alpha(s))$. We know that the optimal parameter vector $\theta^*_{\alpha(s + \epsilon)}$ for infinitesimally close transformation parameters  $\alpha(s + \epsilon)$ on the curve $\alpha(s)$ is within the $\delta$-minimal region around $\theta^*_{\alpha(s)}$. Therefore, to a small segment from $\alpha(s)$ to $\alpha(s + \epsilon)$ we assign a finite segment from $\theta^*_{\alpha(s)}$ to $\theta^*_{\alpha(s + \epsilon)}$ completely within the $\delta$-minimal region around $\theta^*_{\alpha(s)}$. Every point on this curve segment corresponds to a network whose loss is either infinitesimally close to the optimal values $E(\theta^*_{\alpha(s)}, \alpha(s))$ or $E(\theta^*_{\alpha(s+\epsilon)}, \alpha(s +\epsilon))$. In other words, the curve segment starts from optimal network parameters $\theta^*_{\alpha(s)}$, ends at optimal network parameters $\theta^*_{\alpha(s + \epsilon)}$, and in between traverses the region with loss values that are infinitesimally close to either of these optimal loss values. This process is repeated, starting from $\alpha(s+ \epsilon)$ and $\theta^*_{\alpha(s + \epsilon)}$. As a result, the two curves $\alpha(s)$ and $\theta(t)$ are connected by a relation $(s, t) \in R$ such that the domains are the intervals of the curve parameters $[0, \hat{s}]$ and $[0, \hat{t}]$. If $(s, t) \in R$ then $\theta(t)$ is optimal for $\alpha(s)$. No points on the curves are missing, \ie without a relation to the other curve. Moreover, the relation $R$ is monotone: If $(s_1, t_1), (s_2, t_2) \in R$ then $(s_1 \geq s_2) \Rightarrow (t_1 \geq t_2)$.
\end{proof}

Note that the assumption that local minima are also global minima is crucial. For example, suppose that for a given $\alpha$ there are two local minima that are separated by a single barrier. Suppose further that by changing $\alpha$, just the height of the barrier reduces until it vanishes completely. At this value of $\alpha$, a small change, \ie a short distance on the curve $\alpha(s)$, leads to a large change in the optimal $\theta$. In other words, given a curve $\alpha(s)$ there may be no corresponding continuous curve $\theta(t)$ that satisfies the properties of the above theorem.

We are also interested in the relation between $\alpha$ and corresponding optimal vectors $\beta^*_\alpha$ that define optimal locations on the linear subspace of admissible network parameters as defined by (\ref{eq:5}). To simplify the discussion, we suppose that $\beta \in \mathbb{B}$ are in a bounded subspace of $\mathbb{R}^D$, and all basis vectors $\theta_j$ that define $f(\beta)$ in (\ref{eq:5}) have bounded elements. Under these assumptions, we can derive a corollary from Theorem~\ref{th:1}. 

\begin{corollary} \label{th:2}
Suppose that the loss function $E(\theta, \alpha)$ satisfies (\ref{eq:1}), and for any $\alpha \in \mathbb{A}$ all local minima of $E(f(\beta), \alpha)$ w.r.t. $\beta$ are global. Then the following holds: For any continuous curve $\alpha(s) \in \mathbb{A}$ with $0 \leq s \leq \hat{s}$ in the parameter space of data transformations there exists a corresponding curve $\beta(t) \in \mathbb{B}$ with $0 \leq t \leq \hat{t}$ on the linear network parameter subspace according to (\ref{eq:5}) and a relation $(s, t) \in R$ such that 
\begin{itemize}
\item 
    the domain and range of $R$ are the intervals $s \in [0, \hat{s}]$ and $t \in [0, \hat{t}]$, respectively, and
\item 
    the relation $R$ is monotone, \ie if $(s_1, t_1), (s_2, t_s) \in R$ then $(s_1 \geq s_2) \Rightarrow (t_1 \geq t_2)$, and 
\item
    for every $(s, t) \in R$ the network parameter vector $\beta(t)$ minimizes the loss function $E(f(\beta), \alpha)$ for the data transformation parameter $\alpha(s)$.
\end{itemize} 
\end{corollary}

\begin{proof}
Sketch: The relation between the parameterization $\beta$ of the linear subspace spanned by $\theta_i$ and the resulting network parameters $\theta = f(\beta)$ is given by (\ref{eq:5}). As $\beta \in \mathbb{B}$ is bounded and the basis vectors $\theta_j$ are finite, $f(\theta)$ is differentiable and Lipschitz constrained. Therefore, the proof as provided for Theorem~\ref{th:1} holds as well by just replacing $\theta = f(\beta)$. Therefore, the results of Theorem~\ref{th:1} hold for $\beta$ as well if for any $\alpha \in \mathbb{A}$ all local minima of $E(f(\beta), \alpha)$ w.r.t. $\beta$ are global.
\end{proof}

\subsection{Small parameter changes}
\label{sec:theorem:smallchanges}

This section shows that small changes to transform parameters $\alpha$ result in small changes of optimal configuration $\beta^*_\alpha$ for suitable loss functions $E(f(\beta), \alpha)$. For the forthcoming analysis, we suppose that $E(f(\beta), \alpha)$ is at least twice differentiable w.r.t. $\alpha$ and $\beta$.

\begin{theorem} \label{th:3}
Suppose that $\beta_0^* \in \mathbb{B}$ locally minimizes $E(f(\beta), \alpha)$ for $\alpha_0 \in \mathbb{A}$. Moreover, the Hessian $\nabla_{(\beta, \alpha)}^2 E(f(\beta),\alpha)$ of the loss function at $\alpha = \alpha_0$ and $\beta = \beta_0^*$ exists, and its submatrix $\nabla_{\beta}^2 E(f(\beta),\alpha)$ is non-singular. 

Then, if $||\alpha_1 - \alpha_0|| \leq \epsilon$ for some small $\epsilon$, then there exists a locally optimal $\beta_1^*$ for $\alpha_1$ such that 
\[
||\beta_1^* - \beta_0^*|| \leq ||(\nabla_{\beta}^2 E(f(\beta),\alpha))^{-1}|| \cdot ||\nabla_\alpha (\nabla_{\beta} E(f(\beta),\alpha)))|| \cdot \epsilon.
\]
\end{theorem}

% \begin{table*}[!htb]
%   \caption{\textbf{Training hyper-parameters for all architecture-dataset pairs.}}
%   \label{train_params}
%   \centering
%   \begin{threeparttable}
%   {\small
%   \begin{tabular}{lllllll}
%     \toprule
%     Hyper-param.    & MLP- & ShallowCNN- & ResNet18- & LeNet5- & M5- \\
%         & FMNIST & SVHN & CIFAR10 & ModelNet10 & SpeechCmds \\
%     \midrule
%     Optimizer       & Adam         & Adam      & Adam      & Adam      & Adam\\
%     LR              & 0.001        & 0.001     & 0.001     & 0.006     & 0.01\\
%     Weight decay    &              &           & 0.0001    &           &  \\
%     LR schedule  & CosineLR  & CosineLR & CosineAnnealing  & CosineAnnealing & CosineAnnealing \\
%       & & & WarmRestarts, & $T_{max}=6'000$, & $T_{max}=100$, \\
%       & & & $T_0=25$, $T_{mult}=25$  & $\eta_{min} = 5\cdot10^{-6}$ & $\eta_{min} = 0$  \\
%     Batch size       & 64               & 256           & 512    & 256      & 256 \\
%     Epochs           & 500             & 500          & 1'000    & 6'000    & 100 \\
%     % Momentum         & 0.9              & 0.9           & 0.9  \\
%     % Weight Decay     & -                & -             & 0.0001  \\
%     Data augment.& Normalization    & Normalization & HorizontalFlip & None & Resample to\\
%         &  &  & HorizontalFlip &  & 16 KHz\\
%     \bottomrule
%   \end{tabular}
%   }
%   \end{threeparttable}
%   \label{table:training hyper-parameters}
% \end{table*}

\begin{proof}
For a network trained to a minimum in $E(f(\beta), \alpha)$ for a given $\alpha$, its first derivative over weights equals to 0, \ie $\nabla_\beta E(f(\beta),\alpha) = 0$ for an optimal vector $\beta^*_\alpha$. We assume that this derivative exists and is abbreviated by a function $F(\beta, \alpha) = \nabla_\beta E(f(\beta),\alpha)$.

Let $F(\beta, \alpha)$ at $\beta = \beta^*_\alpha$ be a differentiable function of $\beta$ and $\alpha$. 
We apply the first-order Taylor expansion of $F$ at a point $(\beta_0^*, \alpha_0)$, \ie $\beta_0^*$ is optimal for $\alpha_0$:
\begin{equation}
    F(\beta, \alpha) = F(\beta_0^*, \alpha_0) + \nabla_{\beta^*} F\Big|_{(\beta_0^*,\alpha_0)} \delta \beta^* + \nabla_{\alpha} F\Big|_{(\beta_0^*,\alpha_0)} \delta \alpha,
\end{equation}
where $\delta \beta^* = \beta_1^* - \beta_0^*$ and  $\delta \alpha = \alpha_1 - \alpha_0$. We have $F(\beta^*, \alpha) = 0$ and $F(\beta_0^*, \alpha_0) = 0$ due to the optimality of the loss function. 

We abbreviate the evaluated partial derivatives $\nabla_{\beta^*} F\big|_{(\beta_0^*, \alpha_0)} := P$ and $\nabla_{\alpha} F\big|_{(\beta_0^*,\alpha_0)} := Q$. Since $\beta^*$ and $\alpha$ are vectors, we find that $P \in \mathbb{R}^{D \times D}$ and $Q \in \mathbb{R}^{S \times D}$ are matrices. Thus,
\begin{equation}
  P \cdot \delta \beta^* + Q \cdot \delta \alpha = 0
\end{equation} 
and therefore, 
\[
  \delta\beta^* = - P^{-1} Q \delta\alpha.
\]
Using basic results from linear algebra we find
\[
  ||\delta\beta^*|| \leq || P^{-1} || \cdot  ||Q || \cdot || \delta\alpha|| 
\]
and therefore
\[
  ||\beta_1^* - \beta_0^*|| \leq || P^{-1} || \cdot  ||Q || \cdot || \alpha_1 - \alpha_0|| \leq || P^{-1} || \cdot  ||Q || \cdot \epsilon.
\]

\noindent From the last equation it follows that small changes of transform parameters $\delta\alpha$ result in small changes $\delta\beta^*$ of the optimal solution $\beta_0^*$ if $P$ is invertible. Note that $P = \nabla_{\beta}^2 E(f(\beta),\alpha)$ and $Q = \nabla_\alpha (\nabla_{\beta} E(f(\beta),\alpha)))$ for $\alpha = \alpha_0$ and $\beta = \beta_0^*$.
\end{proof}
Note that the condition of the theorem is crucial, \ie the Hessian of the loss function with respect to the parameters $\beta$ of the linear subspace at the optimal solution is invertible. This excludes cases with saddle points, where there is no optimal point vector in the neighborhood after a small change in $\alpha$. Moreover, we can only make statements about local minima of the loss function due to the use of the Taylor expansion. 

%If $\delta\beta^*$ is in the Null-Space of $P$, then we may have a saddle point in this direction and there may be a lower loss function in this direction. If $Q \delta\alpha$ is not perpendicular to the Null-Space, the loss function for an $\alpha$ in the environment of $\alpha_0$ may have a minimum that is far away from $\theta_0^*$. It is not clear though, whether and when the condition of the theorem is satisfied.

%It may well be, that there is a lower loss function than $E(\theta_1^*, \alpha_1)$ in the whole subspace of weights $\theta$, i.e., there exists a $\theta_1^{**}$ with $E(\theta_1^{**}, \alpha_1) < E(\theta_1^{*}, \alpha_1)$ and large $||\theta_1^{*} - \theta_1^{**}||$. For example, if we change $\alpha$ in a certain direction, we first are still finding optimal $\theta^*$ in the environment of previous weights. After a certain point, the loss gets larger, but is still locally optimal. But at some distant weight, the loss is now lower. In other words, we may have separated optimality weight regions that are separated by weights that are not optimizing any $\alpha$. Again, I am not sure whether and how we can avoid this. A possible approach could be to use some Lipschitz bounds.

\section{Implementation details} \label{sec:implementationdetails}

% The source code of all experiments is available online.\footnote{\url{https://github.com/osaukh/subspace-configurable-networks.git}}
We trained over 1'000 models on a workstation featuring two NVIDIA GeForce RTX 2080 Ti GPUs to evaluate the performance of SCNs presented in this work. Training a model takes up to several hours and depends on the SCN dimensionality and model complexity. WandB\footnote{\url{https://wandb.ai}} was used to log hyper-parameters and output metrics from runs.

\subsection{Datasets and architectures}

\paragraph{Configuration network.}
Throughout all experiments we used the configuration network featuring one fully-connected layer comprising 64 neurons. Depending on the input (whether 2 values for 2D rotation and translation, 6 values for 3D rotation, and 1 value for all other considered transformations), the configuration network contains $64\cdot(input\_size+1) + 65\cdot D$ trainable parameters. Note that $D$ is the size the the configuration network's output. The architecture includes the bias term. For example, for 2D rotation with 3 outputs, the configuration network has 387 parameters.

We test the performance of SCNs on five dataset-architecture pairs described below. For MLPs and ShallowCNNs we vary architectures' width and depth to understand the impact of network capacity on efficiency of SCNs for different $D$. To scale up along the width dimension, we double the number of neurons in each hidden layer. When increasing depth, we increase the number of layers of the same width. To improve training efficiency for deeper networks (deeper than 3 layers), we use BatchNorm layers~\citep{ioffe2015batch} when scaling up MLPs and ShallowCNNs along the depth dimension. 
The number of parameters for the network architectures specified below (excluding BatchNorm parameters) is only for a single inference network $\mathcal{G}$. $D$ base models of this size are learned when training a SCN.

\paragraph{MLPs on FMNIST.} 
FMNIST~\citep{xiao2017fashionmnist} is the simplest dataset considered in this work. The dataset includes 60'000 images for training and 10'000 images for testing. The dataset is available under the MIT License.\footnote{\url{https://github.com/zalandoresearch/fashion-mnist}} We use MLPs of varying width $w$ and depth $l$ to evaluate the impact of the dense network capacity on SCNs. The number of parameters of the MLP inference network for 10 output classes scales as follows: 
\[
(32^2+1)\cdot w + (l-1)\cdot(w^2+w) + 10\cdot(w+1).
\]

\paragraph{ShallowCNNs on SVHN.} 
SVHN~\citep{SVHN} digit classification dataset contains 73'257 digits for training, 26'032 digits for testing, and 531'131 additionally less difficult digits for assisting training. No additional images are used. The dataset is available for non-commerical use.\footnote{\url{http://ufldl.stanford.edu/housenumbers/}} Shallow convolutions (ShallowCNNs) were introduced by \citet{neyshabur2020learning}. We scale the architecture along the width $w$ and depth $d$ dimensions. The number of parameters scales as follows: 
\[
(9\times 9\times 3 + 1)\cdot w + (l-1)\cdot(13\times 13 \times w + 1)\cdot w + 10\cdot(w+1). 
\]

\paragraph{LeNet5 on ModelNet10.} 
ModelNet10~\citep{3dShapeNets} is a subset of ModelNet40 comprising  a clean collection of 4,899 pre-aligned shapes of 3D CAD models for objects of 10 categories. We use this dataset to evaluate SCN performance on images of 3D rotated objects. We first rotate an object in the 3D space, subsample a point cloud from the rotated object, which is then projected to a fixed plane. The projection is then used as input to the inference network. Rotation parameters $\alpha$ are input to the trained hypernetwork to obtain the parameters in the $\beta$-space to construct an optimal inference network. We use LeNet-5~\citep{lecun1998} as inference network architecture with 138'562 parameters.

\paragraph{ResNet18 on CIFAR10.} 
This work adopts the ResNet18 implementation by \citet{he2015resnet18} with around 11 million trainable parameters. We use ResNet18 on CIFAR10~\citep{cifar100}, one of the most widely used datasets in machine learning research. The dataset comprises 60'000 color images from 10 classes and is publicly available.\footnote{\url{https://www.cs.toronto.edu/~kriz/cifar.html}}

\paragraph{M5 on Google Speech Commands.}
We explore the performance of SCNs in the audio signal processing domain by adopting
M5~\citep{DaiDQLD16} convolutional architecture to classify keywords in the Google Speech Commands dataset~\citep{warden2019SC}). M5 networks are trained on the time domain waveform inputs. The dataset consists of over 105,000 WAV audio files of various speakers saying 35 different words and is available under the Creative Commons BY 4.0 license. It is part of the Pytorch common datasets.\footnote{\url{https://pytorch.org/audio/stable/datasets.html}}

\subsection{Training hyper-parameters}

Table \ref{train_params} summarizes the set of hyper-parameters used to train different networks throughout this work. 
\begin{table}[!htb]
  \caption{\textbf{Training hyper-parameters} for all architecture-dataset pairs.}
  \label{train_params}
  \centering
  \begin{threeparttable}
  {\small
  \begin{tabular}{lllllll}
    \toprule
                    & MLP- & ShallowCNN- & ResNet18- & LeNet5- & M5- \\
    Hyper-param.    & FMNIST & SVHN & CIFAR10 & ModelNet10 & SpeechCmds \\
    \midrule
    Optimizer       & Adam         & Adam      & Adam      & Adam      & Adam\\
    LR              & 0.001        & 0.001     & 0.001     & 0.006     & 0.01\\
    Weight decay    &              &           & 0.0001    &           &  \\
    LR schedule  & CosineLR  & CosineLR & CosineAnnealing  & CosineAnnealing & CosineAnnealing \\
      & & & WarmRestarts, & $T_{max}=6'000$, & $T_{max}=100$, \\
      & & & $T_0=25$        & $\eta_{min} = 5\cdot10^{-6}$ & $\eta_{min} = 0$  \\
      & & & $T_{mult}=25$   & \\
    Batch size       & 64               & 256           & 512    & 256      & 256 \\
    Epochs           & 500             & 500          & 1'000    & 6'000    & 100 \\
    % Momentum         & 0.9              & 0.9           & 0.9  \\
    % Weight Decay     & -                & -             & 0.0001  \\
    Data augment.& Normali-    & Normali- & Normalization, & None & Resample \\
                 & zation      & zation   & HorizontalFlip &      & to 16 KHz \\
    \bottomrule
  \end{tabular}
  }
  \end{threeparttable}
  \label{table:training hyper-parameters}
\end{table}

\subsection{Transformations}

\begin{figure*}[!h]
\centering
    \subfloat[\textbf{2D rotation transformation parameterized by an angle $\phi$ in the range (0--$2\pi$)}. The transformation preserves angles and distances and can be undone with little loss of image quality (the edges of the input image may get cropped, rounding effects may occur).]{\includegraphics[width=\linewidth]{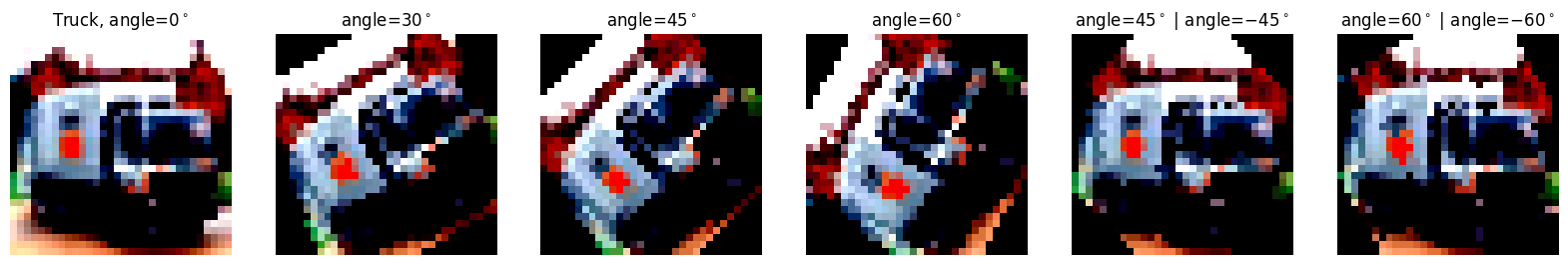}}
    
    \subfloat[\textbf{Scaling transformation parameterized by a scaling factor in the range (0.2--2.0)}. Preserves only angles, not fully invertible, reduces input quality, large portions of the input image may get cropped.]{\includegraphics[width=\linewidth]{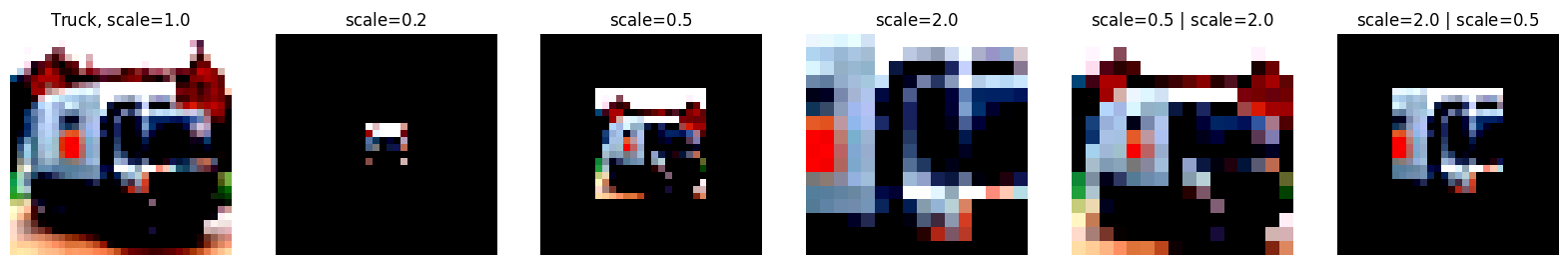}}
    
    \subfloat[\textbf{Translation transform with a shift in (-8,-8)--(8,8)}. Fully invertible only for the part of the input image inside the middle square (8,8) to (24,24).]{\includegraphics[width=\linewidth]{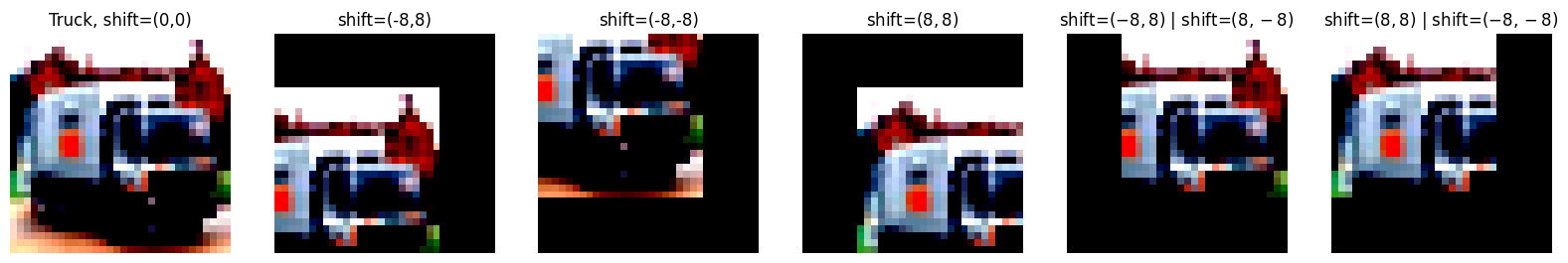}}
    
    % \subfloat[\textbf{2D Permutation transform (tiling 2$^i$,$i=1..5$)}. Discrete transform, does not preserve angles and distances, fully invertible.]{\includegraphics[width=\linewidth]{figs/sample_transformations/transform_permutation}}
    
    \subfloat[\textbf{3D rotation transform}. We rotate an object in 3D along XY, YZ, and XZ planes using 3 angles $(\phi_1, \phi_2, \phi_3)$, $\phi_i \in (-\pi, \pi)$ and sample a point cloud of 4'096 points. Rotations in XZ (\eg angles=$(0, \frac{\pi}{2}, 0)$) and YZ (\eg angles=$(\frac{\pi}{2}, 0, 0)$) planes can block some pixels (\eg the table surface, which is not visible in the picture).]{\includegraphics[width=\linewidth]{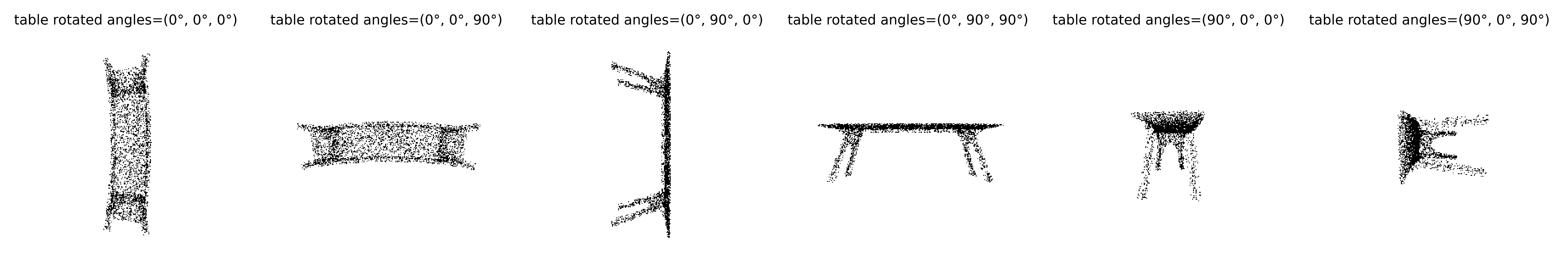}}
    \caption{\textbf{Geometric transformations} used in this work applied to a sample input. Notice how the images get impacted when inverse transformation is applied, showing a loss of input quality due to rounding, re-scaling and cropping.}
    \label{fig:transforms}
\end{figure*}

This paper evaluates SCNs on the following computer vision and audio signal transformations: 2D rotation, scaling, translation, 3D rotation-and-projection, brightness, contrast, saturation, sharpness, pitch shift and speed change described below. \Figref{fig:transforms} illustrates examples of transformations applied to a sample input, showcasing various non-obvious effects that result in a decrease in input quality. Consequently, this decrease in quality adversely affects the accuracy of a trained classifier.

\paragraph{2D rotation.} The rotation transformation is parameterized by a single angle $\phi$ in the range 0--$2\pi$. We use $\alpha = (\cos(\phi), \sin(\phi))$ as input to the configuration network when learning SCNs for 2D rotations. The transformation preserves distances and angles, yet may lead to information loss due to cropped image corners and rounding effects. It can be inverted with little loss of image quality, as can be observed in \Figref{fig:transforms}.

\paragraph{Scaling.} The transformation is parameterized by the scaling factor in the range 0.2--2.0, which is input to the hypernetwork to learn the configuration $\beta$-space for this transformation. Scaling transformation leads to a considerable loss of image quality. When inverted, the image appears highly pixelated or cropped.

\paragraph{Translation.} We consider image shifts within the bounds (-8,-8) and (8,8) pixels. A shift is represented by two parameters $\alpha = (\alpha_x, \alpha_y)$ reflecting the shift along the $x$ and $y$ axes. Note that an image gets cropped when translation is undone. In the FMNIST dataset the feature objects are positioned at the center of the image, which mitigates the negative impact of translations compared to other datasets like SVHN and CIFAR10.

\paragraph{3D rotation.} The 3D rotation transformation is parameterized by the three Euler angles that vary in the range $(-\pi, \pi)$. We use $\alpha=(\cos(\phi_1), \sin(\phi_1),\cos(\phi_2),\sin(\phi_2),\cos(\phi_3),\sin(\phi_3))$
as the input to the hypernetwork for learning SCNs on 3D rotations. Note that a different order of the same combination of these three angles may produce a different transformation output. We apply a fixed order $(\phi_1, \phi_2, \phi_3)$ in all 3D rotation experiments. After rotation the 3D point cloud is projected on a 2D plane. When applying 3D rotations, it is possible to lose pixels in cases where the rotation axis is parallel to the projection plane. An example is shown in \Figref{fig:transforms}.

\paragraph{Color transformations.} 
We explore SCN performance on four common color transformations: brightness, contrast, saturation and sharpness. 
The \emph{brightness} parameter governs the amount of brightness jitter applied to an image and is determined by a continuously varying brightness factor. The \emph{contrast} parameter influences the distinction between light and dark colors in the image. \emph{Saturation} determines the intensity of colors present in an image. Lastly, \emph{sharpness} controls the level of detail clarity in an image. We vary the continuously changing $\alpha$ parameter between 0.2 and and 2.0 for all considered color transformations.

\paragraph{Audio signal transformations.} 
We use SCNs with pitch shift and speed adjustment transformations. Pitch shift modifies the pitch of an audio frame by a specified number of steps, with the parameter adjusted within the range of -10 to +10. Similarly, speed adjustment alters the playback speed by applying an adjustment factor, with speed changes applied within the range of 0.1 to 1.0.

\section{Configuration subspaces and SCN efficiency}
\label{sec:cn:accuracy}

\subsection{SCN performance}

\begin{figure*}[!h]
\centering
    \includegraphics[width=.24\linewidth]{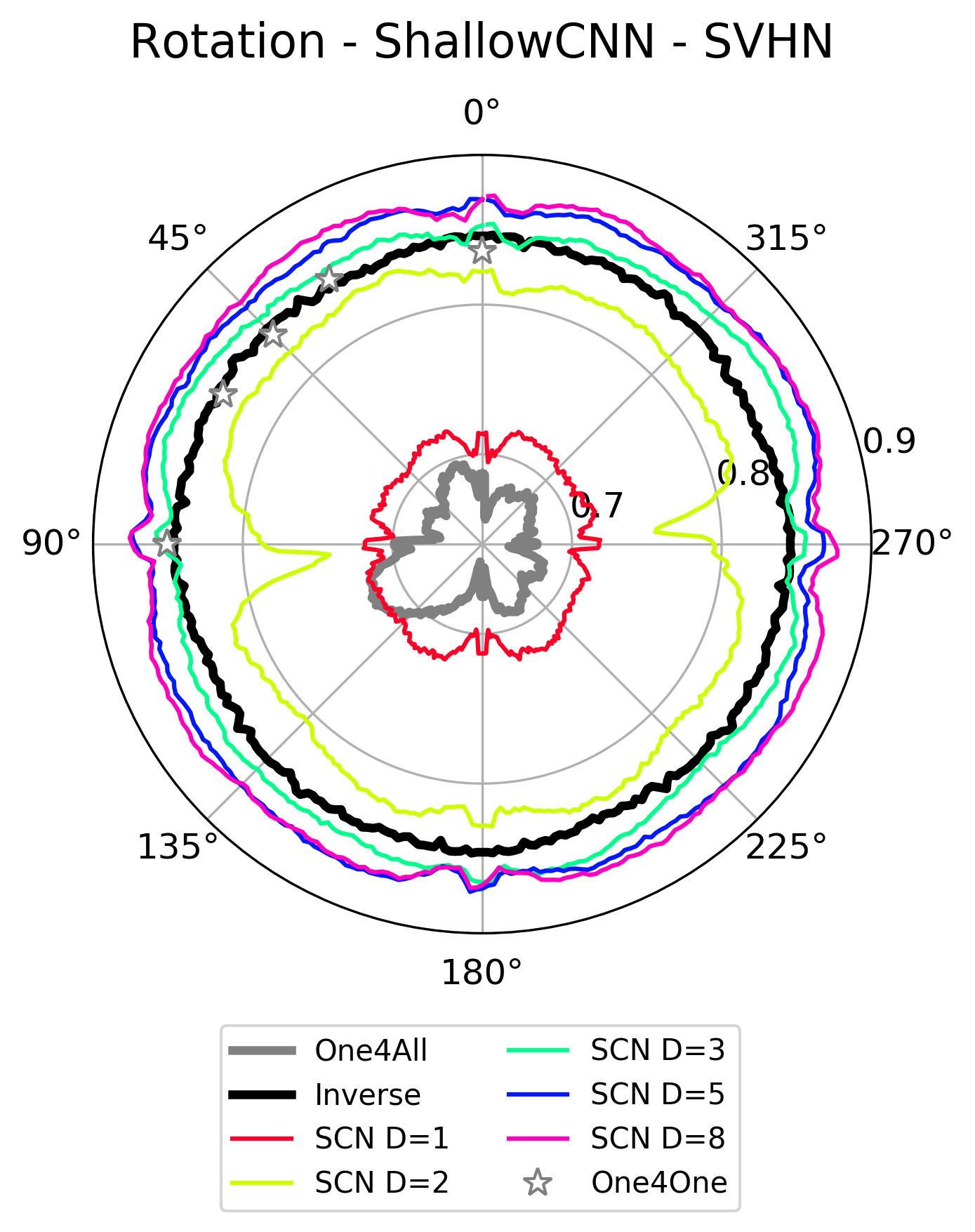}
    \includegraphics[width=.24\linewidth]{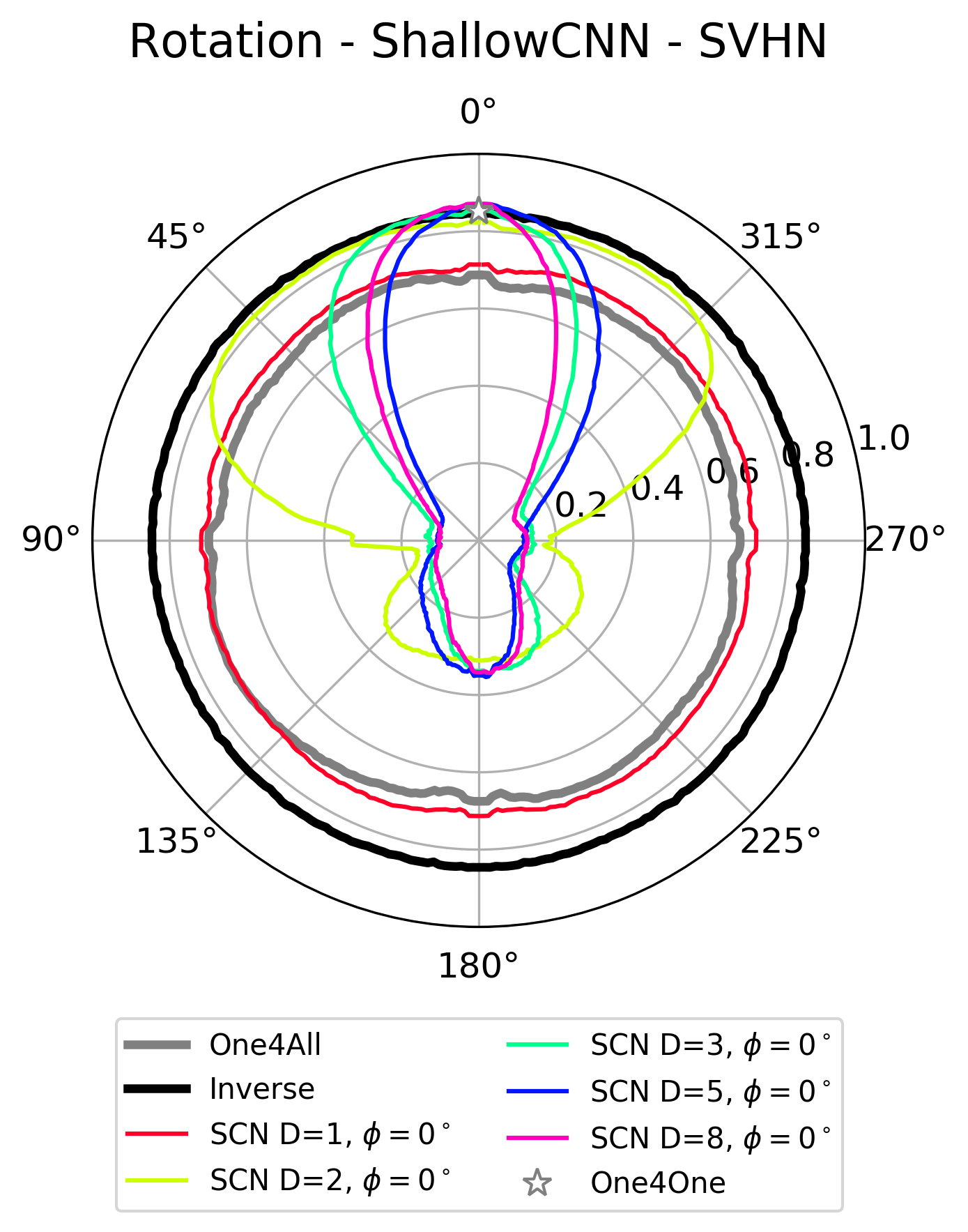}
    \includegraphics[width=.24\linewidth]{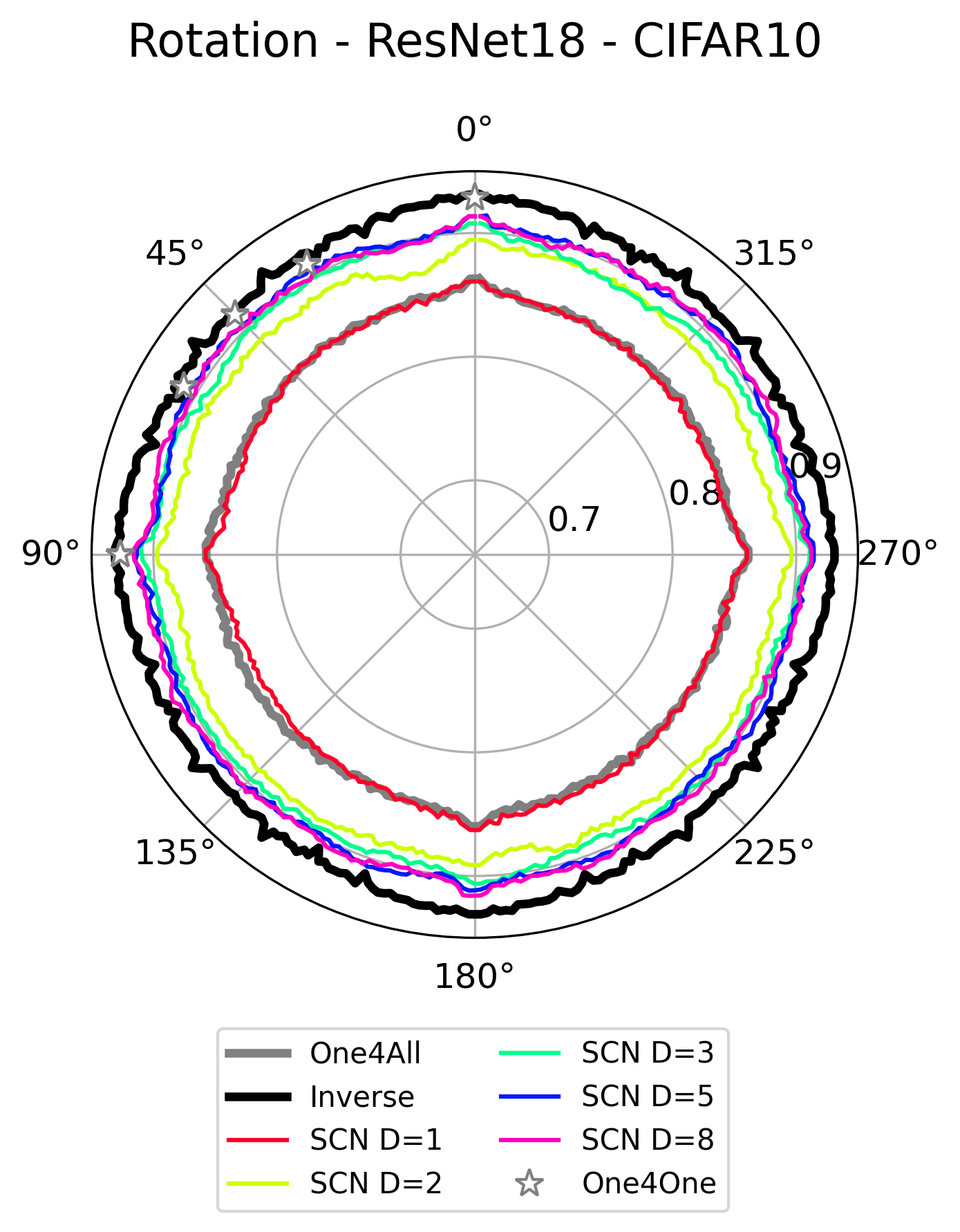}
    \includegraphics[width=.24\linewidth]{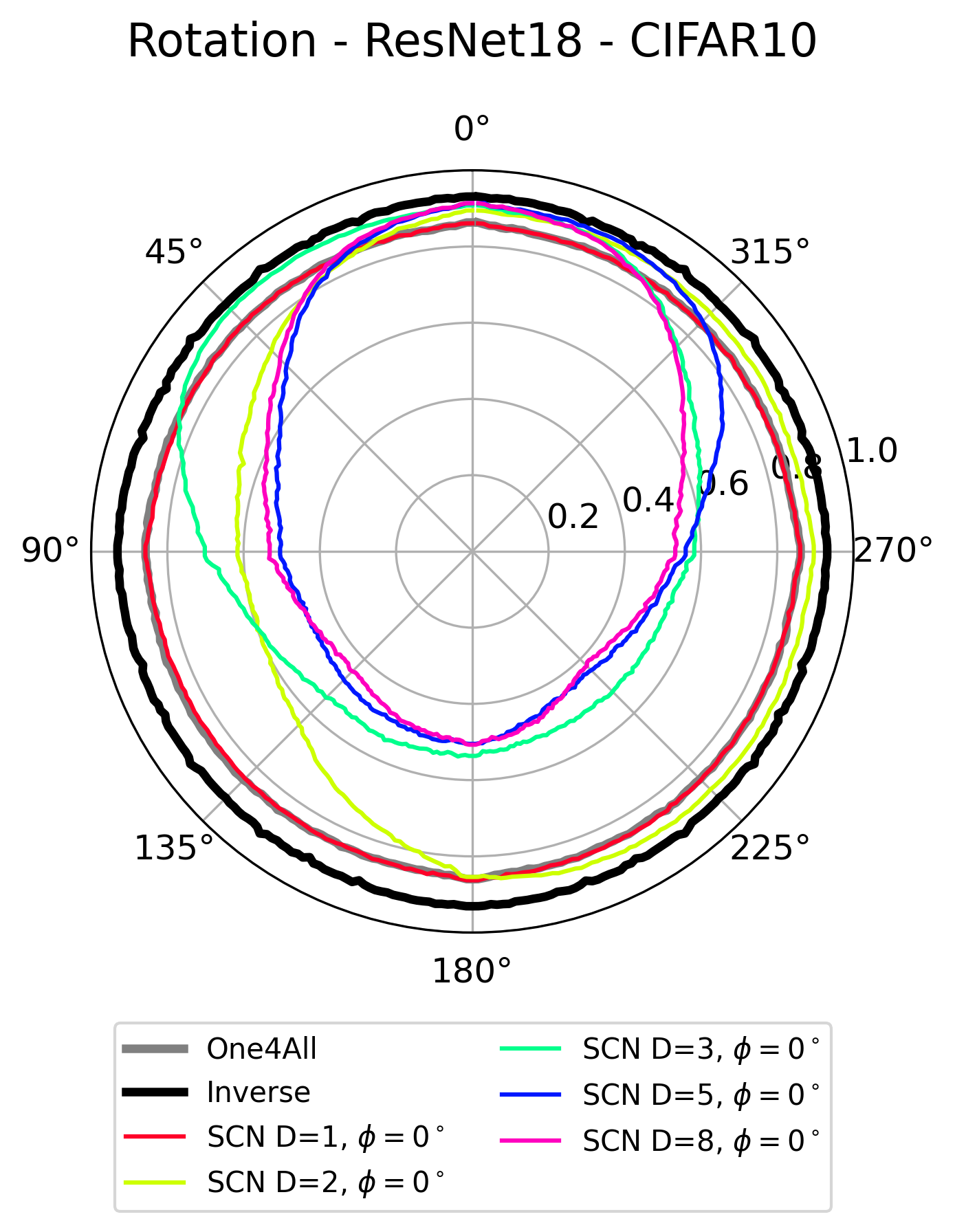}
    \caption{\textbf{SCN test set accuracy on 2D rotations}. \textbf{From left to right:} A pair of plots for ShallowCNN--SVHN and ResNet18--CIFAR10. The models in each pair show SCN's performance for changing input $\alpha=(\cos(\phi),\sin(\phi))$ and for a fixed $\alpha$ with $\phi=0^\circ$.}
    \label{fig:sconv:resnet:accuracy}
\end{figure*}

\paragraph{SCN performance on geometric transformations.}
\Figref{fig:sconv:resnet:accuracy} complements \Figref{fig:mlpb:accuracy} in the main paper and presents the performance of SCNs for 2D rotation on ShallowCNN--SVHN and ResNet18--CIFAR10. We observe high efficiency of SCNs compared to the baselines even for low $D$. A close inspection of models for a fixed degree ($\phi=0^\circ$) shows their increasingly higher specialization to the respective transformation parameter setting.

\begin{figure*}[h]
    \centering
     \includegraphics[width=.24\linewidth]{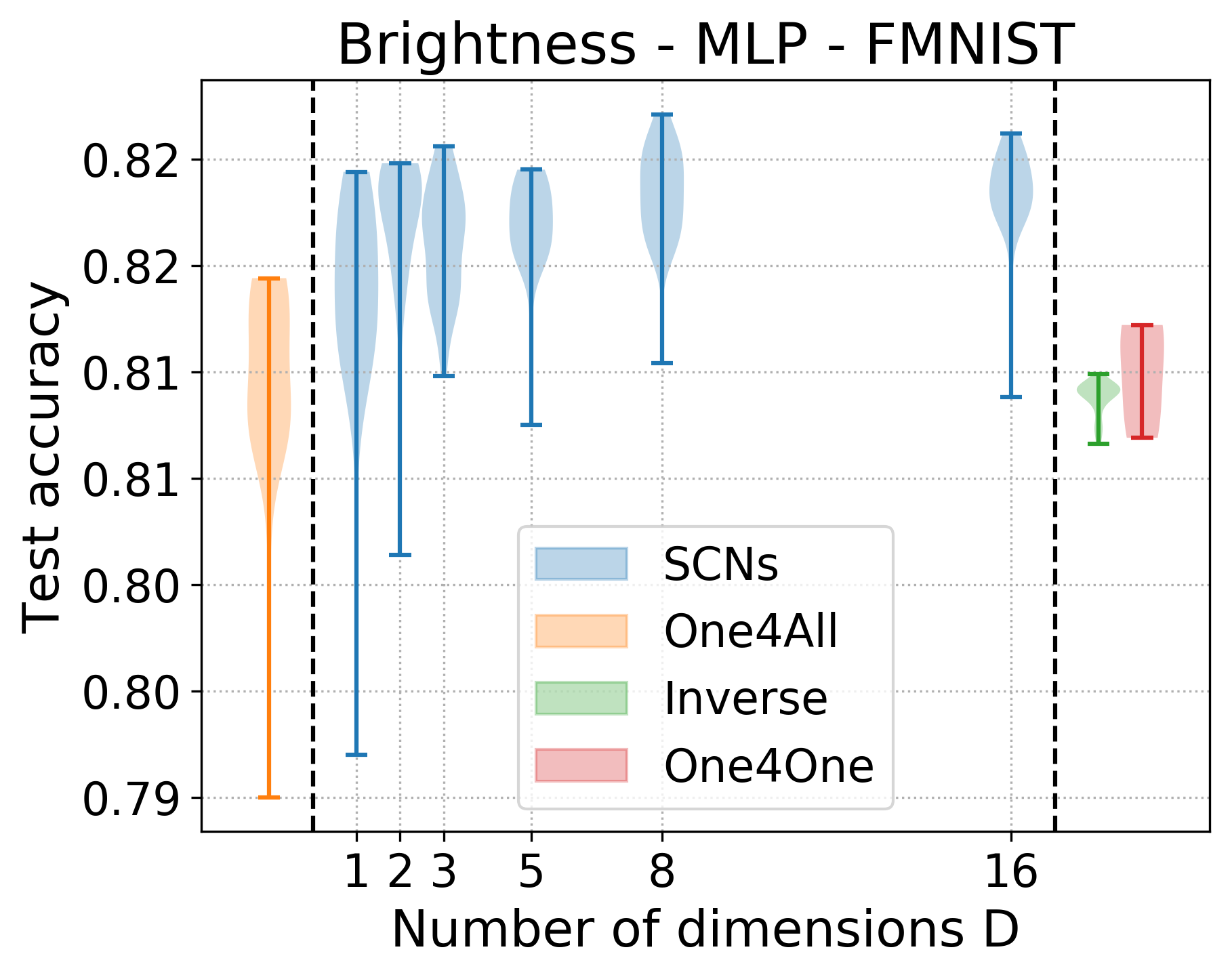}
     \includegraphics[width=.24\linewidth]{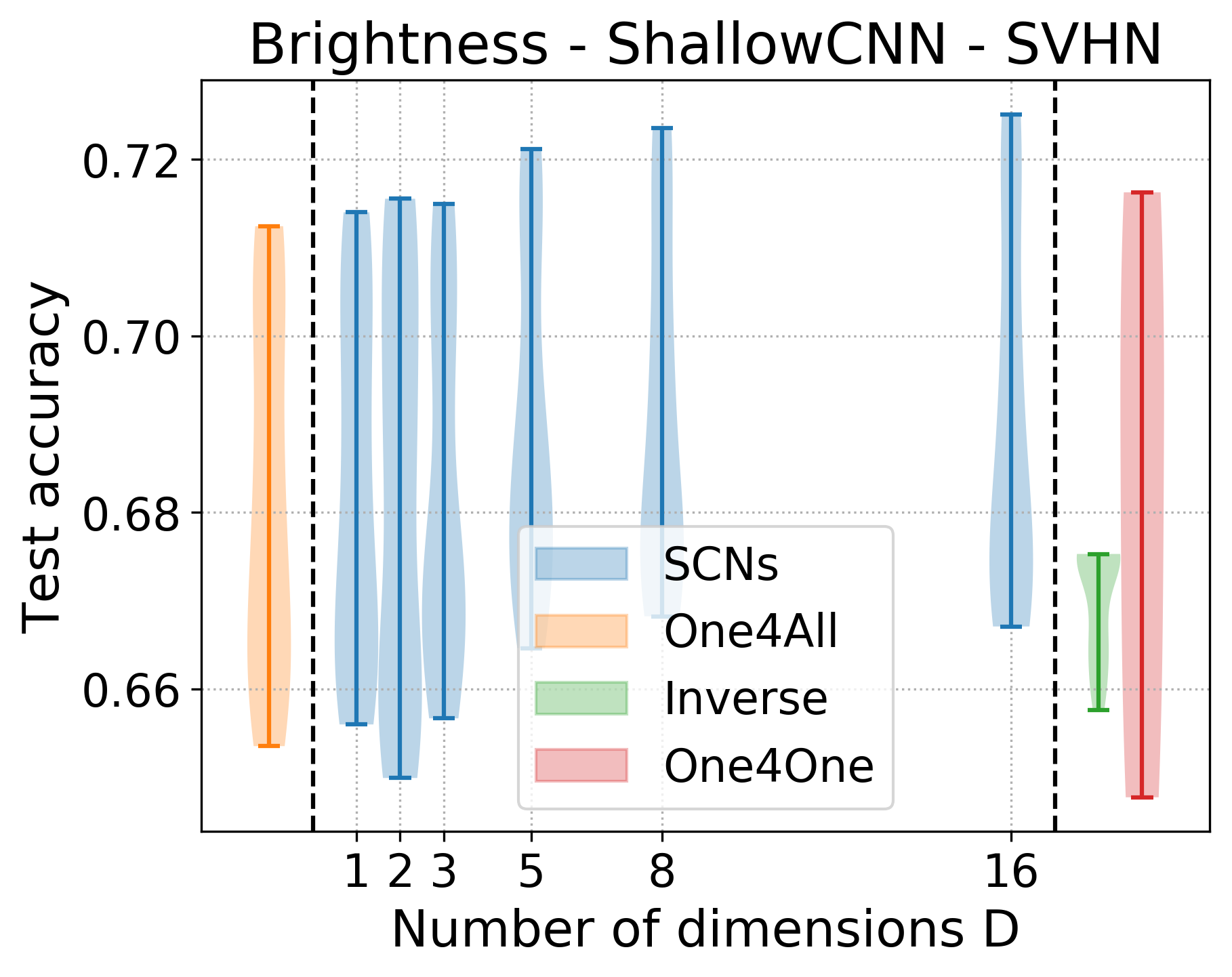}
     \includegraphics[width=.24\linewidth] {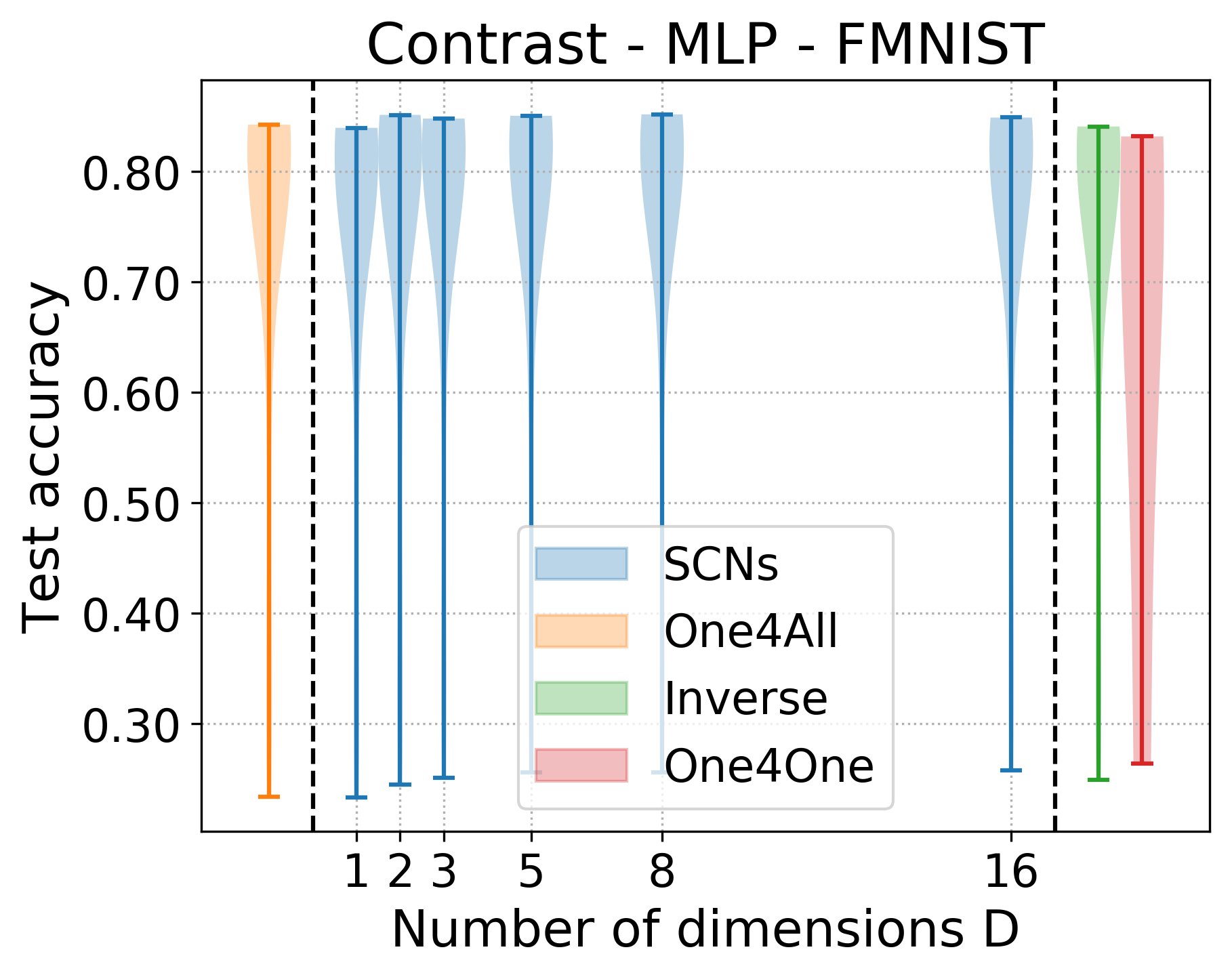}
     \includegraphics[width=.24\linewidth] {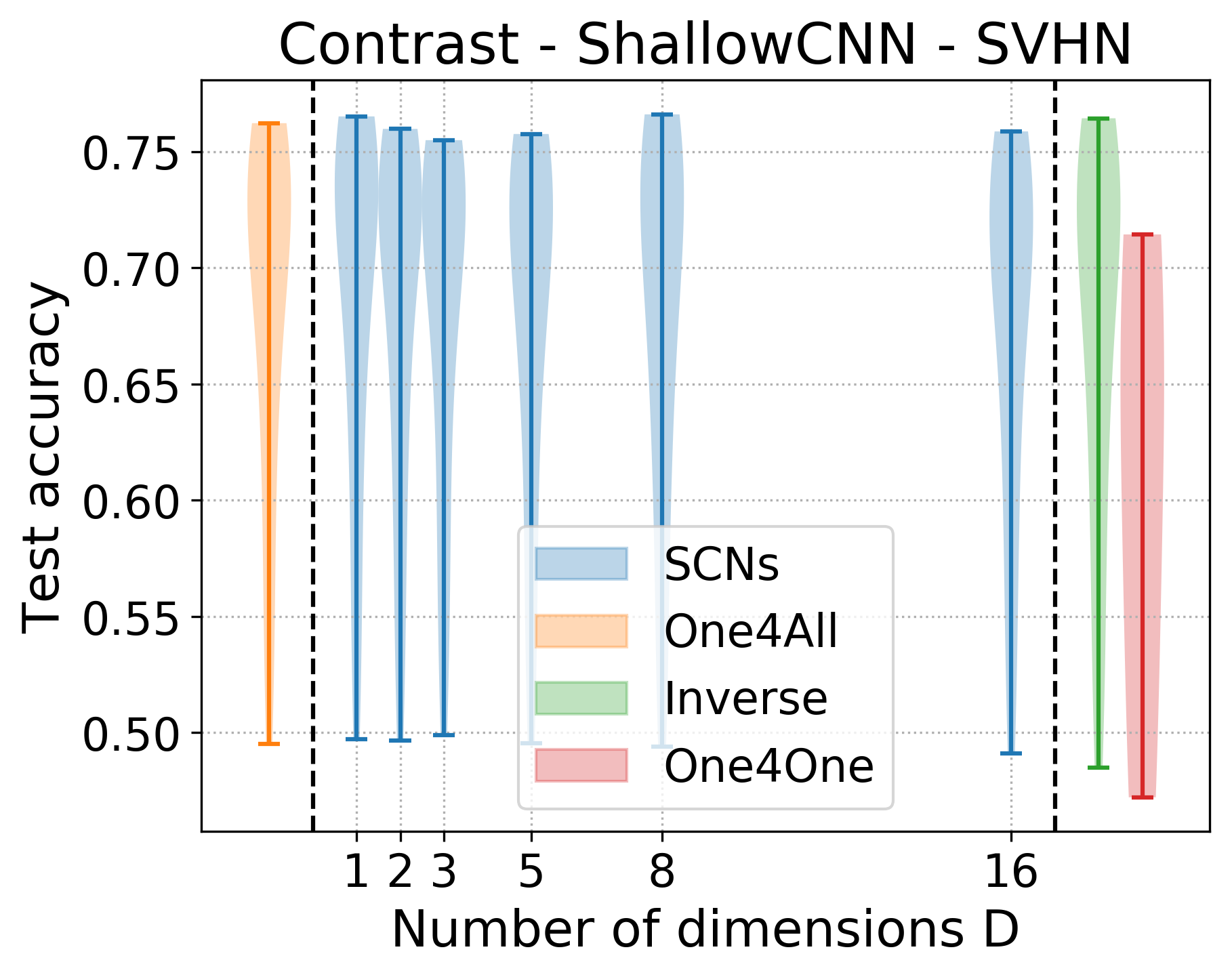}
     \includegraphics[width=.24\linewidth]{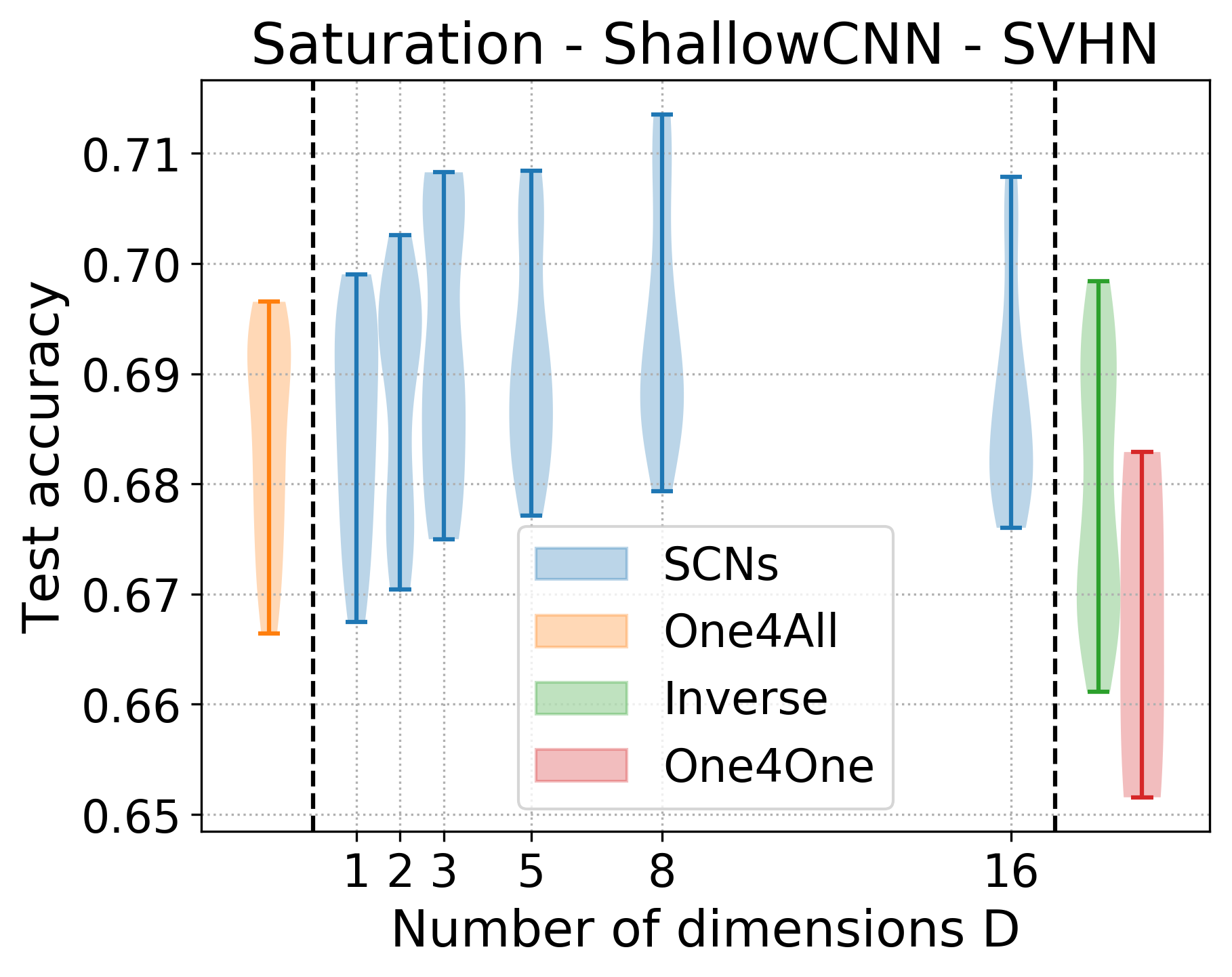}
     \includegraphics[width=.24\linewidth] {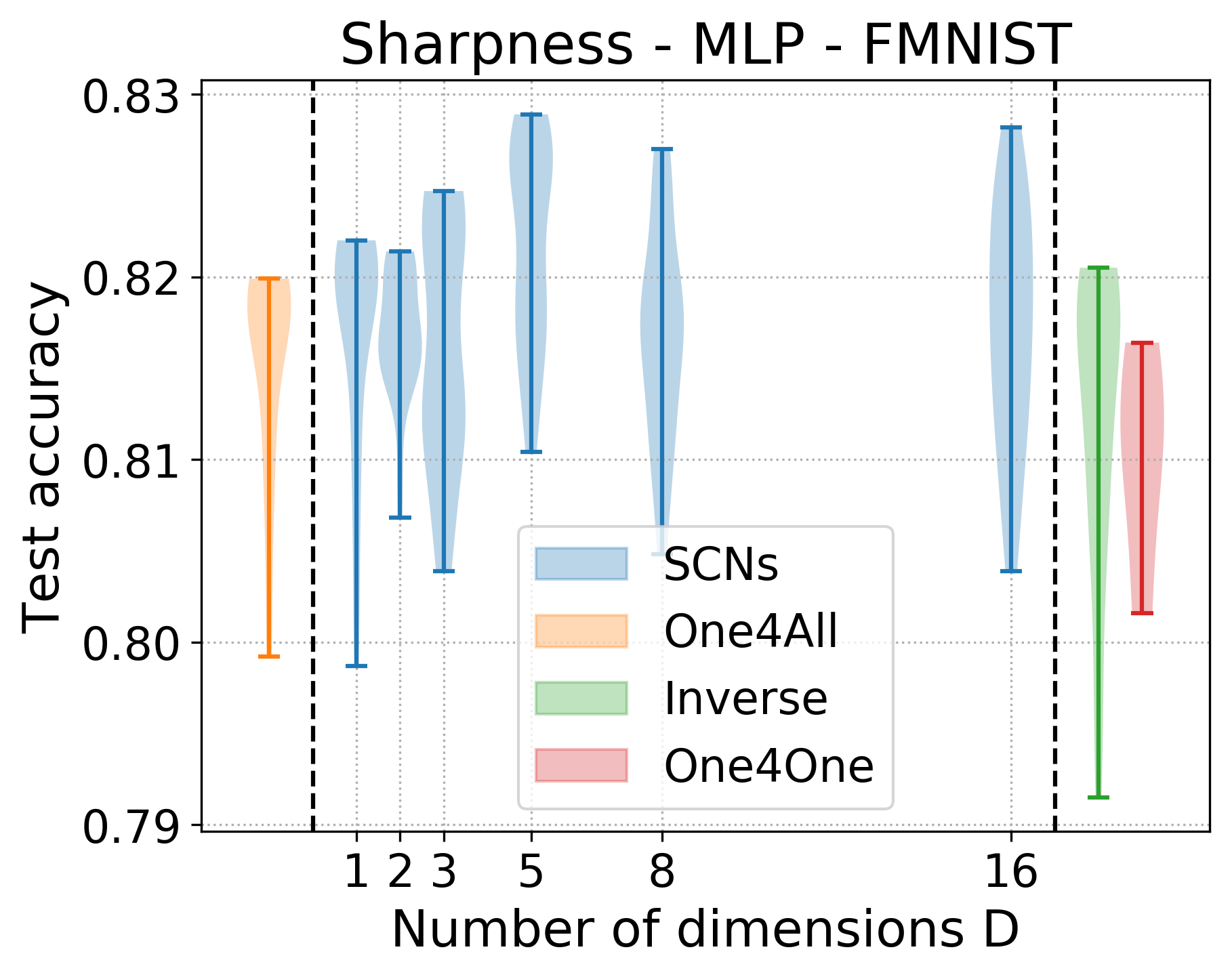}
     \includegraphics[width=.24\linewidth] {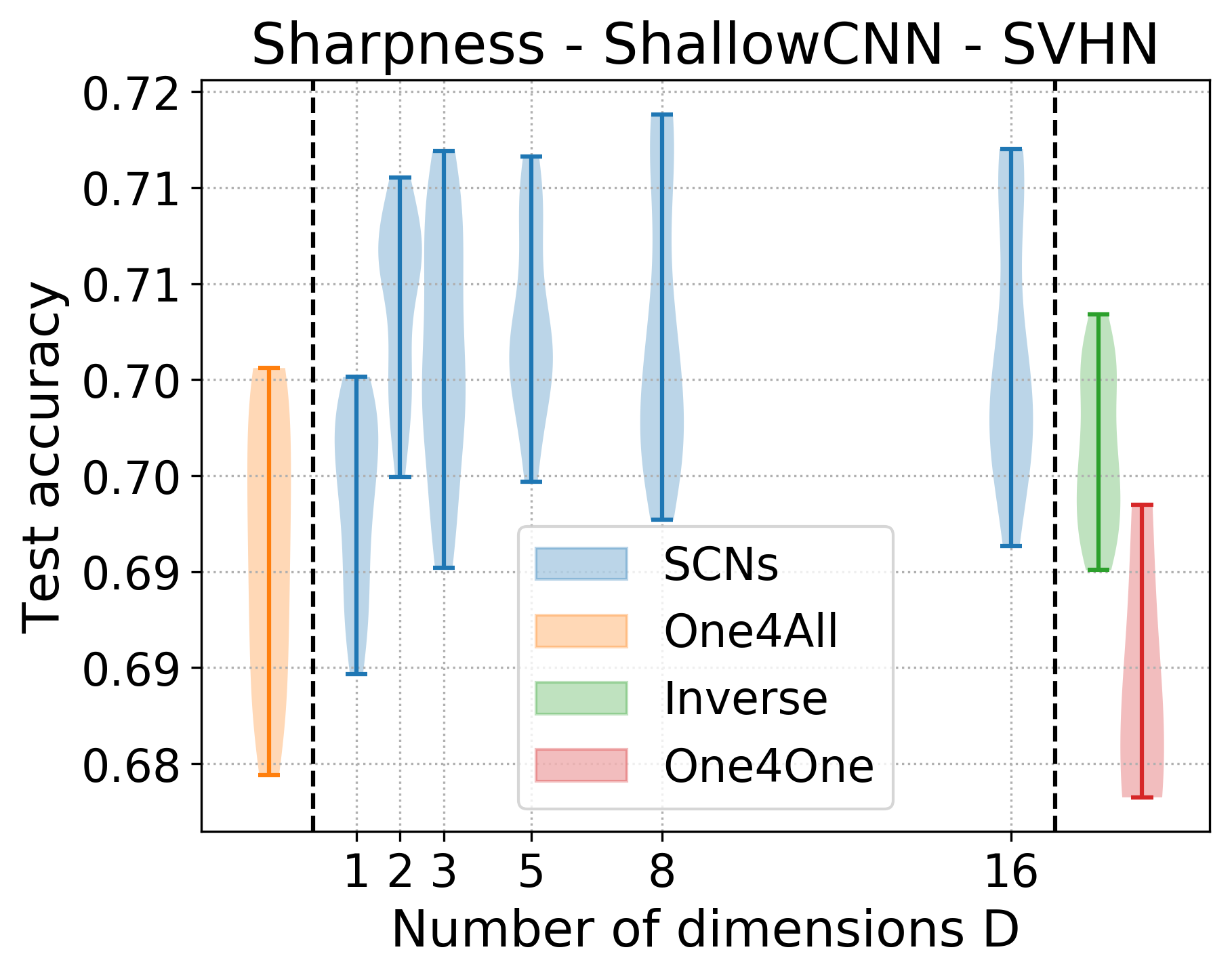}
    \caption{\textbf{Summary of SCN performance on color transformations:} brightness, contrast, saturation and sharpness. We present the results for MLP-FMNIST and ShallowCNN-SVHN architecture-dataset pairs. All transformations are simple. SCNs match the baselines for very low $D$. Note that saturation has no effect on black-white images. Therefore, for MLP-FMNIST the difference in model performance is the same up to the choice of a random seed.}
    \label{fig:effect:d:appendix:cv}
\end{figure*}

\paragraph{SCN performance on color transformations.}
Color transformations are simple. SCNs achieve high performance already for very low $D=2$ or $D=3$ (see \Figref{fig:effect:d:appendix:cv}). There is little performance difference between the baselines One4All, One4One and Inverse despite the small inference model size (1-layer MLP with 32 hidden units for FMNIST and 2-layer ShallowCNN with 32 channels for SVHN).

\begin{figure*}[h]
    \centering
     \includegraphics[width=.24\linewidth]{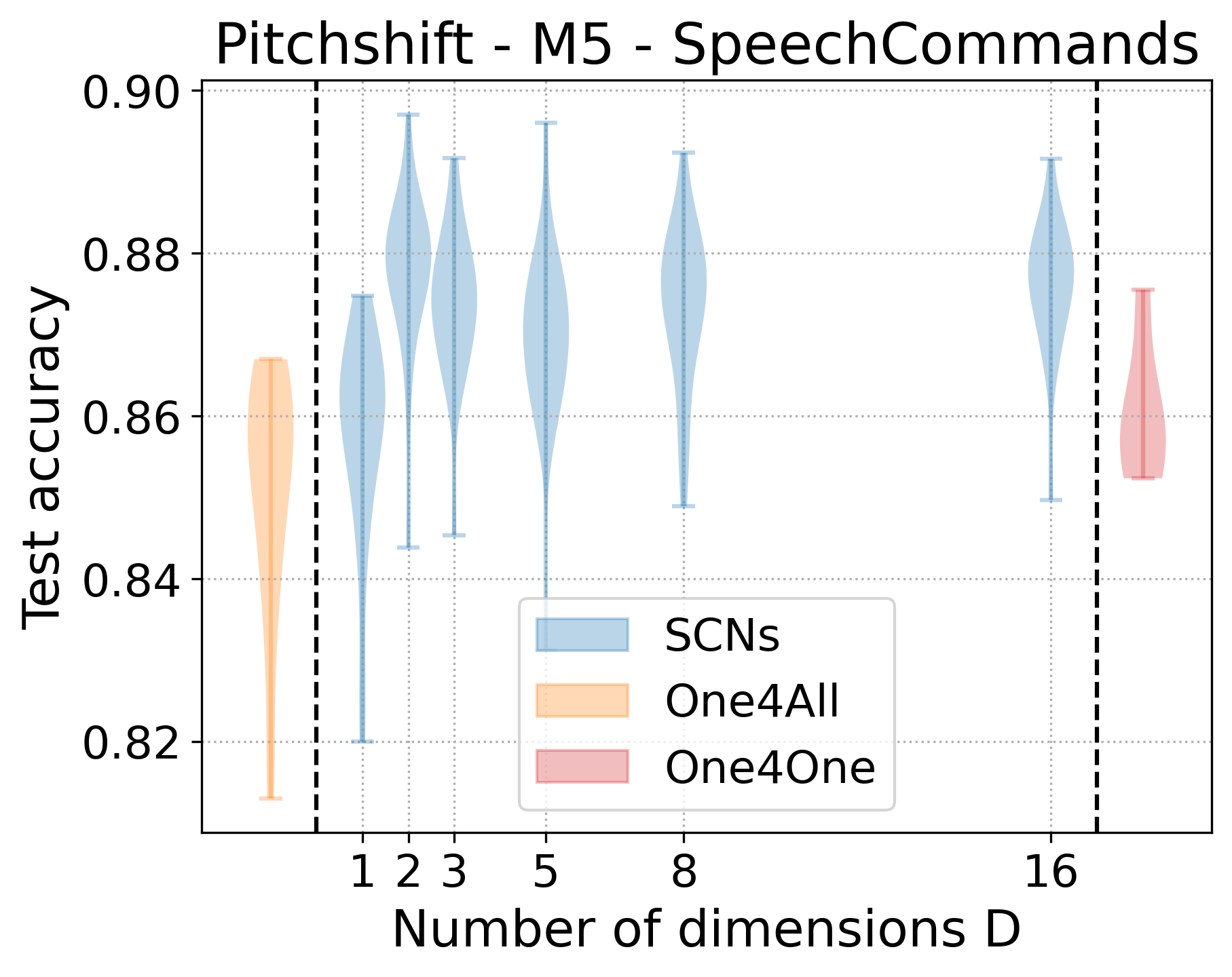}
     \includegraphics[width=.24\linewidth]{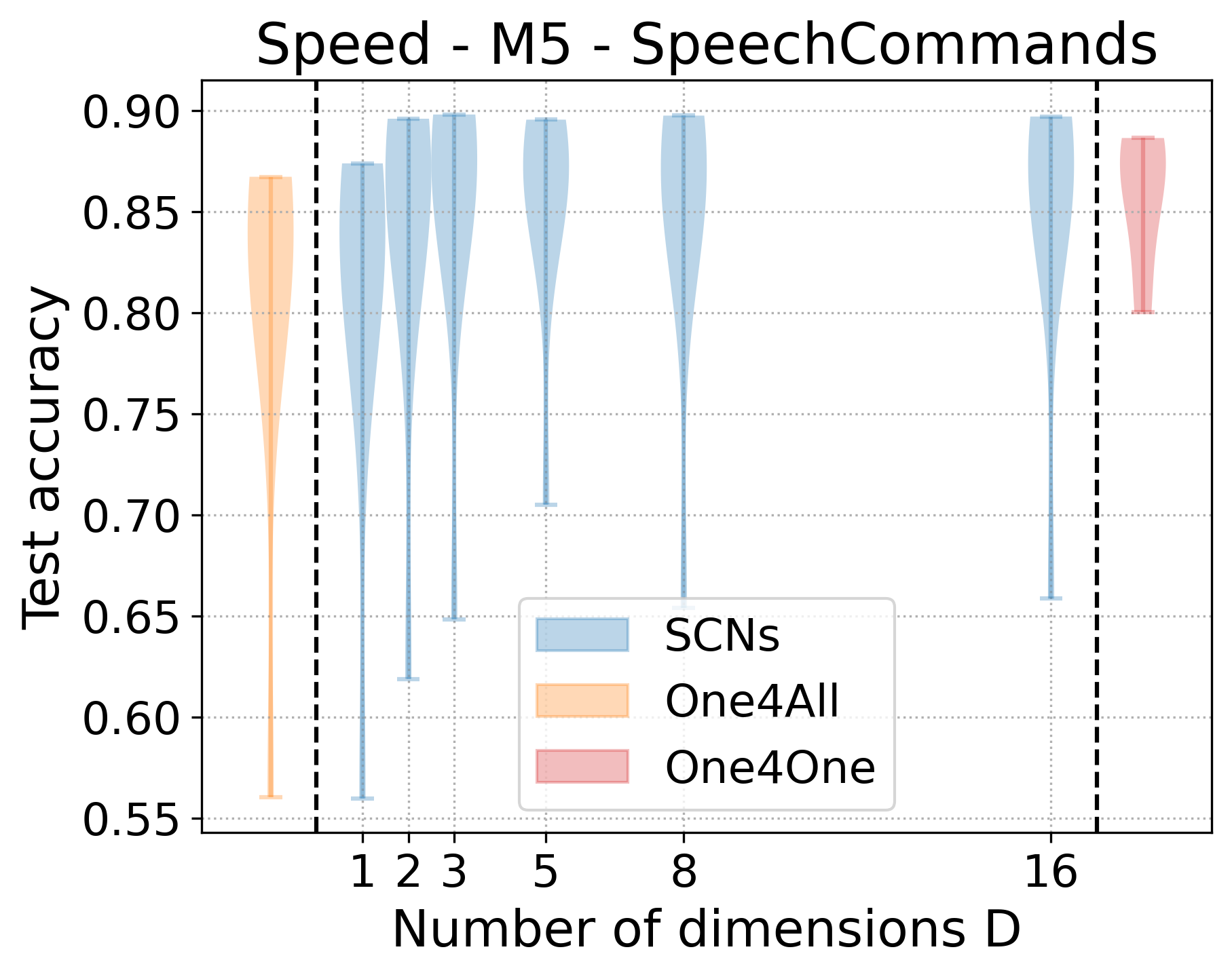}
    \caption{\textbf{Summary of SCN performance on audio signal transformations:} pitch shift and speed using M5 as inference architecture. SCNs match the performance of baselines already for small $D$.}
    \label{fig:effect:d:appendix:audio}
\end{figure*}

\paragraph{SCN performance on audio signal transformations.}
\Figref{fig:effect:d:appendix:audio} shows the performance of SCNs on two audio signal transformations: pitch shift and speed change. For both transformations a low $D$ is sufficient for SCN to match or outperform the baselines. Note that M5 takes a raw waveform in the time domain as input rather than a spectrogram.

\subsection{Configuration $\beta$-space visualization}
\label{sec:betaspace}

\begin{figure*}[!th]
\centering
    \includegraphics[width=\linewidth]{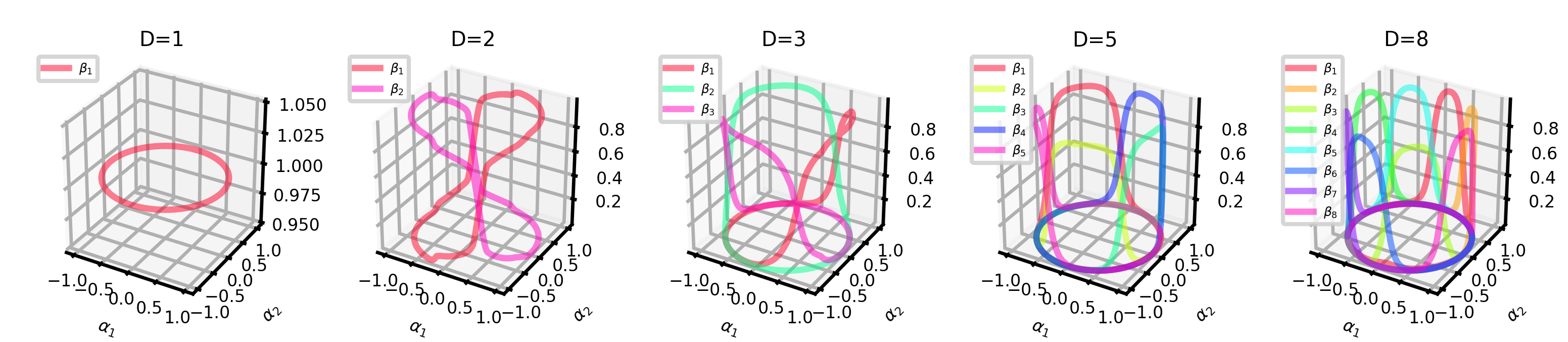}
    \includegraphics[width=\linewidth]{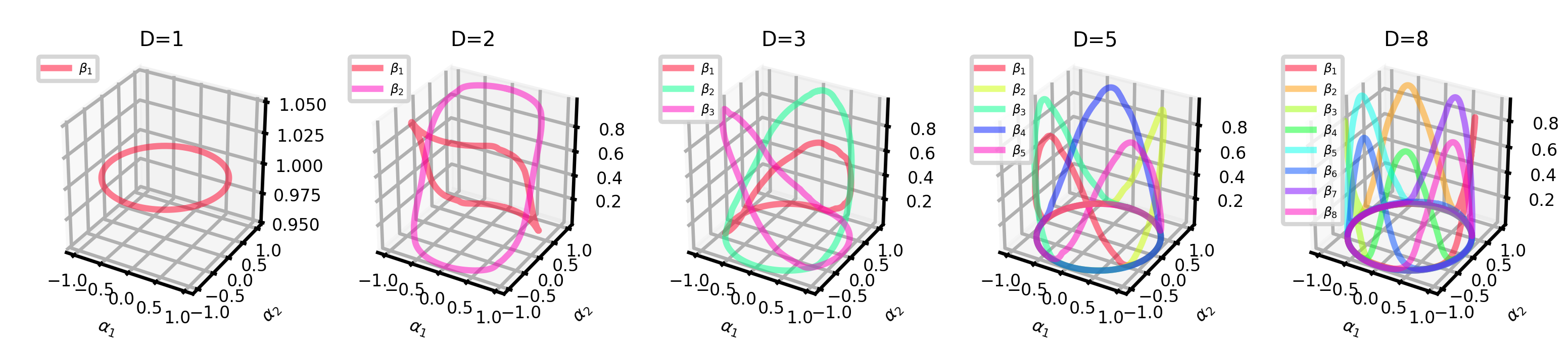}
    \caption{\textbf{Configuration $\beta$-space of SCNs trained for 2D rotation} on further datasets and inference network architectures, complementing \Figref{fig:beta}. \textbf{Top:} 1-layer MLP with 32 hidden units on FMNIST. \textbf{Bottom:} 2-layer ShallowCNNs with 32 filters in the hidden layers on SVHN. Transform parameters are $\alpha = (\alpha_1, \alpha_2) = (\cos(\phi), \sin(\phi))$, with $\phi$ being a rotation angle.}
    \label{fig:beta:rotation}
\end{figure*}

\begin{figure*}[ht]
\centering
    \includegraphics[width=\linewidth]{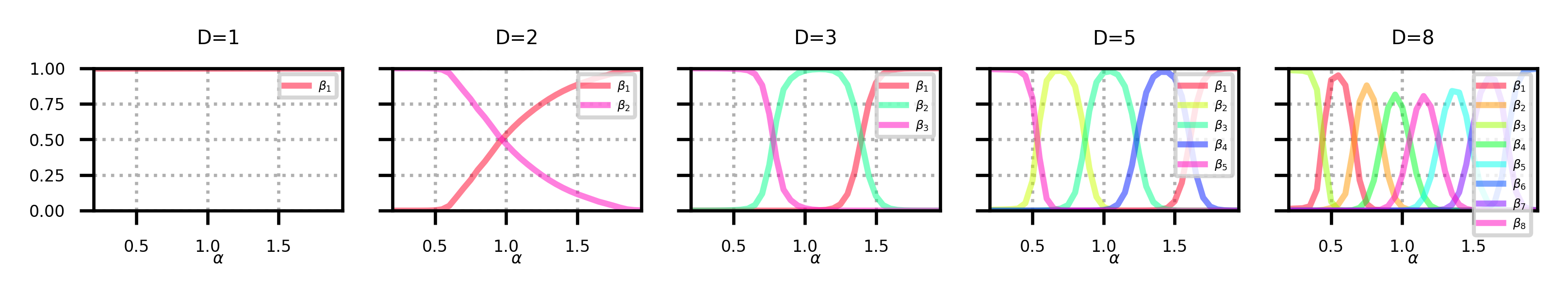}
    \caption{\textbf{Configuration $\beta$-space of SCNs trained for scaling using MLP on FMNIST}. Scaling factor $\alpha$ is between 0.2--2.0. The $\beta$-space looks similarly to the one shown in the main paper for a different dataset-architecture pair.}
    \label{fig:beta:scaling}
\end{figure*}

Different dataset-architecture pairs exhibit a similar structure in the $\beta$-space. \Figref{fig:beta:rotation} and \Figref{fig:beta:scaling} present the learned configuration parameters $\beta$ as a function of the transformation parameters $\alpha$ for 2D rotation and scaling, respectively, complementing the findings in \Figref{fig:beta} of the main paper. It is worth noting the slight variations in the shape of the learned curves, which are specific to the architecture-dataset pairs used to train SCNs. Based on the consistent $\beta$-space across different dataset-architecture pairs, we infer that the configuration space primarily relies on the transformation and the characteristics of its parameter vector $\alpha$.

\subsection{SCN at a block level}

\begin{figure}[ht]
\centering
    \includegraphics[width=.5\linewidth]{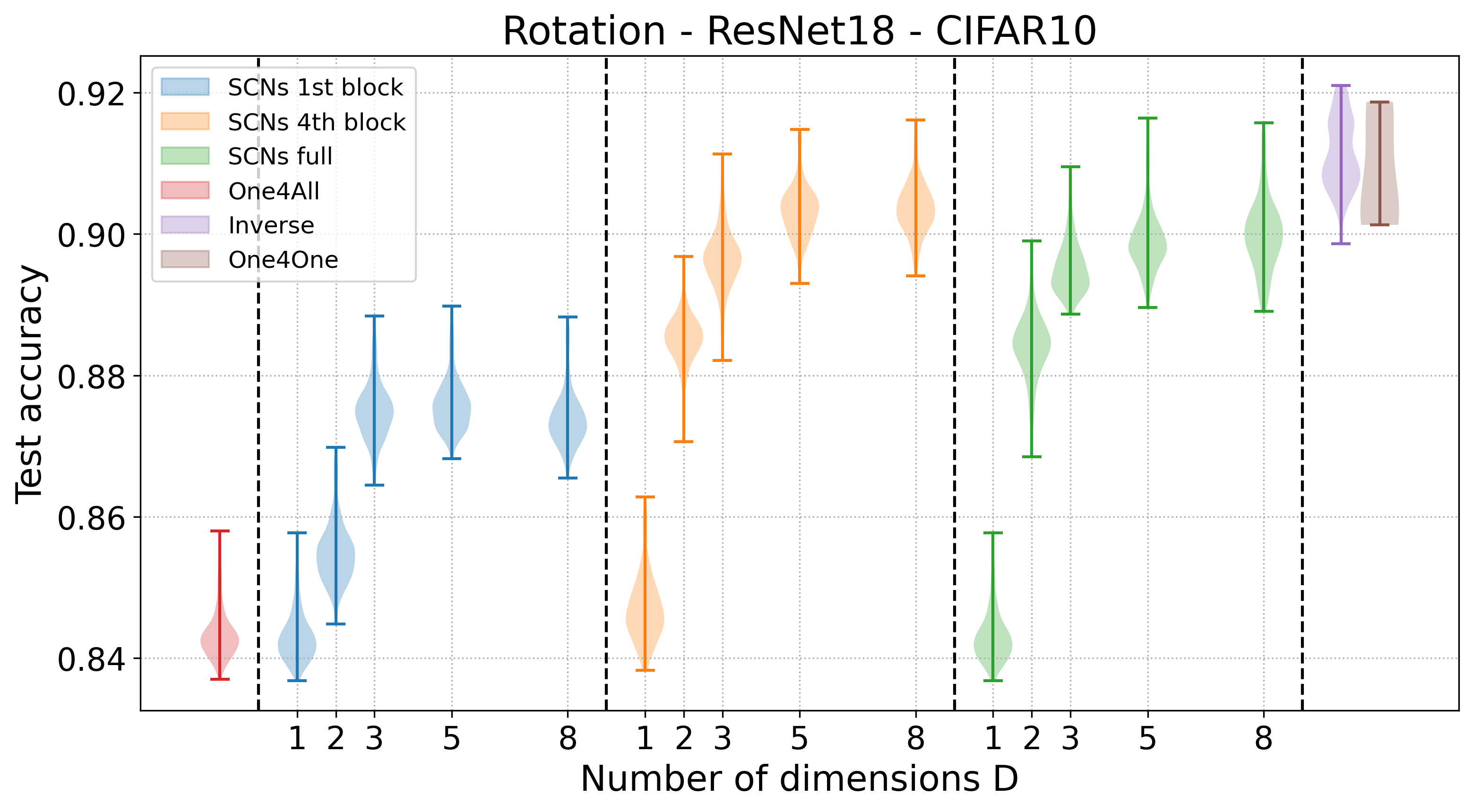}
    \caption{textbf{SCN at a block level}. Performance results for the 2D rotation transformation applied to ResNet18 on CIFAR10. Applying SCN solely on the 4th block yields results comparable to SCN application across all blocks. Applying SCN exclusively to the 1st block results in inferior performance.}
    \label{fig:partial:scn}
\end{figure}

We investigate the performance of SCNs when applied to just one block of ResNet18 on CIFAR10 for 2D rotation transformations in \Figref{fig:partial:scn}. Applying SCN solely on the 4th block yields results comparable to SCN's application across all blocks. However, applying SCN exclusively to the 1st block results in inferior performance.

\subsection{Comparison to domain adaptation methods}

\begin{table}[b]
\centering
\begin{tabular}{l|c|c|c}
% \hline
\toprule
               & \textbf{FMNIST Rotated[\%]} & \textbf{FMNIST Rotated[\%]} & \textbf{FMNIST Rotated[\%]} \\
\textbf{Model} & \textbf{degree=10}          & \textbf{degree=90}          & \textbf{degree=0...360} \\
% \hline
\midrule
SAFN               & $42.41\pm6.93$        & $9.61\pm0.54$         & $10.11\pm0.16$           \\
Data Calibrator    & $77.16\pm0.75$        & $9.28\pm0.67$         & $19.77\pm0.39$           \\
\textbf{SCN(D=3)[ours]} & $86.15\pm0.21$        & $86.75\pm0.35$        & $73.75\pm2.40$           \\
\textbf{SCN(D=5)[ours]} & $86.51\pm0.56$        & $86.92\pm0.77$        & $85.71\pm3.35$           \\
% \hline
\bottomrule
\end{tabular}
\caption{Comparison of SCNs to SAFN and Data Calibrator for models adapted from FMNIST to the target domain. With $D$=5 SCN surpasses SAFN and Data Calibrator in multiple target domains}
\label{table:domain adaptation:accuracy}
\end{table}

Since robustness to input transformations can be framed as a domain shift problem, we compare SCNs with two domain adaptation baselines on the 2D-rotation task on FMNIST using LeNet-5 network architecture in Table~\ref{table:domain adaptation:accuracy}. The first baseline is the Stepwise Adaptive Feature Norm (SAFN)~\citep{xu2019larger}, which defines a distance measure between the source and the target domains in the feature space, and minimizes this distance when training the network to reduce the domain shift. Furthermore, we compare SCNs to the Data Calibrator~\citep{ye2020light}, which fixes the source classifier and recovers discrimination power in the target domain, while preserving the source domain's performance.

\section{Translation transformation}
\label{sec:translation}

\begin{figure*}[h]
\centering
    \includegraphics[width=\linewidth]{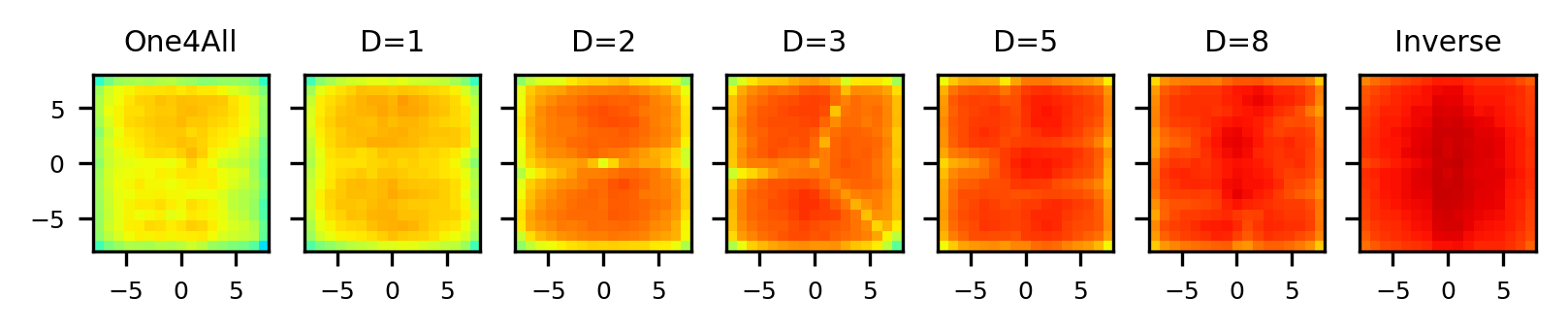} \\
    \includegraphics[width=\linewidth]{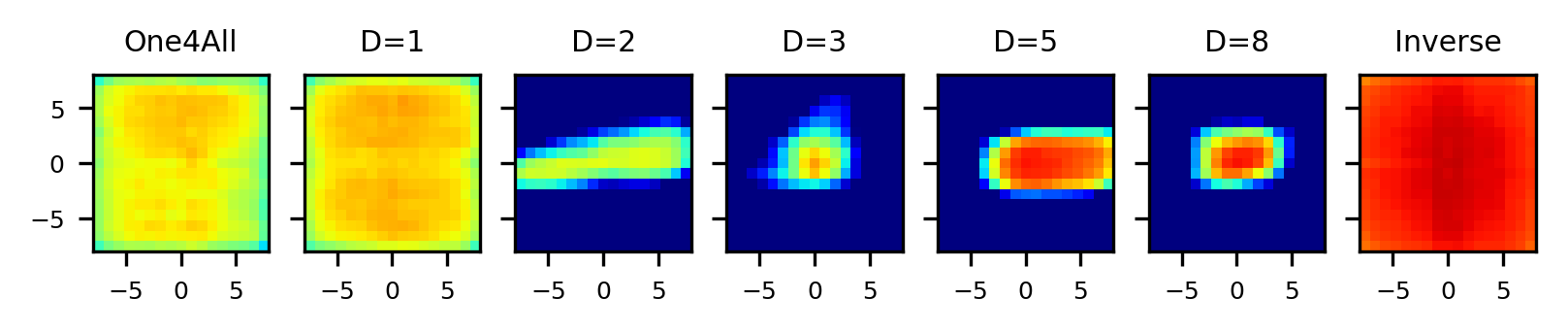}
    \caption{\textbf{SCN performance for translation trained with MLP inference network on FMNIST.} In this example, applying translation to an input image leads to information loss, since the part of the image shifted outside the image boundary gets cut. We use 1-layer MLPs with 32 hidden units and a bias term. This architecture is not translation-invariant. In all plots the color map is "rainbow" ranging uniformly between 0.5 or below (dark blue) to 0.9 (dark red). X and Y axes are horizontal and vertical shift parameters $(\alpha_x, \alpha_y)$ applied to the input. \textbf{Top:} Test accuracy of SCNs for $D=1..8$ for every $(\alpha_x, \alpha_y)$ combination, compared to One4All and Inverse baselines. \textbf{Bottom:} Test accuracy of SCNs for the dedicated fixed (0,0) shift evaluated on shifted inputs. The area of high accuracy decreases with higher $D$, leading to higher degree of model specialization, higher accuracy of the dedicated model for each setting, and a better overall performance of SCNs.}
    \label{fig:translation:noninv}
\end{figure*}

\paragraph{SCNs on architectures NOT invariant to translation.}
\Figref{fig:translation:noninv} shows SCN performance on translation transformation for 1-layer MLP with 32 hidden units as inference network trained on the FMNIST dataset. SCN's test accuracy increases with higher $D$ matching the accuracy of the Inverse baseline. The visualization allows identifying high accuracy areas of each base model. With higher $D$, the area of a dedicated model for a specific parameter setting decreases, whereas its test accuracy increases. 

\begin{figure*}[h]
\centering
    \includegraphics[width=.7\linewidth]{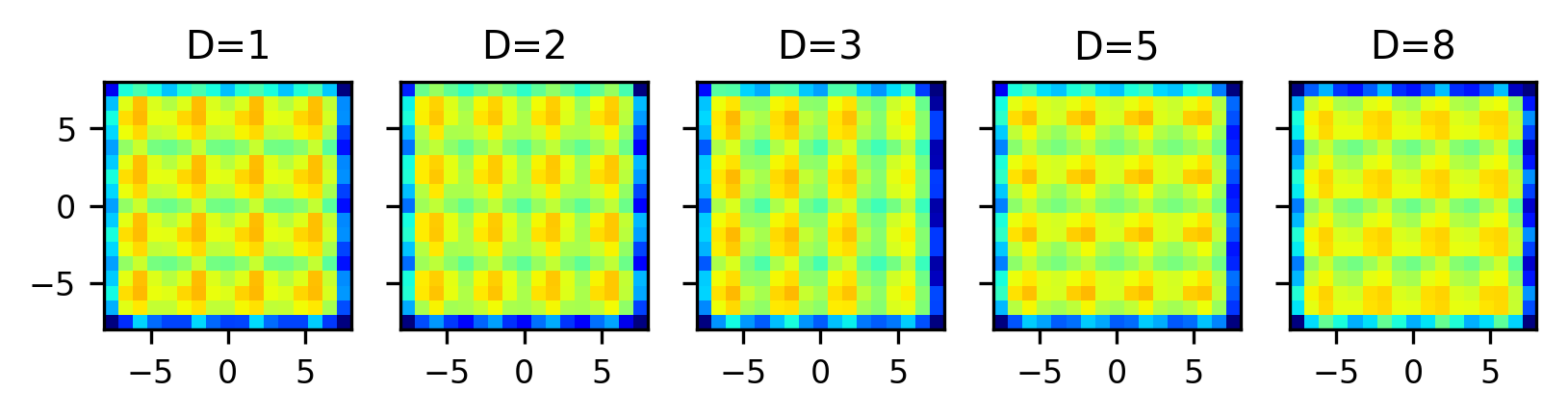} \\
    \includegraphics[width=\linewidth]{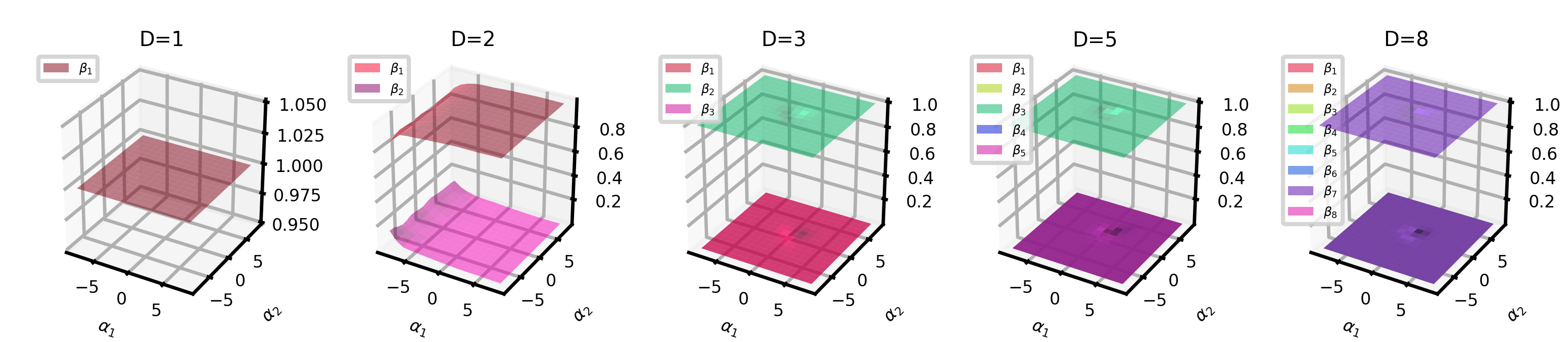}

    \caption{\textbf{SCN performance for translation trained on translation-invariant CNN architecture on FMNIST.} SCNs for all $D$ learn a degenerated $\beta$-space with only one active model (only one $\beta_i$=1) for all inputs. \textbf{Top:} Independently trained SCNs for different $D$ yield very similar accuracy of 0.85. The checkered structure of the plots reflects the size of the filters, which is 4x4. \textbf{Bottom:} Configuration $\beta$-space showing that only one $\beta_i$ equals 1.0 for all input parameters $\alpha$.}
    \label{fig:translation:inv}
\end{figure*}

\paragraph{SCNs on translation-invariant architectures.}
Using translation-invariant architecture as inference network part of SCN trained for translation results in a degenerated $\beta$-space with only one base model. This special case is exemplified in \Figref{fig:translation:inv}. On the FMNIST dataset side, we scale the input images down by 50\% and apply padding of 8 to ensure that shifting the image within (-8,8) along horizontal and vertical axes leads to a pure translation of the object in the image without information loss. As translation-invariant network architecture, we use a 2-layer CNN built only of convolutional and max pooling layers with kernel size of 4 and 16 channels. For any $D$, SCNs learn a single model with only one $\beta_i$=1 and other $\beta_j, j \neq i$ being zero. The checkered structure of the test accuracy plot reflects the size of the filters. A detailed explanation of its origin and its relation to the Nyquist–Shannon sampling theorem is given in \citep{zhang2019makingConvs}.

\section{3D rotation transformation}
\label{sec:3d:appendix}

\begin{figure*}[h]
    \centering
    \includegraphics[width=\linewidth]{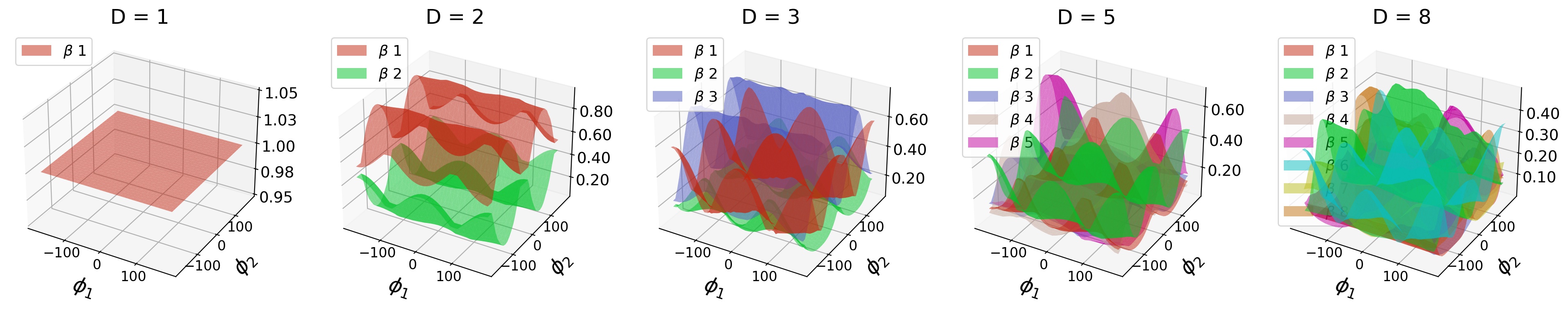}
     
    \includegraphics[width=\linewidth]{figs/3D/beta_alpha13.jpg}

    \includegraphics[width=\linewidth]{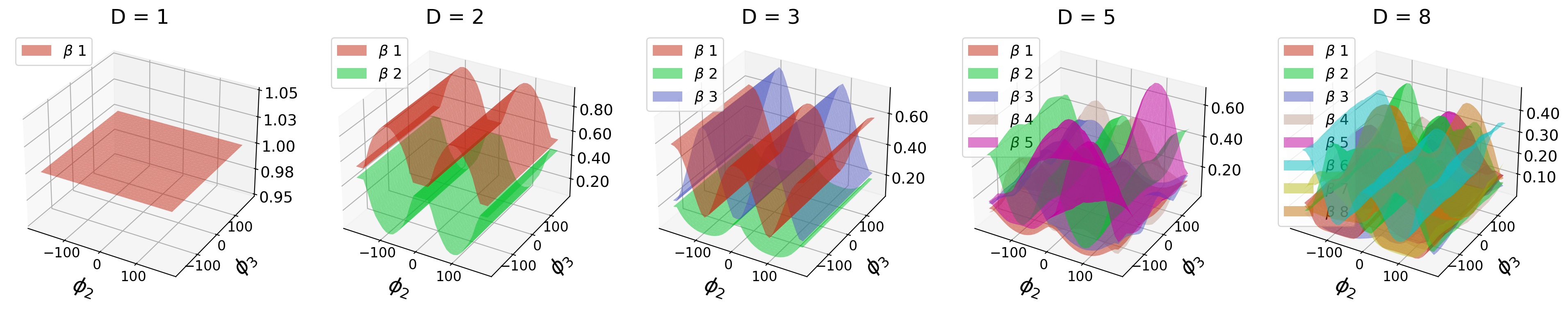}
     
    \caption{    
    \textbf{$\alpha-\beta$ space of SCNs trained for 3D rotation on ModelNet10 with LeNet5 inference architecture for $D=1..8$}.  
    Transformation parameters $\alpha$ result from applying $\cos(\cdot)$ and $\sin(\cdot)$ functions to the vector of rotation angles $(\phi_1, \phi_2, \phi_3)$, with each $\phi_i$ in the range $(-\pi, \pi)$. \textbf{Top:} Subspace of SCNs when changing $(\phi_1, \phi_2)$, and fixing $\phi_3=-\pi$. \textbf{Middle:} Subspace of SCNs when changing $(\phi_1, \phi_3)$, and fixing $\phi_2=-\pi$. \textbf{Bottom:}  Subspace of SCNs when changing $(\phi_2, \phi_3)$, and fixing $\phi_1=-\pi$.}
    \label{fig:3d:beta:all}
\end{figure*}

\Figref{fig:3d:beta:all} shows all views of the $\beta$-space of SCN for 3D rotation as a function of input parameters $\alpha=(\cos(\phi_1), \sin(\phi_1), \cos(\phi_2),$  $\sin(\phi_2), \cos(\phi_3), \sin(\phi_3))$, where $\phi_i, i=1..3$ is a rotation angle in the YZ, XZ and XY planes, respectively.
\Figref{fig:3d:beta:all} shows the whole $\beta$-space for 3D rotation presented as a function of all pairwise combinations of $\phi_i$.
In \Figref{fig:3d:beta:all} middle and bottom, $\beta$s show a stable trend along the $\phi_3$-axis, indicating that the 3D rotation in the XY plane keeps all object pixels (and is basically the same as 2D rotation in this case). In \Figref{fig:3d:beta:all} (top), $\beta$-space has cosine-like trend along both $\phi_1$ and $\phi_2$ axes, reflecting the 3D rotations in YZ and XZ planes. These transformations lead to information loss as some parts of an object rotate out of the view and get blocked. 
In all plots $\beta$-surfaces are not flat or degenerated. 
By observing the similarities and changing trends in the learned $\beta$-space for 3D rotation, it can be inferred that the shape of this configuration space primarily relies on the transformation itself and its associated parameters, namely $(\phi_1, \phi_2, \phi_3)$.
We provide a link\footnote{\url{https://subspace-configurable-networks.pages.dev/}} to an interactive website visualizing the $\beta$-space of sample SCNs, including those trained for 3D rotation.

\section{Search algorithm in the $\alpha$-space}
\label{sec:search:appendix}

This section provides details on the performance of the search algorithm which estimates $\alpha$ from a stream of input data. As mentioned in the main paper, we can leverage the fact that the correct input parameters $\alpha$ should produce a confident low-entropy classification result~\citep{Wortsman2020supsup,Hendrycks2016}. Therefore, our search algorithm estimates $\alpha$ from a batch of input data by minimizing the entropy of the model output on this batch by exploring the output of optimal models in the learned low-dimensional subspace. We use the basin hopping\footnote{\url{http://tinyurl.com/yp9ve2dd}} method to find the solution (with default parameters, \texttt{iter=100}, \texttt{T=0.1}, \texttt{method=BFGS}). 

The following code snippet runs the search in the $\alpha$-space to estimate the best rotation angle $\alpha$ from a batch of data $X$ by minimizing the function \texttt{f()}. The angle transformation function converts an input angle in degrees to the corresponding $(\cos, \sin)$ pair.

\lstset{firstnumber=1} 
\begin{lstlisting}
from scipy import optimize

# function to be minimized by the basin hopping algorithm
def f(z, *args):
    alpha = transform_angle(((1+z)*180)%360-180)
    X = args[0]
    logits = model(Tensor(X), hyper_x=Tensor(alpha))
    b = (F.softmax(logits, dim=1)) * (-1 * F.log_softmax(logits, dim=1)) # entropy
    return b.sum().numpy()

# given a batch of images find the rotation angle alpha using basin hopping algorithm
def findalpha(X):
    mkwargs = {"method": "BFGS", "args":X}
    res = optimize.basinhopping(f, 0.0, minimizer_kwargs=mkwargs, niter=100, T=0.1)
    alpha = ((1+res.x[0])*180)%360-180
    return alpha

# test search algorithm performance on a test set
result = 0.0
for (X, y) in test_loader:
    angle = random.uniform(-180, 180)
    X = TF.rotate(X, angle)

    alpha = findalpha(X)

    # compute model prediction with the estimated alpha
    logits = model(X, hyper_x=transform_angle(alpha))
    # y is the true label --> calculate accuracy
    correct = (logits.argmax(1) == y).type(torch.float).sum().item() / batch_size
    result += correct

result /= len(test_loader.dataset) / batch_size
print(f"Test accuracy: {(100*result):>0.1f}%")
\end{lstlisting}

To improve the accuracy of the search, SCN training is enhanced with an additional regularizer to minimize the model output entropy value for the correct $\alpha$ and maximise it for a randomly sampled $\alpha$. The train function is sketched in the listing below.

\lstset{firstnumber=1} 
\begin{lstlisting}
loss_fn = nn.CrossEntropyLoss()
optimizer = torch.optim.Adam(model.parameters(), lr=args.lr)
cos = nn.CosineSimilarity(dim=0, eps=1e-6)

def train(dataloader, model, loss_fn, optimizer):
    for (X, y) in dataloader:
        X, y = X.to(device), y.to(device)
        angle = random.uniform(0, 360)
        X = TF.rotate(X, angle)

        # make prediction and compute the loss
        pred = model(X, hyper_x=transform_angle(angle).to(device))
        loss = loss_fn(pred, y)

        # regularize (cosine similarity squared) in the beta space
        beta1 = model.hyper_stack(transform_angle(angle).to(device))
        angle2 = random.uniform(0, 360)
        beta2 = model.hyper_stack(transform_angle(angle2).to(device))
        loss += pow(cos(beta1, beta2),2)

        # minimize entropy for the correct degree
        b1 = (F.softmax(pred, dim=1)) * (-1 * F.log_softmax(pred, dim=1))
        loss += 0.01*b1.sum()

        # maximize entropy for a wrong / random degree
        logits = model(X, hyper_x=transform_angle(angle2).to(device))
        b2 = (F.softmax(logits, dim=1)) * (-1 * F.log_softmax(logits, dim=1))
        loss -= 0.01*b2.sum()

        optimizer.zero_grad()
        loss.backward()
        optimizer.step()
\end{lstlisting}

The interested reader can check the source code for further details.\footnote{\url{https://github.com/osaukh/subspace-configurable-networks/}}

Note that the basin hopping algorithms is computationally expensive. For the 2D rotation transformation on FMNIST dataset, the method may run several hundreds of model inferences to find a good solution. Optimizing the running time of the method is beyond the scope of this paper, because in practice $\alpha$-search can be avoided, \eg by using an additional sensor modality as input or by discretizing the search space to a manageable number of models. The expensive $\alpha$-search aims to show that the problem of estimating $\alpha$ and building I-SCNs is solvable in principle.

\section{SCNs on Low-resource Devices Vision Task}

\label{sec:eval:iot}

In Section~\ref{sec:eval:iot:fruit}, we showcase and measure the SCN's performance on fruit classification using RGB sensor. In this section, we show SCNs' performance on classifying traffic signs from 2D-rotated images on IoT devices.
For all tasks, additional sensor data is used to derive the input $\alpha$, elevating the need to perform $\alpha$-search on the device. The data gathered in our experiments is online.\footnote{\url{https://github.com/osaukh/subspace-configurable-networks/tree/main/IoT}}

We evaluate the models' test accuracy on the held-out datasets. On IoT devices, we compare SCN to the original One4All baseline, as well as to the wider and deeper One4All variants. To evaluate the time-efficiency of SCN, we measure the latency of three phases: executing the configuration network to obtain the vector $\beta$ ("Hypernet Inference"), computing $\theta$ from base models $\theta_i$ ("Configuration"), and executing the inference network $\mathcal{G}$ ("Inference"). All reported times present averages over 100 measurements.
The running time of the One4All model is solely determined by its inference time. We measure the required storage capacity by separately quantifying the flash and RAM usage of the SCN models, original One4All models, and modified One4All models. Our embedded experiments utilize two MCUs: Tensilica Xtensa 32-bit LX7 dual-core for traffic sign classification and nRF52840 for fruit classsification. 
One copy of the initialized model weights is stored in flash and loaded into RAM upon program start. The $D$ base models of SCNs are stored exclusively in flash, thereby conserving valuable RAM resources.

\begin{figure*}[!t]
    \centering
\begin{minipage}{0.45\textwidth}
\centering
  \includegraphics[angle=90,height=4.2cm]{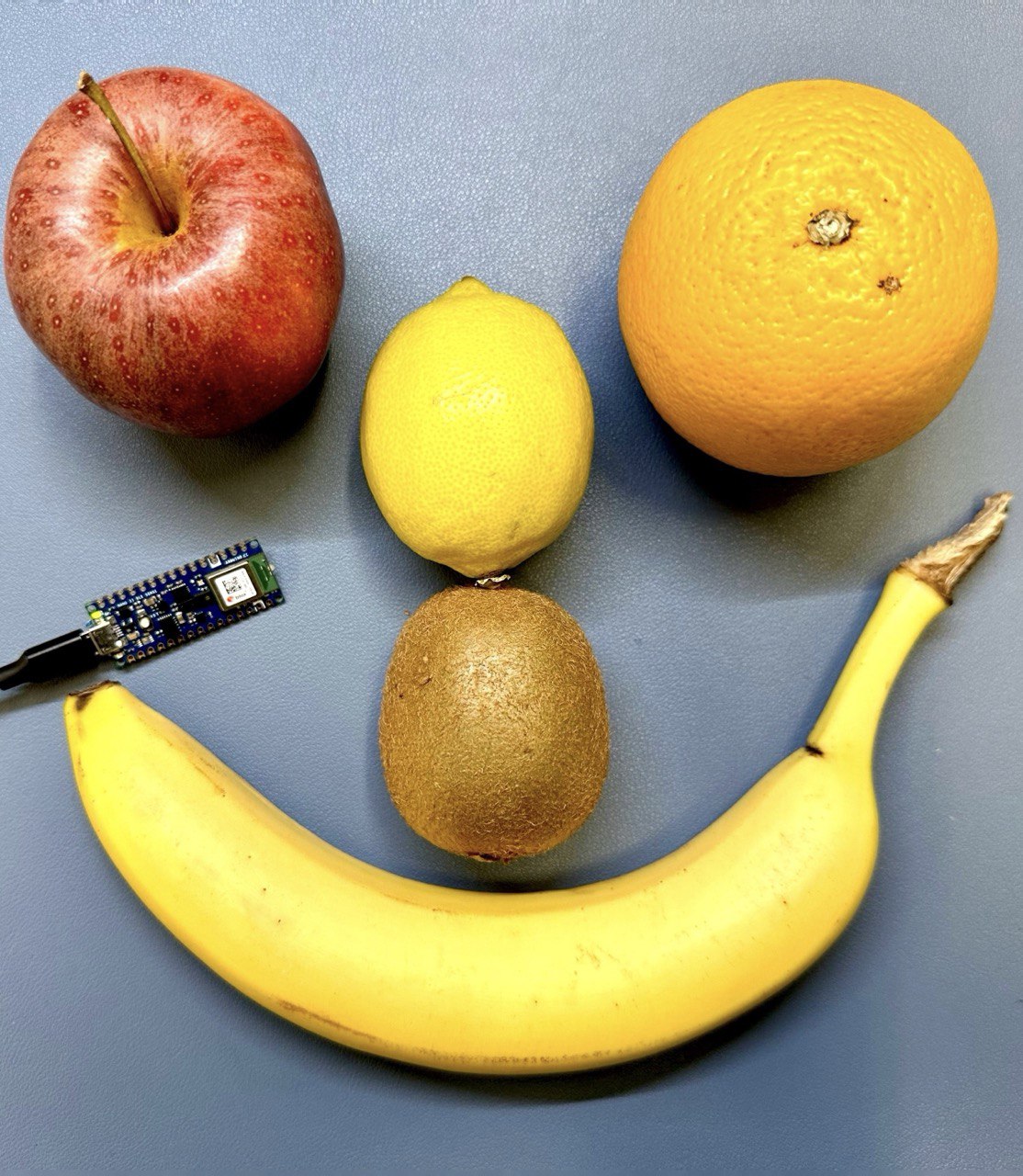}  
  \includegraphics[angle=90,height=3.5cm]
  {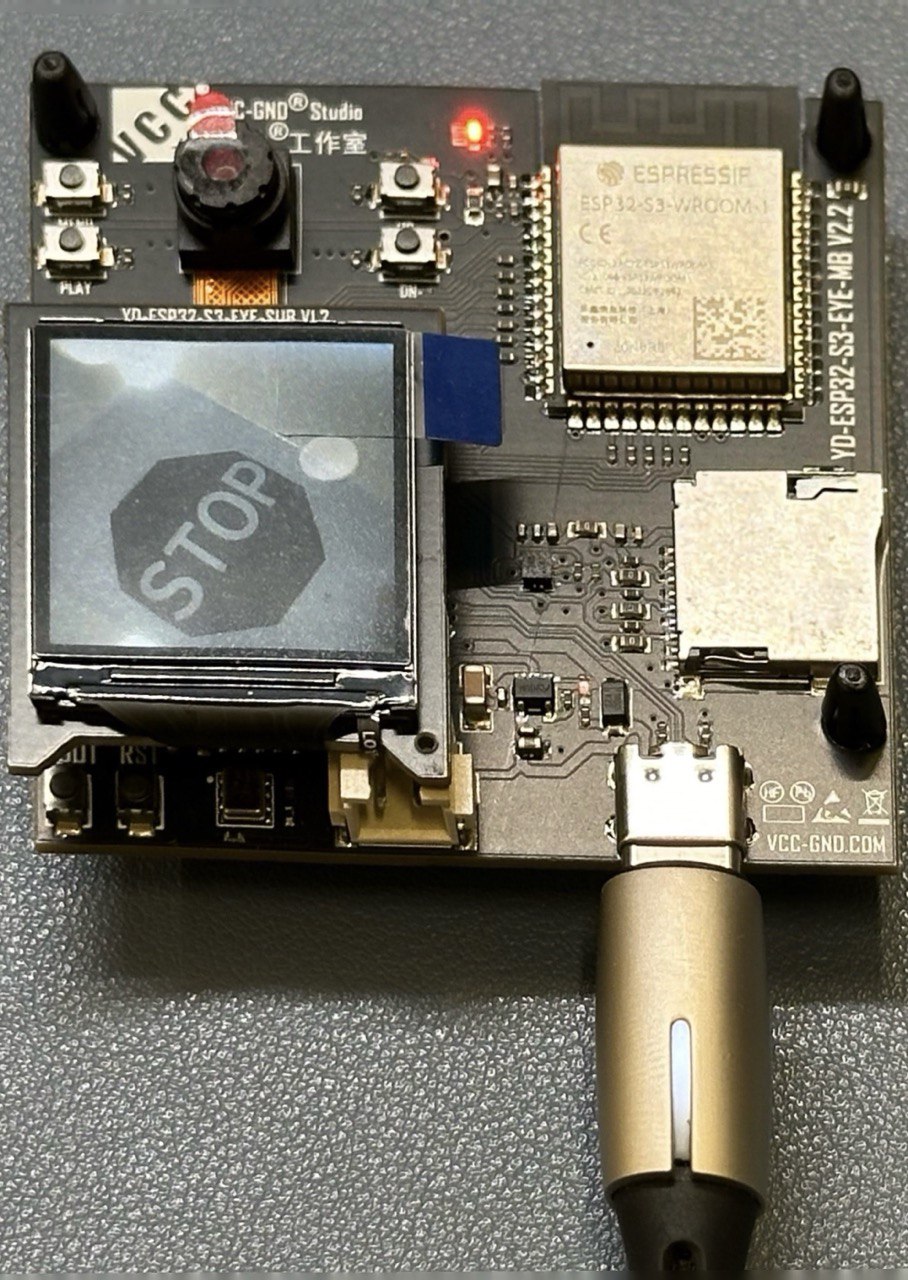}
\end{minipage}
% \hfill
\begin{minipage}{0.45\textwidth}
  \includegraphics[width=\linewidth]{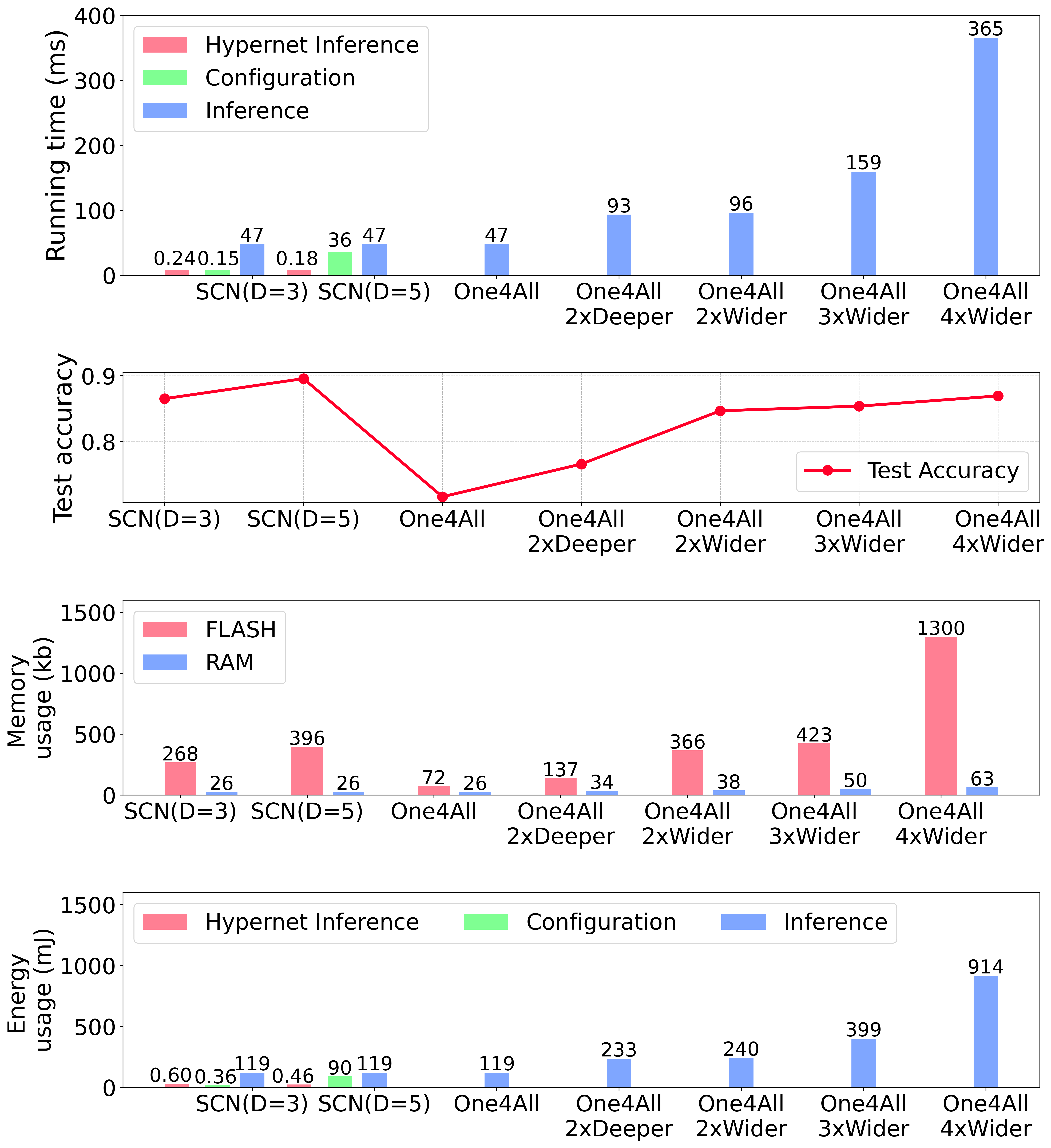}    
\end{minipage}    

    \caption{\textbf{SCN performance on the traffic sign classification task on ESP32-S3-EYE }, along with the visualization of both setups of the fruit classification task(see section~\ref{sec:eval:iot:fruit}) and traffic sign classification (left). From top to bottom the plots show: (1) Inference time in milliseconds, showcasing the efficiency of SCN, where the deeper and wider One4All variants lead to increased inference times. For SCNs, we also measure the execution latencies of the configuration network used to obtain vector $\beta$ ("Hypernet Inference"), and the computation time for generating $\theta$ from base models $\theta_i$ ("Configuration"). These latencies are only incurred when the deployment environment changes. (2) Test accuracy across various architectures, highlighting SCN's highly competitive performance. (3) RAM and flash memory usage in kB, indicating the increased resource consumption as the One4All model expands. (4) Energy consumption in mJ.}
    \label{fig:iot:performance}
\end{figure*}

\subsection{Traffic Sign Classification}
The German Traffic Sign Benchmark contains 39,209 images of 43 traffic signs captured on German roads~\citep{gtsdb}. During training, we rotate each traffic sign image at an arbitrary angle, supplied as $\alpha$ to the SCN's configuration network. We evaluate SCN's performance on 12,630 self-gathered traffic sign images, each fixed at a randomly chosen rotation angle. These images were collected using the onboard camera of the ESP32S3-EYE development board featuring Tensilica Xtensa 32-bit LX7 dual-core processor with 8 MB Octal PSRAM and 8 MB flash~\citep{esp32s3eye}. The data measured with the on-board IMU sensor of the ESP32S3-EYE development board is used to calculate the rotation angle during testing.

\Figref{fig:iot:performance} (right) presents the performance of SCN($D=3$) and SCN($D=5$) on classifying traffic signs from 2D-rotated images. The One4All model used in this experiment shares the same architecture as SCN: It features a sequence of three fully connected layers, with 12, 8, and 5 neurons respectively, each followed by a ReLU activation function. For the modified One4All models, including the $N$xDeeper and $N$xWider modifications, the number of layers or the number of hidden units per layer are increased. Figure \ref{fig:iot:performance} (right) shows that the One4All architecture, when made four times wider, matches the accuracy of SCN($D=3$). However, the resource consumption of SCN($D=3$) is notably lower: inference time and energy usage are reduced by the factor of 7.6, flash usage by the factor of 4.9, and RAM usage by the factor of 2.4. The overhead of reconfiguring the SCN model constitutes only a tiny fraction of the SCN inference time and is almost negligible. Despite sharing the same network architecture, One4All may lack sufficient capacity to store all the necessary augmentations for the desired input transformations. This could result in a significant drop in accuracy, which becomes more pronounced under tighter resource constraints.

\section{Related work}
\label{sec:relatedwork}
Networks trained on extensive datasets lack robustness to common transformations of the input, such as rotation~\citep{gandikota2021}, scaling~\citep{Ye2021}, translation~\citep{Biscione2021} or small deformations~\citep{Engstrom2017}. For example, \citet{Gong2014} showed that CNNs achieve neither rotation nor scale invariance, and their translation invariance to identify an object at multiple locations after seeing it at one location is limited~\citep{kauderer2018cnninv,Blything2020,Biscione2021}. Moreover, deep networks remain susceptible to adversarial attacks with respect to these transformations, and small perturbations can cause significant changes in the network predictions~\citep{gandikota2021}. There are three major directions of research to address the problem in the model design phase: Modifying the training procedure, the network architecture, or the data representation. Alternatively, the problem can be treated as a domain adaptation challenge and solved in the post-deployment phase. Below, we summarize the related literature.

\paragraph{Modifying the training scheme}
The methods that modify the training scheme replace the loss function $\mathcal{L}$ with a function that considers all parameters of transformations $T$ in a range where the solution is expected to be invariant. Common choices are minimizing the mean loss of all predictions $\{G(u(x_i), \theta) | u \in T\}$ resulting in training with data augmentation~\citep{Botev2022}, or maximizing the loss among all predictions leading to adversarial training~\citep{Engstrom2017}. Both training schemes do not yield an invariant solution with respect to transformations such as rotation, as discussed in \citep{gandikota2021}. The use of regularization can also improve robustness, yet provides no guarantees~\citep{Simard1991,yang2019invarianceinducing}. Overall, modifications of the training procedure are popular in practical applications, since they do not require characterization of the transformations applied to the input data, which are often unknown and may include a mix of complex effects.

\paragraph{Designing invariant network architectures}
Dedicated network architectures can be designed to be invariant to structured transformations based on a symmetric group action that preserves class labels. For example, it is commonly believed that convolutional neural networks are architecturally invariant to translation due to the design characteristics of their convolution and max pooling layers~\citep{Marcus2018,kauderer2018cnninv}. However, multiple studies argue that the translation invariance of CNNs is rather limited~\citep{Biscione2021,Blything2020}. 
Nevertheless, designing invariant architectures to a particular transformation is the subject of many recent works~\citep{Weiler2019,della2019survery} due to the desirable robustness properties they offer in practice~\citep{gandikota2021,Schneuing2022}.

Rotation invariant architectures play in important role in computer vision tasks. For instance, for a successful object classification, the orientation of the coordinate system should not affect the meaning of the data. Therefore, a broad research literature is devoted to designing rotation invariant and equivariant architectures. \citep{Cohen2016,Marcos2016,Veeling2018} use rotated filters to achieve layer-wise equivariance to discrete rotation angles. For continuous rotations, \citet{Worrall2016} proposed circular harmonic filters at each layer. These approaches were consolidated in \citep{Weiler2019}. 
\cite{Jaderberg2015,Tai2019} align transformed images using different methods, \eg using principal component analysis. \citet{pmlr-v162-wang22aa} explore approximately equivariant networks which relax symmetry-preserving constraints, since the real world rarely conforms to strict mathematical symmetry either due to noisy or incomplete data.
\citet{Weiler2018} and \citet{Thomas2018} propose 3D rotation equivariant kernels for convolutions. \citet{Esteves2017} propose a polar transformer network by learning a transformation in a polar space in which rotations become translations and so CNNs become effective to achieve rotation invariance. When 3D objects are presented as point clouds, this solves problems that arise due to object discretization, but leads to a loss of information about the neighbor relationship between individual points. \citet{Zhang2018graphCNN} add graph connections to compensate for this information loss and use graph convolutions to process the cloud points. \citet{Qi2016pointnet,Qi2017pointnetPlus} additionally include hierarchical and neighborhood information.

\paragraph{Canonicalization of data representation}
Input canonicalization is the process of converting the data into a specific form to simplify the task to be solved by a deep model. For example, by learning to map all self-augmentations of an image to similar representations is the main idea behind contrastive learning methods such as SimCLR~\citep{chen2020simclr} and Supervised Contrastive Learning~\citep{khosla2020supcon}. 
Canonicalization can also be achieved by learning to undo the applied transformation or learning a canonical representation of the data~\citep{kaba2022equivariance}. For example, \citet{Jaderberg2015} propose a Spatial Transformer layer to transform inputs to a canonical pose to simplify recognition in the following layers. BFT layers~\citep{tridao2019} can be used to learn linear maps that invert the applied transform. Earlier works on the topic considered manual extraction of features in the input that are robust, or invariant, to the desired transformation~\citep{Manthalkar2003,Yap2010}.

\paragraph{Domain adaptation}
Robustness to input transformations can be framed as a domain shift problem~\citep{loghmani2020}. Domain adaptation methods described in the literature follow different strategies as to how they align the source and the target. For example, \citet{ruijiaxu2018} define a distance measure between source and target data in the feature space and minimize this during training the network to reduce domain shift. \citet{RussoCTC17} train a generator that makes the source and target data indistinguishable for the domain discriminator. Another group of methods uses self-supervised learning to reduce domain shift~\citep{self-supervised-da2019}. In many real world scenarios, the data from the target domain are available only in the post-deployment phase. Therefore, domain adaptation methods often face memory and computing resource constraints making the use of backpropagation too costly. Partial model updates, especially those executed sequentially, may reduce model quality~\citep{Vucetic_2022}.

\paragraph{Linear mode connectivity, generalization and robustness}
In this work we show that optimal model weights that correspond to parameterized continuous transformations of the input reside in a low-dimensional linear subspace. This finding connects this work to recent research on the properties of the loss landscape and its relationship with generalization and optimization~\citep{geiger2019jamming, nguyen2018loss, fort2019deep, csimcsek2021geometry, juneja2022linear, entezari2021role, keller2022repair}. In particular, the existence of linear paths between solutions trained from independent initializations~\citep{entezari2021role}, those that share a part of their learning trajectories~\citep{frankle2020linear}, or trained on data splits~\citep{ainsworth2022git}. \citet{pmlr-v139-wortsman21a} learn neural network subspaces containing diverse and at the same time linear mode connected~\citep{frankle2020linear,nagarajan2019deterministic} solutions that can be effectively weight-space ensembled. This work builds upon and extends these works to linear mode connectivity between optimal models trained for different input transformation parameters.

\paragraph{Mixture-of-Experts}
Mixture-of-experts (MoE) networks split model parameters into several expert modules, designed to fit specialized sub-tasks~\citep{jacobs1991adaptive,shazeer2017outrageously}. In NLP, MoE successfully increases model capacity by adding multiple experts without exponential growth in computation cost~\protect\citep{gao2022parameter,caccia2022multi}. MoE utilizes a routing function to assign data to one or a subset of appropriate expert modules for further processing. The router takes weighted averaging of the outputs of the experts. In contrast, the configuration network of SCNs can also be regarded as a router to effectively generate low-dimensional coefficients and configure the inference model weights from the base models.\\\\
To the best of our knowledge, this is the first work looking into model robustness and adaptation to input transformations in the face of severe resource constraints.

\end{document}